\documentclass[10pt,journal,compsoc]{IEEEtran}

\usepackage{ragged2e}
\usepackage{times}
\usepackage{soul}
\usepackage{url}

\usepackage{hyperref}

\usepackage{microtype}
\usepackage{graphicx}
\usepackage{subfigure}
\usepackage{booktabs} 

\usepackage{hyperref}
\usepackage{algorithmic}

\makeatletter
\newcommand{\ShiftedIF}[1]{%
  \ALC@it~~~~\algorithmicif\ #1\ \algorithmicthen%
  \begin{ALC@if}}
\newcommand{\ShiftedELSE}{%
  \end{ALC@if}\ALC@it~~~~\algorithmicelse%
  \begin{ALC@if}}
\newcommand{\ShiftedENDIF}{%
  \end{ALC@if}\ALC@it~~~~\algorithmicendif}
\makeatother

\usepackage{amsmath}
\usepackage{amssymb}
\usepackage{mathtools}
\usepackage{amsthm}

\usepackage{enumitem}

\usepackage[colorinlistoftodos,prependcaption,textsize=tiny]{todonotes}

\usepackage{amsmath}

  \def\mB{{\mathcal B}}
  \def\mC{{\mathcal C}}
  \def\mD{{\mathcal D}}

  \def\mL{{\mathcal L}}

  \def\mX{{\mathcal X}}

  \DeclareMathAlphabet\mathbfcal{OMS}{cmsy}{b}{n}

  \def\0{{\bf 0}}
  \def\1{{\bf 1}}

  \def\bC{{\bf C}}

  \def\bZ{{\bf{Z}}}

  \def\bc{{\bf c}}

  \def\bx{{\bf x}}
  
  \def\bz{{\bf z}}

  \def\mmE{{\mathrm E}}

  \def\bx{{\bf x}}

  \def\bc{{\bf c}}
  \def\bz{{\bf z}}

\def\eg{\emph{e.g.}} 
\def\ie{\emph{i.e.}} 
 
\def\etc{\emph{etc.}} \def\vs{\emph{vs.}}
\def\wrt{{w.r.t.~}}

\usepackage[capitalize,noabbrev]{cleveref}

\newcommand{\rpami}[1]{{\color{black}#1}}

\usepackage{balance}
\usepackage{xspace}
\usepackage{pifont}
\usepackage{dashrule}

\newcommand{\methodname}{SAR\xspace}
\newcommand{\mysar}{SAR\xspace}
\newcommand{\mysarE}{SAR$^2$\xspace}

\definecolor{chengreen}{RGB}{17, 120, 100}

\usepackage{threeparttable}
\usepackage{multicol,multirow}
\usepackage{booktabs} 
\usepackage{bbding}

\theoremstyle{plain}

\theoremstyle{definition}

\theoremstyle{remark}

\usepackage{colortbl}  
\usepackage[textsize=tiny]{todonotes}

\usepackage{wrapfig}
\usepackage[table]{xcolor}

\ifCLASSOPTIONcompsoc
  \usepackage[nocompress]{cite}
\else
  \usepackage{cite}
\fi
\ifCLASSINFOpdf
\else
\fi

\usepackage{algorithm}
\usepackage{algorithmic}

\def\mytitle{
Adapt in the Wild: Test-Time Entropy Minimization with Sharpness and Feature Regularization
}

\begin{document}
%
\title{\mytitle}
%
%
%
%


\author{Shuaicheng Niu*,~\IEEEmembership{Member, IEEE,} Guohao~Chen*, Deyu Chen, Yifan Zhang,~\IEEEmembership{Member, IEEE,} \\ Jiaxiang Wu, Zhiquan Wen, Yaofo Chen, Peilin Zhao,~\IEEEmembership{Senior Member, IEEE,} \\ Chunyan Miao,~\IEEEmembership{Fellow, IEEE,} Mingkui~Tan$^\dagger$,~\IEEEmembership{Member, IEEE}  
\IEEEcompsocitemizethanks{
\IEEEcompsocthanksitem Shuaicheng Niu, Guohao Chen, and Chuanyan Miao are with the College of Computing and Data Science, Nanyang Technological University, Singapore. Email: \{shuaicheng.niu, guohao.chen, ascymiao\}@ntu.edu.sg.
\IEEEcompsocthanksitem Deyu Chen, Zhiquan Wen, Yaofo Chen, Mingkui Tan are with School of Software Engineering, South China University of Technology. Email: \{deyuchen, sewenzhiquan\}@mail.scut.edu.cn, \{chenyaofo, mingkuitan\}@scut.edu.cn.
\IEEEcompsocthanksitem Jiaxiang Wu is with ByteDance, China. The majority of this work was conducted while at Tencent AI Lab. Email: jiaxiang.wu.90@gmail.com.
\IEEEcompsocthanksitem Yifan Zhang is with the School of Computing, National University of Singapore. Email: yifan.zhang@u.nus.edu.
\IEEEcompsocthanksitem Peilin Zhao is with the School of Artificial Intelligence, Shanghai Jiao Tong University, Shanghai, China. Email: peilinzhao@hotmail.com.
\IEEEcompsocthanksitem * Authors contributed equally. $\dagger$ Corresponding author.
}

}

\markboth{Journal of \LaTeX\ Class Files, 2025}%
{Shell \MakeLowercase{\textit{et al.}}: \mytitle}

\IEEEtitleabstractindextext{%
\begin{abstract}
\justifying
Test-time adaptation (TTA) has shown to be effective at tackling distribution shifts between training and testing data by adapting a given model on test samples. However, the online model updating of TTA may be unstable and this is often a key obstacle preventing existing TTA methods from being deployed in the real world. Specifically, TTA may fail to improve or even harm the model performance when test data have: 1) mixed distribution shifts, 2) small batch sizes, and 3) online imbalanced label distribution shifts, which are quite common in practice. In this paper, we investigate the unstable reasons and find that the batch norm layer is a crucial factor hindering TTA stability. Conversely, TTA can perform more stably with batch-agnostic norm layers, \ie, group or layer norm. However, we observe that TTA with group and layer norms does not always succeed and still suffers many failure cases, \ie, the model collapses into trivial solutions by assigning the same class label for all samples. By digging into this, we find that, during the collapse process: 1) the model gradients often undergo an initial explosion followed by rapid degradation, suggesting that certain noisy test samples with large gradients may disrupt adaptation; and 2) the model representations tend to exhibit high correlations and classification bias. To address the above collapse issue, we first propose a sharpness-aware and reliable entropy minimization method, called SAR, for stabilizing TTA from two aspects: 1) remove partial noisy samples with large gradients, 2) encourage model weights to go to a flat minimum so that the model is robust to the remaining noisy samples. Based on SAR, we further introduce \mysarE to prevent representation collapse with two regularizers: 1) a redundancy regularizer, which reduces inter-dimensional correlations among centroid-invariant features; and 2) an inequity regularizer, which maximizes the prediction entropy of a prototype centroid, thereby penalizing biased representations toward any specific class. Promising results demonstrate that our methods perform more stably over prior methods and are computationally efficient under the above wild test scenarios. The source code is available at \url{https://github.com/mr-eggplant/SAR}.
\end{abstract}

\begin{IEEEkeywords}
Out-of-Distribution Generalization, Test-Time Adaptation, Sharpness-Aware Minimization, Feature Regularization.
\end{IEEEkeywords}}

\maketitle

\IEEEdisplaynontitleabstractindextext

\IEEEpeerreviewmaketitle

\ifCLASSOPTIONcompsoc
\IEEEraisesectionheading{\section{Introduction}\label{sec:introduction}}
\else
\section{Introduction}
\label{sec:introduction}
\fi

\IEEEPARstart{D}{eep}
neural networks achieve excellent performance when training and testing domains follow the same distribution~\cite{he2016deep,wang2018nonlocal,choi2018stargan}. However, when domain shifts exist, deep networks often struggle to generalize. Such domain shifts usually occur in real applications, since test data may unavoidably encounter natural variations or corruptions~\cite{hendrycks2019benchmarking,koh2021wilds}, such as the weather changes (\eg, \textit{snow, frost, fog}), sensor degradation (\eg, \textit{Gaussian noise, defocus blur}), and many other reasons. Unfortunately, deep models can be sensitive to the above shifts and suffer from severe performance degradation even if the shift is mild~\cite{recht2018cifar}. However, deploying a deep model on test domains with distribution shifts is still an urgent demand, and model adaptation is needed in these cases. 

Recently, numerous test-time adaptation (TTA) methods~\cite{sun2020test,wang2021tent,iwasawa2021test,bartler2022mt3,foa,mgtta,niu2022CLI} have been proposed to conquer the above domain shifts by online updating a model on the test data, which include two main categories, \ie, Test-Time Training (TTT)~\cite{sun2020test,liu2021ttt++} and Fully TTA~\cite{wang2021tent,niu2022EATA}. In this work, we focus on Fully TTA since it is more generally to be used than TTT in two aspects: i) it does not alter training and can adapt arbitrary pre-trained models to the test data without access to original training data; ii) it may rely on fewer backward passes (only one or less than one) for each test sample than TTT (see efficiency comparisons of TTT, Tent and EATA in Table~\ref{tab:methods_summary_supp}).

\begin{figure*}[t]
    \centering
    \includegraphics[width=1.\linewidth]{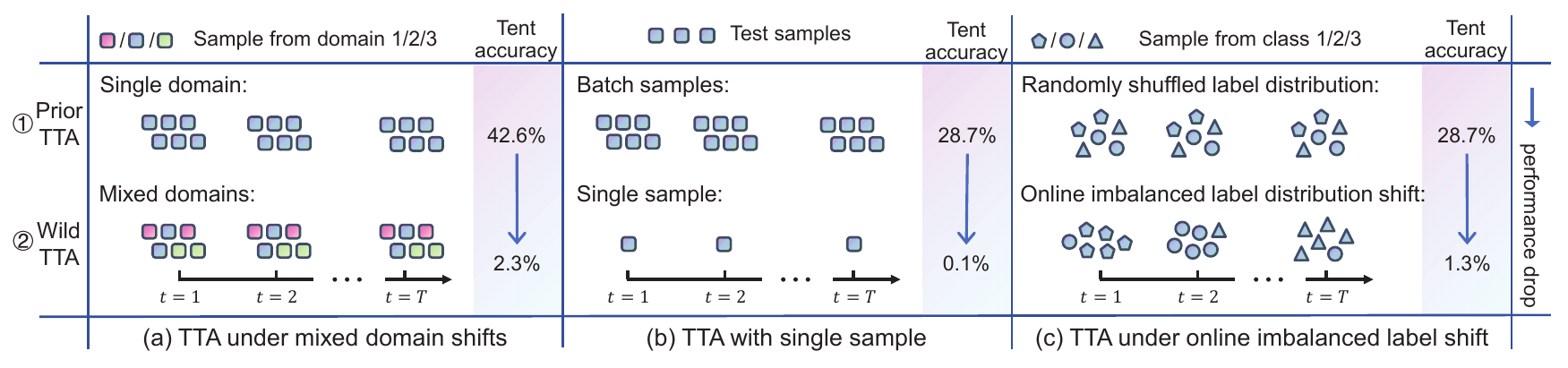}
    \vspace{-0.2in}
    \caption{An illustration of practical/wild test-time adaptation (TTA) scenarios, in which prior online TTA methods may degrade severely. The accuracy of Tent~\cite{wang2021tent} is measured on ImageNet-C of level 5 with ResNet50-BN (15 mixed corruptions in \textit{(a)} and Gaussian in \textit{(b-c)}).}
    \label{fig:3_weak_points_tta}
\end{figure*}

TTA has been shown boost model robustness to domain shifts significantly. However, its excellent performance is often obtained under some mild test settings, \eg, adapting with a batch of test samples that have the same distribution shift type and randomly shuffled label distribution (see Figure \ref{fig:3_weak_points_tta} \ding{192}). In the complex real world, test data may come arbitrarily. As shown in Figure~\ref{fig:3_weak_points_tta} \ding{193}, the test scenario may meet: i) mixture of multiple distribution shifts, ii) small test batch sizes (even single sample), iii) the ground-truth test label distribution $Q_t(y)$ is online shifted and $Q_t(y)$ may be imbalanced at each time-step $t$. 
In these wild test settings, online updating a model by existing TTA methods may be unstable, \ie, failing to help or even harming the model's robustness.

\begin{figure*}[t]
\centering
\includegraphics[width=0.95\linewidth]{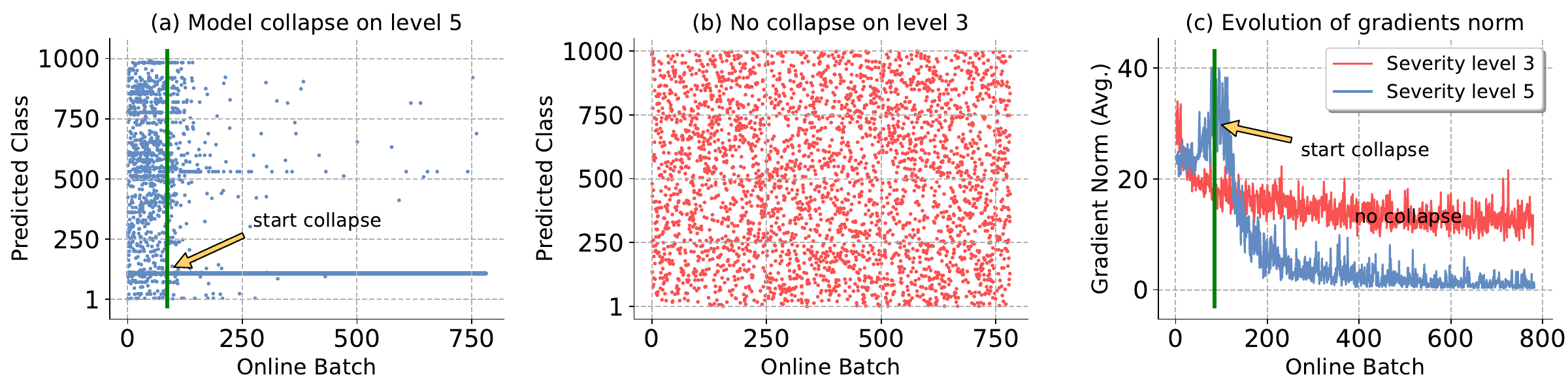}
\vspace{-0.15in}
\caption{Failure case analyses (a-c) of online test-time entropy minimization~\cite{wang2021tent}. (a) and (b) record the model predictions during online adaptation. (c) illustrates how gradients norm evolves with and without model collapse. All experiments are conducted on shuffled ImageNet-C of Gaussian noise with ResNet50 (GN), and a larger (severity) level denotes a more severe distribution shift.}
\label{fig:method_motivation}
\vspace{-0.05in}
\end{figure*}

In this paper, we first point out that the batch norm (BN) layer~\cite{ioffe2015batch} is a key obstacle since the mean and variance estimation in BN layers will be biased under the above wild scenarios. In light of this, we investigate the effects of norm layers in TTA (see Section~\ref{sec:empirical_norm_effects}) and find that pre-trained models with batch-agnostic norm layers (\ie, group norm (GN)~\cite{wu2018group} and layer norm (LN)~\cite{ba2016layer}) are more beneficial for stable TTA. However, TTA on GN/LN models does not always succeed and still has many failure cases. Specifically, GN/LN models optimized by online entropy minimization~\cite{wang2021tent} tend to occur collapse, \ie, predicting all samples to a single class  (see Figure~\ref{fig:method_motivation}), especially when the distribution shift is severe.  To address this issue, we propose a \textbf{s}harpness-\textbf{a}ware and \textbf{r}eliable entropy minimization method (namely \methodname). Specifically, we find that indeed some noisy samples that produce gradients with large norms harm the adaptation and thus result in model collapse. To avoid this, we filter test samples with large and noisy gradients out of adaptation according to their entropy. For the remaining samples, we introduce a sharpness-aware learning scheme to ensure that the model weights are optimized to a flat minimum, thereby being robust to the large and noisy gradients/updates.

\rpami{Recently, our observations have been widely adopted in the TTA community~\cite{wang2024search, liang2025comprehensive,tan2025uncertainty} and SAR has inspired a series of follow-up studies for TTA in the wild, including DeYO~\cite{lee2024deyo} that selects samples based on shape information for model updates,  ROID~\cite{marsden2024universal} that directly adjusts prediction priors under imbalanced shift, \etc~Nevertheless, they still analyze and devise methods to conquer the collapse issue at the model's output level, while the intrinsic factors driving model collapse remain underexplored.}

\rpami{
Therefore, we further move the lens one level deeper and ask:
\textit{What happens inside the model while collapse is (or is not) unfolding?} 
We start by analyzing the representation dynamics of model throughout TTA. Specifically, we define two feature-level statistics: 1)~feature redundancy to estimate dimension correlation; and 2) feature inequity that calculates the prediction entropy regarding a central feature, which quantifies the bias in representation. We compare the trajectories of these two statistics during a collapse and a non-collapse TTA process in Figure~\ref{fig:feature_motivation}. \linebreak Our results uncover a strong correlation between feature redundancy/inequity and TTA effectiveness. Feature redundancy declines or remains stable when TTA effectively enhances accuracy, but explodes immediately when collapse occurs. 
Meanwhile, feature inequity shows accelerated growth during early TTA under severe shift, which reveals an unfolding collapse inside the model.}

\rpami{
Inspired by the above insights, we propose to enhance \mysar with feature regularization during TTA, namely \mysarE. However, naively applying feature regularization to raw samples from the wild test data stream can harm the TTA performance (\eg, pushing features of the same class apart) or become ill-posed (\eg, using a single sample).
To address this, instead of regularizing the features of individual samples, \mysarE applies regularization to the class centroids, which effectively preserves intra-class compactness while penalizing degraded representations.
\mysarE introduces an exponentially updated feature bank to maintain a centroid per class over the test data stream, and provides a stable reference for the missing classes in the current mini-batch. 
Feature redundancy and inequity are then applied on class centroids augmented by the feature bank. Extensive experiments demonstrate that both feature redundancy and inequity regularization reduce the risk of collapse, and \mysarE substantially enhances the stability and efficacy of TTA in the wild.
Our contributions are summarized as follows:
}

\begin{itemize}[leftmargin=*]
    \item \rpami{We propose a Sharpness-Aware and Reliable (\mysar) optimization scheme for test-time entropy minimization. We analyze and verify empirically that batch-agnostic norm layers (\ie, group norm and layer norm) are more stable than batch norm for test-time adaptation under wild test settings, \ie, mix domain shifts, small test batch sizes, and online imbalanced label distribution shifts (Figure~\ref{fig:3_weak_points_tta}). Based on group/layer norm–equipped models, we further identify that the model collapse of test-time entropy minimization in wild test scenarios is linked to large/noisy gradients. Therefore, we devise \mysar to effectively mitigate this, by excluding unreliable samples from TTA and performing sharpness-aware updates to favor flat entropy minima.
   }
    
    \item \rpami{We further extend \mysar with feature \textbf{r}egularization (namely \mysarE) at test time. \mysarE introduces two metrics, \ie, feature redundancy and inequity, to provide deeper insights into the unstable learning process of wild TTA, revealing that declines in feature redundancy coincide with effective adaptation, while the accelerated growth in feature inequity foretells a collapse in representation (Figure~\ref{fig:feature_motivation}).
    Motivated by this, \mysarE develops a feature redundancy and inequity regularizer for TTA, and further introduces a feature bank and a centroid-based regularization method to stabilize test-time feature regularization in the wild.}

    \item \rpami{Extensive experiments demonstrate that \mysar consistently improves TTA stability under the challenging, wild test settings. Moreover, by explicitly measuring feature drifts and penalizing representation degradation, \mysarE introduces a more interpretable solution for mitigating collapse and substantially enhances the stability and effectiveness of TTA across various wild scenarios.}

\end{itemize}

\rpami{A short version of this work was published at ICLR 2023~\cite{niu2023sar} as oral presentation. This manuscript extends our preliminary version from the following aspects: 1) We study the model representation dynamics during TTA to understand how TTA affects a model and explain when it leads to effective or collapsing updates;
2) Based on our findings, we extend \mysar with feature redundancy and inequity regularizers for more effective TTA, and further introduce a feature bank and a centroid-based approach to stabilize feature regularization under the wild test scenarios;
3) We provide analyses of different regularizers for penalizing classification bias, verifying that our inequity regularizer outperforms information maximization's diversity regularizer~\cite{hu2017learning} in accuracy and calibration~\cite{naeini2015ece} by avoiding its implicit confidence flattening effects for online TTA;
}\rpami{
4) We provide extensive new empirical evaluations in the wild test settings with various model architectures, demonstrating that \mysarE achieves markedly better efficacy and stability over \mysar, \eg, +21.9\% accuracy and up to \textit{10}-fold performance variance under continuous TTA with imbalanced label shifts on ResNet50-GN.
}

\section{Related Work}\label{sec:related_work}

We relate our \mysar to existing adaptation methods without and with target data, and further review the sharpness-aware optimization mechanisms and the feature-space regularization approaches.

\textbf{Adaptation without Target Data.}
The problem of conquering distribution shifts has been studied in a number of works at \textit{training time}, including domain generalization~\cite{shankar2018generalizing,li2018learning,dou2019domain}, increasing the training dataset size~\cite{orhan2019robustness}, various data augmentation techniques~\cite{lim2019fast,hendrycks2020augmix,li2021feature,yao2022improving,wang2024inter,lin2025drivegen}, to name just a new. These methods aim to pre-anticipate or simulate the possible shifts of test data at training time, so that the training distribution can cover the possible shifts of test data. However, pre-anticipating all possible test shifts at training time may be infeasible and these training strategies are often more computationally expensive. Instead of improving generalization ability at training time, we conquer test shifts by directly learning from the test data.

\textbf{Adaptation with Target Data.} We divide the discussion on related methods that exploit target data into 1) unsupervised domain adaptation (adapt offline) and 2) test-time adaptation (adapt online).

$\bullet$
\textit{\textbf{Unsupervised domain adaptation (UDA).}} Conventional UDA jointly optimizes on the labeled source and unlabeled target data to mitigate distribution shifts, such as devising a domain discriminator to align source and target domains at feature level~\cite{pei2018multi,saito2018maximum,zhang2020collaborative,zhang2020covid} and aligning the prototypes of source and target domains through a contrastive learning manner~\cite{lin2022prototype}. Recently, \textit{source-free UDA} methods have been proposed to resolve the adaptation problem when source data are absent, such as generative-based methods that generate source images or prototypes from the model~\cite{li2020model,kundu2020universal,Qiu2021CPGA}, and information maximization~\cite{liang2020we,liang2021source}. These methods adapt models on a whole test set, in which the adaptation is offline and often requires multiple training epochs, and thus are hard to be deployed in online testing scenarios. 

$\bullet$
\textbf{\textit{Test-time adaptation (TTA).}} According to whether alter training, TTA methods can be mainly categorized into two groups. i) \textit{Test-Time Training (TTT)}~\cite{sun2020test} jointly optimizes a source model with both supervised and self-supervised losses, and then conducts self-supervised learning at test time. The self-supervised losses can be rotation prediction~\cite{gidaris2018unsupervised} in TTT or contrastive objectives~\cite{chen2020simple} in TTT++~\cite{liu2021ttt++} and MT3~\cite{bartler2022mt3}, \etc~ ii) \textit{Fully Test-Time Adaptation}~\cite{wang2021tent,niu2022EATA,hong2023mecta,lin2024monotta} does not alter the training process and can be applied to any pre-trained model, including adapting the statistics in batch normalization layers~\cite{schneider2020improving,hu2021mixnorm,khurana2021sita,lim2023ttn,zhao2023delta}, unsupervised entropy minimization~\cite{wang2021tent,niu2022EATA,zhang2021memo,cema}, prediction consistency maximization~\cite{zhang2021memo,wang2022continual,chen2022contrastive,spa}, \etc~Nevertheless, prior TTA methods are shown to be unstable in the online adaptation process and are sensitive to insufficient data (small batches), mixed domains, and online imbalanced shifted label distribution (see Figure~\ref{fig:3_weak_points_tta}). 
Our \mysar~\cite{niu2023sar} first analyzes what hinders the stability of TTA under the above practical test settings and proposes associated solutions to stabilize TTA under various wild test scenarios.

\rpami{Recently, the observations in SAR have been widely noted in the TTA community~\cite{liang2023ttasurvey} and SAR has encouraged a series of studies for TTA in the wild~\cite{gong2023sotta,marsden2024universal,lee2024deyo,hu2025beyond}. Memory-bank  methods~\cite{yuan2023robust,gong2023sotta,li2023generalized} introduce a data bank at testing to perform class-balanced sampling under online imbalanced label shifts.  In contrast, ROID~\cite{marsden2024universal} assumes a uniform target class prior and directly corrects label-imbalance via prior adjustment.
Considering the potential risk of model collapse in TTA, RDump~\cite{press2023rdumb} periodically resets model parameters during the continuous adaptation process, and TTA-Monitor~\cite{schirmer2025monitoring} tracks predictive performance and raises alerts when the model is at risk of collapse. To enhance learning stability, DeYO~\cite{lee2024deyo} exploits shape information to select test samples for entropy minimization, while ReCAP~\cite{hu2025beyond} further introduces local consistency to ensure entropy minimization favors predictions with large margins to all class boundaries. 
Nevertheless, prior approaches focus on mitigating collapse through input- or output-level strategies, while leaving the intrinsic causes of collapse underexplored.
In this work, we reveal the effects of TTA on model representation through the lens of redundancy and inequity, and further introduce test-time feature regularization as a more robust and interpretable framework to enhance the stability and efficacy of TTA across various test scenarios in the wild.
}

\textbf{Sharpness-aware Minimization (SAM).} SAM~\cite{foret2020sharpness} optimizes both a supervised objective (\eg, cross-entropy) and the sharpness of loss surface, aiming to find a flat minimum that has good generalization ability~\cite{hochreiter1997flat}. SAM and its variants~\cite{kwon2021asam,zheng2021regularizing,du2022efficient,chen2022vision} have shown outstanding performance on several deep learning benchmarks. In this work, when we analyze the failure reasons of test-time entropy minimization, we find that some noisy samples that produce gradients with large norms harm the adaptation and thus lead to model collapse. To alleviate this, we propose to minimize the sharpness of the test-time entropy loss surface so that the online model update is robust to those noisy/large gradients. 

\rpami{
\textbf{Feature-space Regularization.} At the training phase, feature regularization is widely used to enhance the generalization ability of the learned representations, \eg, self-supervised methods~\cite{zbontar2021barlow,bardes2022vicreg,hua2021feature} propose to reduce feature redundancy to prevent the risk of learning collapsed representations~\cite {Li2022understand}, decorrelated-normalization techniques~\cite {huang2018decorrelated,zhang2021stochastic} performs activation whitening within a batch to improve the efficiency of supervised training, \etc~While effective, these methods often require modifications to training or rely on the assumption of i.i.d. data and large batches to stabilize regularization, limiting their applicability. 
Unlike these methods, we exploit feature dynamics to study the collapse issue in fully TTA, and further introduce a feature bank and a centroid-based regularization method to stabilize feature regularization under wild testing scenarios.
}

\section{Preliminaries}\label{sec:preliminary}

We revisit two main categories of test-time adaptation methods in this section for the convenience of further analyses.

\textbf{Test-time Training (TTT).}
Let $f_{\Theta}(\bx)$ denote a model trained on $\mD_{train}=\{(\bx_i,y_i)\}_{i=1}^{N}$ with parameter $\Theta$, where $\bx_i\in\mX_{train}$ (training data space) and $y_i\in\mC$ (label space). The goal of test-time adaptation~\cite{sun2020test,wang2021tent} is to boost $f_{\Theta}(\bx)$ on out-of-distribution test samples $\mD_{test}=\{\bx_j\}_{j=1}^{M}$, where $\bx_{j}\in\mX_{test}$ (testing data space) and $\mX_{test}\neq\mX_{train}$. Sun et al., \cite{sun2020test} first propose the TTT pipeline, in which at \textbf{training phase} a model is trained on $\mD_{train}$ via both cross-entropy $\mL_{CE}$ and self-supervised rotation prediction $\mL_{S}$:
\begin{equation}
\min_{\Theta_b,\Theta_c,\Theta_s}\mathbb{E}_{\bx\in\mD_{train}}[\mL_{CE}(\bx;\Theta_b,\Theta_c) + \mL_{S}(\bx;\Theta_b,\Theta_s)],
\end{equation}
where $\Theta_b$ is the task-shared parameters (shadow layers), $\Theta_c$ and $\Theta_s$ are task-specific parameters (deep layers) for $\mL_{CE}$ and $\mL_{S}$, respectively. At \textbf{testing phase}, given a test sample $\bx$, TTT first updates the model with self-supervised task: $\Theta_b' \leftarrow \arg\min_{\Theta_b}\mL_{S}(\bx;\Theta_b,\Theta_s)$ and use the updated model weights $\Theta_b'$ to perform final prediction via $f(\bx;\Theta_b',\Theta_c)$.
\rpami{Methods like TTT++~\cite{liu2021ttt++} and TTT-MAE~\cite{gandelsman2022tmae} also explore using contrastive or reconstruction objectives as $\mL_S$.}

\textbf{Fully Test-time Adaptation (TTA).} The pipeline of TTT needs to alter the original model training process, which may be infeasible when training data is unavailable due to privacy/storage concerns. To avoid this, Tent~\cite{wang2021tent} proposes fully TTA, which provides the flexibility to adapt arbitrary pre-trained models for a given test mini-batch by conducting entropy minimization, formulated as: 
\begin{equation}\label{eq:entropy}
    \min E(\bx;\Theta)=-\sum_{c}f_{\Theta}(c|\bx)\log f_{\Theta}(c|\bx)
\end{equation}
where $c$ denotes the $c$-th class. \rpami{This method removes the need for an auxiliary self-supervised branch and is more efficient and broadly applicable than TTT, as shown in Table \ref{tab:methods_summary_supp}.}

\section{What Causes Unstable Test-time Adaptation under Wild Scenarios?}\label{sec:causes}
Although prior TTA methods have exhibited great potential for out-of-distribution generalization, its success may rely on some underlying test prerequisites (as illustrated in Figure~\ref{fig:3_weak_points_tta}): 1) test samples have the same distribution shift type; 2) adapting with a batch of samples each time, 3) the test label distribution is uniform during the whole online adaptation process, which, however, are easy to be violated in the wild world. In wild scenarios (Figure \ref{fig:3_weak_points_tta} \ding{193}), prior methods may perform poorly or even fail.

\begin{figure*}[t]
\centering
\includegraphics[width=.95\linewidth]{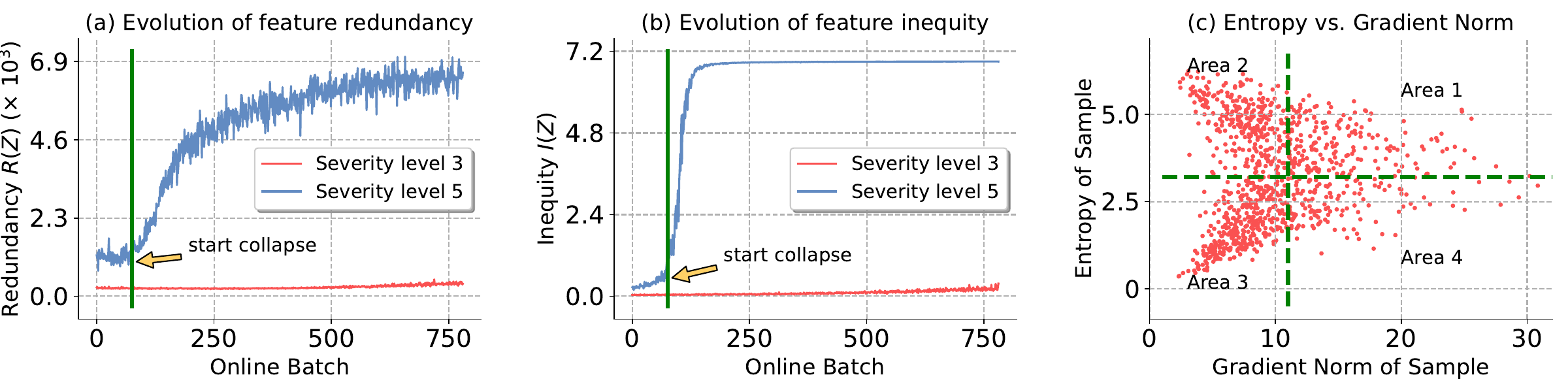}
\vspace{-0.15in}
\caption{Failure case analyses (a-b) of online test-time entropy minimization~\cite{wang2021tent}. (a) and (b) illustrates how feature redundancy and inequity evolve during TTA with and without model collapse. (c) investigates the relationship between the sample's entropy and the gradient norm. All experiments are conducted on shuffled ImageNet-C of Gaussian noise with ResNet50 (GN), and a larger (severity) level denotes a more severe distribution shift.}
\label{fig:feature_motivation}
\vspace{-0.1in}
\end{figure*}

\rpami{In this section, we seek to analyze the underlying reasons why TTA fails under wild testing scenarios described in Figure~\ref{fig:3_weak_points_tta} from a unified perspective and then propose associated solutions (c.f. Section~\ref{sec:overall_sar_method}). We start by investigating the impacts of norm layer on TTA stability and further dig into the failure modes of entropy-based methods with batch-agnostic norms, \eg, group norm, from both the output and feature level. Our findings are depicted below.}

\textbf{Batch Normalization Hinders Stable TTA.}\label{sec:norm_effects}
In TTA, prior methods often conduct adaptation on pre-trained models with batch normalization (BN) layers~\cite{ioffe2015batch}, and most of them are built upon test-time BN statistics calibration~\cite{schneider2020improving,nado2020evaluating,khurana2021sita,wang2021tent,niu2022EATA,hu2021mixnorm,zhang2021memo}. Specifically, for a layer with $d$-dimensional input $\bx=\left(x^{(1)} \ldots x^{(d)}\right)$, the batch normalized output are:
$
    y^{(k)}=\gamma^{(k)} \widehat{x}^{(k)}+\beta^{(k)}, \text{~~where~~} \widehat{x}^{(k)}=\big(x^{(k)}-\mmE\left[x^{(k)}\right]\big)\big/{\sqrt{\operatorname{Var}\left[x^{(k)}\right]}}.
$
Here, $\gamma^{(k)}$ and $\beta^{(k)}$ are learnable affine parameters. BN adaptation methods calculate mean $\mmE[x^{(k)}]$ and variance $\operatorname{Var}[x^{(k)}]$ over (a batch of) \textbf{test samples.} 

However, in wild TTA, all three practical adaptation settings (in Figure~\ref{fig:3_weak_points_tta}) in which TTA may fail will result in problematic mean and variance estimation.  \textbf{First}, BN statistics indeed represent a distribution and ideally each distribution should have its own statistics. Simply estimating shared BN statistics of multiple distributions from mini-batch test samples unavoidably obtains limited performance, such as in multi-task/domain learning \cite{wu2021rethinking}. \textbf{Second}, the quality of estimated statistics relies on the batch size, and it is hard to use very few samples (\ie, small batch size) to estimate it accurately. \textbf{Third}, the imbalanced label shift will also result in biased BN statistics towards some specific classes in the dataset. Based on the above, we posit that batch-agnostic norm layers, \ie, agnostic to the way samples are grouped into a batch, are more suitable for performing TTA, such as group norm (GN)~\cite{wu2018group} and layer norm (LN)~\cite{ba2016layer}, and devise our method based on GN/LN models in Section \ref{sec:overall_sar_method}.
To verify the above claim, we empirically investigate the effects of different normalization layers (including BN, GN, and LN) in TTA (including TTT and Tent) in Section~\ref{sec:empirical_norm_effects}. From the results, we observe that models equipped with GN and LN are more stable than models with BN when performing online test-time adaptation under three practical test settings (in Figure~\ref{fig:3_weak_points_tta}) and have fewer failure cases. The detailed empirical studies are put in Section~\ref{sec:empirical_norm_effects} for the coherence of presentation.

\textbf{Online Entropy Minimization Tends to Result in Collapsed Trivial Solutions, \ie, Predict All Samples to the Same Class.}
Although TTA performs more stable on GN and LN models, it does not always succeed and still faces several failure cases.
For example, entropy minimization (Tent) on GN models (ResNet50-GN) tends to collapse, especially when the distribution shift extent is severe. In this paper, we aim to stabilize online fully TTA under various practical test settings. To this end, we first analyze the failure reasons, in which we find models are often optimized to collapse trivial solutions. We illustrate this issue in the following.

During the online adaptation process, we record the predicted class and the gradients norm (produced by entropy loss) of ResNet50-GN on shuffled ImageNet-C of Gaussian noise. By comparing Figures~\ref{fig:method_motivation} (a) and (b), entropy minimization is shown to be unstable and may occur collapse when the distribution shift is severe (\ie, severity level 5). From Figure~\ref{fig:method_motivation} (a), as the adaptation goes by, the model tends to predict all input samples to the same class, even though these samples have different ground-truth classes, called model collapse. Meanwhile, we notice that along with the model starts to collapse the $\ell_2$-norm of gradients of all trainable parameters \rpami{increases significantly} and then degrades to almost 0 (as shown in Figure~\ref{fig:method_motivation} (c)), while on severity level 3 the model works well and the gradients norm keep in a stable range all the time. This indicates that some test samples produce large gradients that may hurt the adaptation and lead to model collapse.

\rpami{\textbf{Feature Redundancy and Inequity are Model Collapsing Signals.} 
Our previous study reveals that TTA may push the network into a trivial solution, yet the underlying mechanisms remain under-explored. 
As shown in Figure~\ref{fig:method_motivation} (c), we notice a continual increase in the gradient norm before model collapse, implying collapse as a cumulative consequence which may be tractable. 
To understand how online TTA gradually affects a model and when it leads to successful or failed adaptations, we further dig into the shifts of model behaviors at the internal \textit{feature level}.
Specifically, we monitor two feature-level statistics throughout TTA:\linebreak 
1)~\textbf{Redundancy}, which measures the similarity between each dimension of model representation, where highly correlated dimensions increase the estimated redundancy; }
\rpami{
and 2) \textbf{Inequity}, which quantifies the extent of model bias towards predicting some particular classes, measured by the prediction entropy of a centroid computed from features across different classes. The intuition is that if each class contributes fairly to the classification task, the centroid of features from all classes should be classified with maximum uncertainty—meaning the classifier would assign nearly uniform probabilities to all classes for this centroid. 
Formally, Let $z_b=g_{\Theta}(x_b)\in \mathbb{R}^{D}$ and $\bZ=[z_1,\ldots,z_B]$ denote the features of a batch, and $\mu_B=\frac{1}{B}\sum_{b} z_{b}$ be the centroid of the batch feature. Feature redundancy $R(\cdot)$ and inequity $I(\cdot)$ are then given by:
\begin{equation}\label{eq:initial_redundancy}
    R(\bZ)=\frac{1}{D-1}\sum_{i\neq j} \mC_{ij}^{2},
~~
\mC=\frac{1}{B}\cdot \frac{(\bZ-\mu_B)^{\top}(\bZ-\mu_B)}{\sigma_B(\bZ)^2};
\end{equation}
\begin{equation}\label{eq:initial_inequity}
I(\bZ)= \log C + \sum_{c} p'_{c}\log p'_{c},
~~
p'=\operatorname{softmax}\bigl(h(\mu_B)\bigr),
\end{equation}
where $\sigma_B(\cdot)$ is the standard variance, $\mC$ is the covariance matrix, $C$ is the number of classes, and $h(\cdot)$ is the classifier head.
}

\rpami{We conduct experiments on the ResNet50-GN model following Figure~\ref{fig:method_motivation} and Eqns.~(\ref{eq:initial_redundancy} \& \ref{eq:initial_inequity}) are calculated over the current test batch. Our results in Figure~\ref{fig:feature_motivation} reveal two key observations: 
1)~\textit{Shift severity determines the initial health of the representation.} A more severe distribution shift (\ie, severity level 5) induces larger redundancy and inequity in model representation before adaptation takes place, as in the first points of Figure 3 (a-b). Appendix~\ref{sec:more_discussions} further confirms such monotonic increment across all five severity levels, providing a new understanding of how distribution shift affects a model's performance.
2)~\textit{The changes in feature redundancy and inequity foretell TTA effectiveness and collapse.} During TTA on severity level 5, feature redundancy $R(\bZ)$ first remains stable (or decreases on VitBase, see Appendix~\ref{sec:vit_feature_evolution}) when TTA still improves accuracy, followed by a significant increase that positions model collapse. Meanwhile, feature inequity $I(\bZ)$ shows accelerated growth during early adaptation, which then results in model collapse and a further explosion in feature inequity. In contrast, when TTA works well under severity level 3, feature inequity and redundancy remain within a stable range. This underscores a strong correlation between feature inequity/redundancy and model collapse, offering new insights to improve TTA stability and effectiveness in the wild.
}

\section{Stable and Effective Test-time Adaptation in Dynamic Wild World}\label{sec:overall_sar_method}

\rpami{Based on the above analyses, we aim to resolve model collapse during TTA in the wild from both the model's output level and the feature level: At the output level, we first select reliable test samples with small entropy values to reduce large and noisy gradients, which are then used for sharpness-aware entropy minimization (\textit{namely} SAR, c.f. Section~\ref{sec:sar_method}); At the feature level, we introduce a prototype feature bank to enable test-time feature regularization for TTA in the wild settings, where we apply both redundancy and inequity regularization along with entropy minimization to mitigate feature collapse and enhance the TTA effectiveness (\textit{namely} \mysarE, c.f. Section~\ref{sec:sar-v2-method}).
} 

\subsection{Sharpness-aware Reliable Entropy Minimization}\label{sec:sar_method}

Based on the above analyses, two most straightforward solutions to \rpami{mitigate} model collapse \rpami{during entropy minimization} are filtering out test samples according to the sample gradients or performing gradients clipping. However, these are not feasible since the gradients norms for different models and distribution shift types have different scales, and thus it is hard to devise a general method to set the threshold for sample filtering or gradient clipping (see Section~\ref{sec:compare_grad_clip} for more analyses). We propose our solutions as follows. 

\textbf{Reliable Entropy Minimization.} Since directly filtering samples with gradients norm is infeasible, we first investigate the relation between entropy loss and gradients norm and seek to remove samples with large gradients based on their entropy. Here, the entropy depends on the model's output class number $C$ and it belongs to $(0,\ln C)$ for different models and data. In this sense, the threshold for filtering samples with entropy is easier to select. As shown in Figure~\ref{fig:method_motivation} (d), selecting samples with small loss values can remove part of samples that have large gradients (\texttt{area@1}) out of adaptation. Formally, let $E(\bx; \Theta)$ be the entropy of sample $\bx$, the selective entropy minimization is defined by:
\begin{equation}
    \min _{{\Theta}} S(\bx) E(\bx; \Theta), ~~\text{where}~~ S(\bx)\triangleq \mathbb{I}_{\left\{E(\bx; \Theta)<E_{0}\right\}}(\bx).
    \label{eq:reliable_entropy}
\end{equation}
Here, $\Theta$ denote model parameters, $\mathbb{I}_{\{\cdot\}}(\cdot)$ is an indicator function and $E_0$ is a pre-defined parameter. Note that the above criteria will also remove samples within \texttt{area@2} in Figure~\ref{fig:method_motivation} (d), in which the samples have low confidence and thus are unreliable~\cite{niu2022EATA}.

\textbf{Sharpness-aware Entropy Minimization.} 
Through Eqn.~(\ref{eq:reliable_entropy}), we have removed test samples in \texttt{area@1\&2} in Figure~\ref{fig:method_motivation} (d) from adaptation. Ideally, we expect to optimize the model via samples only in \texttt{area@3}, since samples in \texttt{area@4} still have large gradients and may harm the adaptation. However, it is hard to further remove the samples in \texttt{area@4} via a filtering scheme. Alternatively, we seek to make the model insensitive to the large gradients contributed by samples in \texttt{area@4}. Here, we encourage the model to go to a flat area of the entropy loss surface. The reason is that a flat minimum has good generalization ability and is robust to noisy/large gradients, \ie, the noisy/large updates over the flat minimum would not significantly affect the original model loss, while a sharp minimum would. To this end, we jointly minimize the entropy and the sharpness of entropy loss by:
\begin{equation}
    \min _{{\Theta}} E^{SA}{(\bx; \Theta)}, ~~\text { where } ~~ E^{SA}{(\bx; \Theta)}  \triangleq  \max _{\|\boldsymbol{\epsilon}\|_{2} \leq \rho} E(\bx; \Theta+\boldsymbol{\epsilon}).
    \label{eq:sa_entropy}
\end{equation}
Here, the inner optimization seeks to find a weight perturbation $\boldsymbol{\epsilon}$ in a Euclidean ball with radius $\rho$ that maximizes the entropy. The sharpness is quantified by the maximal change of entropy between $\Theta$ and $\Theta+\boldsymbol{\epsilon}$. This bi-level problem encourages the optimization to find flat minima.
To address problem~(\ref{eq:sa_entropy}), we follow SAM~\cite{foret2020sharpness} that first approximately solves the inner optimization via first-order Taylor expansion, \ie,
\begin{equation}
\begin{aligned}
    \boldsymbol{\epsilon}_{E}^{*}(\bx; \Theta) &\triangleq \underset{\|\boldsymbol{\epsilon}\|_{2} \leq \rho}{\arg \max }E(\bx; \Theta+\boldsymbol{\epsilon}) \\
    &\approx \underset{\|\boldsymbol{\epsilon}\|_{2} \leq \rho}{\arg \max } E(\bx;\Theta) + \boldsymbol{\epsilon}^{T} \nabla_{\Theta} E(\bx;\Theta) \\
    &= \underset{\|\boldsymbol{\epsilon}\|_{2} \leq \rho}{\arg \max } \boldsymbol{\epsilon}^{T} \nabla_{\Theta} E(\bx;\Theta).
\end{aligned}
\label{eq:epsilon_star}
\end{equation}

Then, $\hat{\boldsymbol{\epsilon}}_{E}(\bx;\Theta)$ that solves this approximation is given by the solution to a classical dual norm problem:
\begin{equation}\label{eq:hat_epsilon}
    \hat{\boldsymbol{\epsilon}}_{E}(\bx;\Theta)=\rho \nabla_{\Theta} E(\bx;\Theta) /\left\|\nabla_{\Theta} E(\bx;\Theta)\right\|_{2}.
\end{equation}
Substituting $\hat{\boldsymbol{\epsilon}}(\Theta)$ back into Eqn.~(\ref{eq:sa_entropy}) and differentiating, by omitting the second-order terms for computation acceleration, the final gradient approximation is:
\begin{equation}
    \left.\nabla_{\Theta} E^{SA}(\bx;\Theta) \approx \nabla_{\Theta} E(\bx;\Theta)\right|_{\Theta+\hat{\boldsymbol{\epsilon}}_{E}(\bx;\Theta)}.
\end{equation}
Then, our sharpness-aware reliable entropy minimization for test-time model update (\textit{termed} \mysar) is given by:
\begin{equation}
    \min _{\tilde{\Theta}} S(\bx) E^{SA}(\bx; \Theta). 
    \label{eq:reliable_and_sharp-aware_entropy}
\end{equation}
\rpami{Here, $\tilde{\Theta}\subset\Theta$ denote learnable parameters. 
In addition, to handle extremely hard cases where \mysar may still fail, we also introduce a Model Recovery Scheme to guarantee a minimum level of stability. We maintain a moving average $e_m$ of entropy loss values and reset $\tilde{\Theta}$ to be the original once $e_m$ falls below a small threshold $e_0$, since collapsed models typically produce abnormally (very) small entropy losses. Here, the additional memory costs are negligible since we only optimize affine parameters in norm layers. Notably, this recovery scheme was triggered only three times by \mysar (\textit{see} Appendix~\ref{sec:more_ablation_for_sar}) and never by \mysarE throughout all experiments across diverse wild settings and corruption/domain-shift scenarios. 
}

\subsection{Test-Time Feature Regularization in the Wild}
\label{sec:sar-v2-method}
\rpami{Our analysis in Section~\ref{sec:causes} reveals that model representations tend to become redundant and unequal (\ie, bias towards certain classes) during a collapsed TTA process.
To address this, a simple solution would be to use feature regularization along with TTA. Methods such as VicReg~\cite{bardes2021vicreg} and Barlow Twins~\cite{zbontar2021barlow} explore using the feature redundancy regularizer in self-supervised learning. Nevertheless, their effectiveness relies on the assumption of i.i.d. data and a large batch size. In online TTA, this assumption generally fails since: \textit{C1}) test data can be highly correlated and produce non-i.i.d. data stream, \ie, the online imbalanced label shift scenario; and \textit{C2}) real-world TTA adapts on micro batches or even a single sample at a time. 
In this case, directly applying the feature regularizer on test data may harm the TTA performance (\eg, enforcing dissimilar features among the same class) or become ill-posed (\eg, using a single sample), and effective feature regularization at test time remains challenging.}

\begin{algorithm}[t]
	\caption{The pipeline of proposed \mysar and \mysarE.}
	\label{alg:sar-algorithm}
	\begin{algorithmic}[1]
    \REQUIRE{Test samples $\mD_{test}{=}\{\bx_j\}_{j=1}^{M}$, model $f_{\Theta}(\cdot)$ with its feature extractor $g(\cdot)$ and classifier $h(\cdot)$, trainable parameters $\tilde{\Theta}\subset\Theta$, feature bank $\tilde{\bC}$, feature bank warm-up threshold $\zeta$.}
    \STATE Initialize $\tilde{\Theta}_0=\tilde{\Theta}$, feature bank $\tilde{\bC}=\varnothing$.
    \FOR{a batch $\mB^t\small{=}\{\bx_j\}_{j=1}^{B}$ in $\mD_{test}$}
    \STATE Compute features $\bz_j\small{=}g(\bx_j)$ and predictions $\hat{y}_j\small{=}h(\bz_j)$. 
    \STATE \texttt{For \mysar:}
    \STATE ~~~~Calculate entropy loss $E(\bx_j;\Theta)$ via Eqn.~(\ref{eq:entropy}).
    \STATE ~~~~Update model ($\tilde{\Theta}\subset\Theta$) with Eqn.~(\ref{eq:reliable_and_sharp-aware_entropy}).
    \STATE \texttt{For \mysarE:}
    \STATE ~~~~Retrieve augmented features $\bC^t$ via Eqn.~(\ref{eq:features_from_bank}).
    \ShiftedIF{$\bC^t$ contains more than $\zeta$ features}
    \STATE ~~~~Calculate entropy loss $E(\bx_j;\Theta)$ via Eqn.~(\ref{eq:entropy}).
    \STATE ~~~~Compute feature redundancy $R(\bC^t;\Theta)$ via Eqn.~(\ref{eq:initial_redundancy}).
    \STATE ~~~~Compute feature inequity $I(\bC^t;\Theta)$ using Eqn.~(\ref{eq:initial_inequity}).
    \STATE ~~~~Update model ($\tilde{\Theta}\subset\Theta$) with Eqn.~(\ref{eq:sar_extend_optimization}).
    \ShiftedENDIF
    \STATE ~~~~Refresh the feature bank $\tilde{\bC}$ via Eqn.~(\ref{eq:bank_update}).
    \ENDFOR
    \ENSURE The predictions $\{\hat{y}_j\}_{j=1}^M$ for all $\bx \in \mD_{test}$.
	\end{algorithmic}
\end{algorithm}

\rpami{\textbf{Feature Bank for Online Representation Regularization under Wild TTA.} 
To address the above pitfalls, we introduce two simple but necessary modifications: 
1)~We reformulate feature regularization to operate on class centroids instead of on individual samples.
This reformulation mitigates collapse by encouraging full utilization of the feature space across classes, while bypassing the risks of pushing the features of the same class apart, \ie, efficiently addressing challenge~\textit{C1}.
2)~We maintain an exponentially updated feature bank that stores one centroid per class over the test data stream. This bank provides a stable reference for the missing classes even when the test batch contains a single sample, \ie, tackling challenge~\textit{C2}.}
\rpami{
Formally, let $\mathcal{B}^t$ be a mini-batch, $g_{\Theta}(\bx)\in\mathbb{R}^{D}$ be the extracted feature of sample $\bx$, and $\hat{y}(\bx)$ denote the predicted label, we compute a centroid $\bc_i$ for every class that \textit{appears} in the mini-batch as:
\begin{equation}
\bc_{i}\;=\;\frac{\sum_{\bx\in\mathcal{B}}\! \mathbf{1}\!\bigl[\hat{y}(\bx)=i\bigr]\cdot g_{\Theta}(\bx)}
{\sum_{\bx\in\mathcal{B}}\! \mathbf{1}\!\bigl[\hat{y}(\bx)=i\bigr]}, ~~~~i\in\hat{\mathcal{Y}}_{\mathcal{B}^t}.
\end{equation}
where $\hat{\mathcal{Y}}_{\mathcal{B}^t}$ is the set of pseudo-labeled classes present in~$\mathcal{B}^t$. Meanwhile, for classes \textit{absent} from $\mathcal B^t$, we refer class feature from the feature bank $\tilde{\bC}{=}\{\tilde{c}_i\}_{i=1}^C$, and the feature matrix $\bC^t$ of a mini-batch $\mathcal{B}^t$ for stable test-time feature regularization becomes:
\begin{equation}~\label{eq:features_from_bank}
\mathbf{C}^{t}_{i}=
\begin{cases}
\mathbf{c}_{i}, & i\in\hat{\mathcal{Y}}_{\mathcal{B}^t},\\[4pt]
\tilde{\mathbf{c}}_{i}, & \text{otherwise},
\end{cases}
\end{equation}
The bank is initialized as empty, and class centroids are inserted or updated into the feature bank via the exponential moving average:
\begin{equation}\label{eq:bank_update}
    \tilde{\bc}'_{i}=(1-\lambda)\,\tilde{\bc}_{i}+\lambda\,\mathbf{c}_{i},~~ i\in\hat{\mathcal{Y}}_{\mathcal{B}^{t}},
\end{equation}
while leaving classes that absent from $\mathcal{B}^{t}$ unchanged.
}

\rpami{
\textbf{Feature Redundancy and Inequity Regularizer.}
Given the feature matrix $\bC^t$ derived from Eqn.~(\ref{eq:features_from_bank}), we can then perform regularization under small test batches (even a single sample) or imbalanced test data, \ie, the wild settings. Specifically, we calculate the feature redundancy $R(\bC^t)$ and inequity $I(\bC^t)$ from matrix $\bC^t$ as test-time regularizers, following Eqns.~(\ref{eq:initial_redundancy} \& \ref{eq:initial_inequity}).

During TTA, we minimize $I(\bC^t)$ to promote unbiased representations, alleviating the potential risk of collapse induced by entropy minimization. 
We also reduce $R(\bC^t)$ to encourage non-redundant representations of test samples for enhancing discrimination ability, which improves generalization against distribution shifts even when used alone during TTA, as shown in Appendix~\ref{sec:standalone_redundancy}. 
In \mysarE, these regularizers are also combined with our sharpness-aware optimization mechanism for stable updates. Following Eqn.~(\ref{eq:hat_epsilon}), let $\hat{\boldsymbol{\epsilon}}_{R}(\bC^t;\Theta)$ and $\hat{\boldsymbol{\epsilon}}_{I}(\bC^t;\Theta)$ be the worst‑case perturbations, we approximate the gradients of the sharpness-aware feature redundancy $R^{SA}(\bC^t;\Theta)$ and inequity $I^{SA}(\bC^t;\Theta)$ \text{regularizer as:}
}
\begin{equation}\label{eq:sharpness_redundancy}
    \left.\nabla_{\Theta} R^{SA}(\bC^t;\Theta) \approx \nabla_{\Theta} R^{SA}(\bC^t;\Theta)\right|_{\Theta+\hat{\boldsymbol{\epsilon}}_{R}(\bC^t;\Theta)},
\end{equation}
\begin{equation}\label{eq:sharpness_inequity}
    \left.\nabla_{\Theta} I^{SA}(\bC^t;\Theta) \approx \nabla_{\Theta} I^{SA}(\bC^t;\Theta)\right|_{\Theta+\hat{\boldsymbol{\epsilon}}_{I}(\bC^t;\Theta)}.
\end{equation}

\rpami{
\textbf{Discussion.} 
It is worth noting that inequity $I (\cdot)$ operates at the \textit{prototype feature centroid} by maximizing the entropy of the centroid prediction, penalizing biased model representation towards a specific class. 
In contrast, redundancy $R(\cdot)$ is \textit{centroid‑invariant} and explicitly targets inter‑dimension correlations: it is computed on mean‑centered features (via centroid subtraction) to penalize statistical dependencies among feature dimensions, thereby encouraging a decorrelated representation independent of the centroid.
Together, they exert complementary effects: inequity ensures well-conditioned centroids by penalizing class bias, while redundancy enforces decorrelated features through centroid-invariant regularization.
As shown in Table~\ref{tab:ablation_fr&fi}, each regularizer individually improves TTA efficacy under wild test settings, and their combination 
yields a significant further boost.
}

\textbf{Overall Optimization.} In summary, our sharpness-aware reliable entropy minimization with feature regularization (\textit{termed} \mysarE) is given by:
\begin{equation}\label{eq:sar_extend_optimization}
    \min _{\tilde{\Theta}} S(\bx) E^{SA}(\bx; \Theta) + \alpha R^{SA}(\bC^t) + \beta I^{SA}(\bC^t), 
\end{equation}
where $S(\bx)$ and $E^{SA}(\bx;\Theta)$ are defined in Eqns.~(\ref{eq:reliable_entropy}) and (\ref{eq:sa_entropy}) respectively, \rpami{$\alpha$ and $\beta$ are balancing coefficients}, and $\tilde{\Theta}\subset\Theta$ denote learnable parameters during TTA.
\rpami{In addition, to prevent unstable optimization during the warm-up phase of the feature bank, we perform Eqn.~(\ref{eq:sar_extend_optimization}) only when $\bC^t$ contains no less than $\zeta$ non-empty class features.}
We summarize our methods in Algorithm~\ref{alg:sar-algorithm}.

\begin{figure*}[t]
\centering
\includegraphics[width=1.\linewidth]{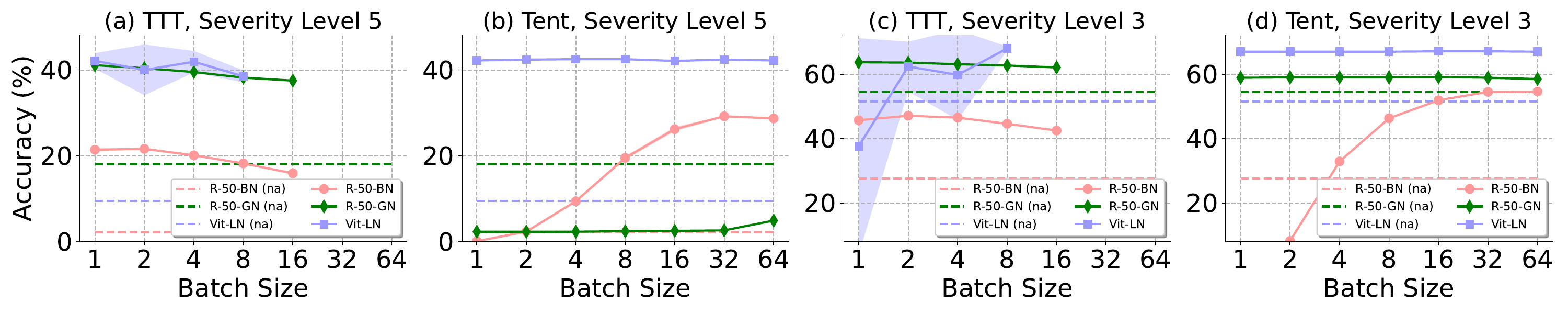}
\vspace{-0.25in}
\caption{Batch size effects of different TTA methods under different models (different normalization layers). Experiments are conducted on ImageNet-C of Gaussian noise. We report mean and standard deviation of 3 runs with different random seeds. `na' denotes no adapt accuracy. Note that except for Vit-LN, the standard deviation is too small to display in the figures.}
\label{fig:bs_effects}
\vspace{-0.1in}
\end{figure*}

\begin{figure*}[t]
\centering
\includegraphics[width=1.\linewidth]{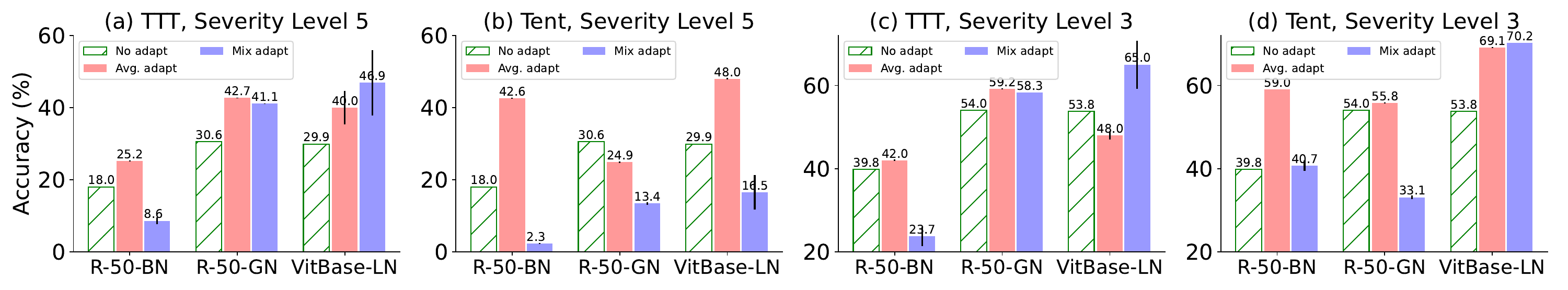}
\vspace{-0.3in}
\caption{Performance of TTA methods on different models (different norm layers) under the mixture of 15 different corruption types (ImageNet-C). We report mean\&stdev. over 3 independent runs. }
\label{fig:mix_domain_effects}
\end{figure*}

\begin{figure*}[t]
\centering
\includegraphics[width=1.\linewidth]{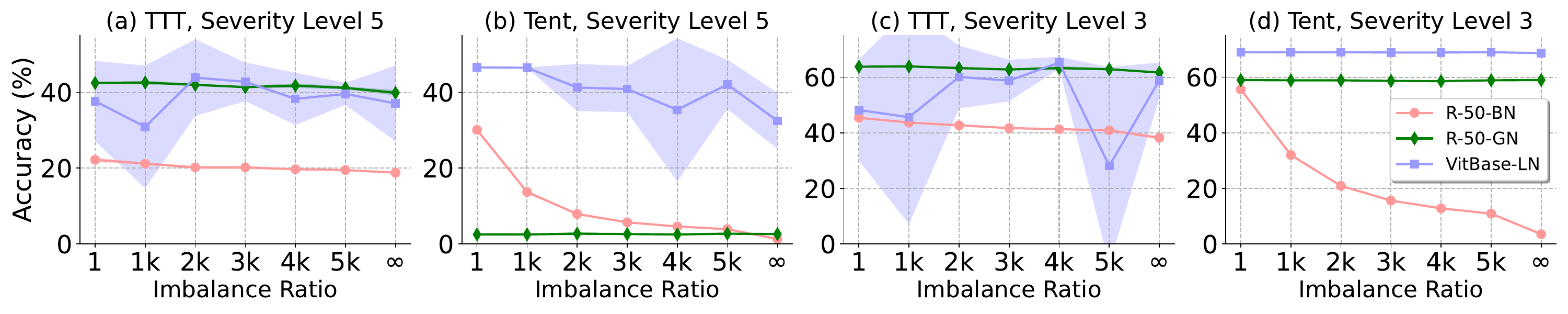}
\vspace{-0.3in}
\caption{Performance of TTA methods with different models (different norm layers) under online imbalanced label distribution shifts on ImageNet-C (Gaussian noise). We report mean\&stdev. results of 3 runs.
Note that except for VitBase-LN, the stdev. is too small to display in the figures.
}
\label{fig:label_shift_effects}
\vspace{-0.1in}
\end{figure*}

\section{Empirical Studies of Normalization Layer Effects in TTA}\label{sec:empirical_norm_effects}

This section designs experiments to illustrate how test-time adaptation (TTA) performs on models with different norm layers (including BN, GN and LN) under wild test settings described in Figure~\ref{fig:3_weak_points_tta}. We verify two representative methods introduced in Section~\ref{sec:preliminary}, \ie, self-supervised \textbf{TTT}~\cite{sun2020test} and unsupervised \textbf{Tent} (a fully TTA method)~\cite{wang2021tent}. Considering that the norm layers are often coupled with mainstream network architectures, we conduct adaptation on ResNet-50-BN (R-50-BN), ResNet-50-GN (R-50-GN) and VitBase-LN (Vit-LN). All adopted model weights are public available and obtained from \texttt{torchvision} or \texttt{timm} repository~\cite{rw2019timm}. Implementation details of experiments in this section can be found in Appendix~\ref{sec:more_exp_details}.

\noindent\textbf{(1) Norm Layer Effects in TTA Under Small Test Batch Sizes.}\label{sec:bs_effects}
We evaluate TTA methods (TTT and Tent) with different batch sizes (BS), selected from \{1, 2, 4, 8, 16, 32, 64\}. Due to GPU memory limits, we only report results of BS up to 8 or 16 for TTT (in original TTT BS is 1), since TTT needs to augment each test sample multiple times (for which we set to 20 by following \cite{niu2022EATA}).

From Figure~\ref{fig:bs_effects}, we have: \textbf{i)} For Tent, compared with R-50-BN, R-50-GN and Vit-LN are less sensitive to small test batch sizes. The adaptation performance of R-50-BN degrades severely when the batch size goes small ($<$8), while R-50-GN/Vit-LN show stable performance across various batch sizes (Vit-LN on levels 5\&3 and R-50-GN on level 3, in subfigures (b)\&(d)). It is worth noting that Tent with R-50-GN and Vit-LN not always succeeds and also has failure cases, such as R-50-GN on level 5 (Tent performs worse than no adapt), which is analyzed in Section~\ref{sec:norm_effects}.  \textbf{ii)} For TTT, all R-50-BN/GN and Vit-LN can perform well under various batch sizes. However, TTT with Vit-LN is very unstable and has a large variance over different runs, showing that TTT+VitBase is very sensitive to different sample orders. Here, TTT performs well with R-50-BN under BS 1 is mainly benefited from TTT applying multiple data augmentations to a single sample to form a mini-batch.

\noindent\textbf{(2) Norm Layer Effects in TTA Under Mixed Distribution Shifts.}\label{sec:mix_domain_effects}
We evaluate TTA methods on models with different norm layers when test data come from multiple shifted domains simultaneously. We compare `no adapt', `avg. adapt' (average accuracy of adapting on each domain separately) and `mix adapt' (adapting on mixed and shifted domains) accuracy on ImageNet-C consisting of 15 corruption types. The larger accuracy gap between `mix adapt' and 'avg. adapt' indicates the more sensitive to mixed domain shifts.

From Figure~\ref{fig:mix_domain_effects} we have: \textbf{i)} For both Tent and TTT, R-50-GN and Vit-LN perform more stable than R-50-BN under mix domain shifts. Specifically, the mix adapt accuracy of R-50-BN is consistently poor than the average adapt accuracy across different severity levels (in all subfigures (a-d)). In contrast, R-50-GN and Vit-LN are able to achieve comparable accuracy of mix and average adapt, \ie, TTT on R-50-GN (levels 5\&3) and Tent on Vit-LN (level 3). \textbf{ii)} For R-50-GN and Vit-LN, TTT performs more stable than Tent. To be specific, Tent gets 3/4 failure cases (R-50-GN on levels 5\&3, Vit-LN on level 5), which is more than that of TTT. \textbf{iii)} The same as Section~\ref{sec:bs_effects} (1), TTT on Vit-LN has large variances over multiple runs, showing TTT+Vit-LN is sensitive to different sample orders.

\noindent\textbf{(3) Norm Layer Effects in TTA Under Online Imbalanced Label Shifts.}\label{sec:label_shift_effects}
As in Figure~\ref{fig:3_weak_points_tta} (c), during the online adaptation process, the label distribution $Q_t(y)$ at different time-steps $t$ may be different (online shift) and imbalanced. To evaluate this, we first simulate this imbalanced label distribution shift by adjusting the order of input samples (from a test set) as follows. 

\textit{Online Imbalanced Label Distribution Shift Simulation.} Assuming that we have totally $T$ time-steps and $T$ equals to the class number $C$. We set the probability vector $Q_t(y)=[q_1,q_2,...,q_C]$, where $q_c=q_{max}$ if $c=t$ and $q_{c}=q_{min}\triangleq (1-q_{max})/(C-1)$ if $c\neq t$. Here, $q_{max}/q_{min}$ denotes the imbalance ratio. Then, at each $t\in\{1,2,...,T\small{=}C\}$, we sample $M$ images from the test set according to $Q_t(y)$. Based on ImageNet-C (Gaussian noise), we construct a new testing set that has online imbalanced label distribution shifts with totally $100 (M)\times 1000 (T)$ images. Note that we pre-shuffle the class orders in ImageNet-C, since we cannot know which class will come in practice.

From Figure~\ref{fig:label_shift_effects}, we have: \textbf{i)} For Tent, R-50-GN and Vit-LN are less sensitive than R-50-BN to online imbalanced label distribution shifts (see subfigures (b)\&(d)). Specifically, the adaptation accuracy of R-50-BN (levels 5\&3) degrades severely as the imbalance ratio increases. In contrast, R-50-GN and Vit-LN have the potential to perform stably under various imbalance ratios (\eg, R-50-GN and Vit-LN on level 3). \textbf{ii)} For TTT, all R-50-BN/GN and Vit-LN perform relatively stable under label shifts, except for TTT+Vit-LN has large variances. The adaptation accuracy will also degrade but not very severe as the imbalance ratio increases. \textbf{iii)} Tent with GN is more sensitive to the extent of distribution shift than BN. Specifically, for imbalanced ratio 1 (all $Q_t(y)$ are uniform) and severity level 5, Tent+R-50-GN fails and performs poorer than no adapt, while Tent+R-50-BN works well.

\noindent\textbf{(4) Overall Observations.} Based on all the above results, we have: 
\textbf{i)} R-50-GN and Vit-LN are more stable than R-50-BN when performing TTA under wild test settings (see Figure~\ref{fig:3_weak_points_tta}). However, they do not always succeed and still suffer from several failure cases. \textbf{ii)} R-50-GN is more suitable for self-supervised TTT than Vit-LN, since TTT+Vit-LN is sensitive to different sample orders and has large variances over different runs. \textbf{iii)} Vit-LN is more suitable for unsupervised Tent than R-50-GN, since Tent+R-50-GN is easily to collapse, especially when the distribution shift is severe.

\begin{table*}[t]
    \caption{Comparisons with state-of-the-art methods on ImageNet-C (severity level 5) under \textbf{\textsc{online imbalanced label shifts}} (imbalance ratio = $\infty$)  regarding \textbf{Accuracy (\%)}.
    ``BN"/``GN"/``LN" is short for Batch/Group/Layer normalization. The \textbf{bold} number indicates the best result.
    }
    \vspace{-0.15in}
    \label{tab:imagenet-c-label-shift-infity}
\newcommand{\tabincell}[2]{\begin{tabular}{@{}#1@{}}#2\end{tabular}}
 \begin{center}
 \begin{threeparttable}
 \LARGE
    \resizebox{1.0\linewidth}{!}{
 	\begin{tabular}{l|ccc|cccc|cccc|cccc|>{\columncolor{blue!8}}c}
 	\multicolumn{1}{c}{} & \multicolumn{3}{c}{Noise} & \multicolumn{4}{c}{Blur} & \multicolumn{4}{c}{Weather} & \multicolumn{4}{c}{Digital}  \\
 	 Model+Method & Gauss. & Shot & Impul. & Defoc. & Glass & Motion & Zoom & Snow & Frost & Fog & Brit. & Contr. & Elastic & Pixel & JPEG & Avg.  \\
    \cmidrule{1-17}
 	
        ResNet50 (BN) & 2.2  & 2.9  & 1.8  & 17.8  & 9.8  & 14.5  & 22.5  & 16.8  & 23.4  & 24.6  & 59.0  & 5.5  & 17.1  & 20.7  & 31.6  & 18.0  \\ 
        ~~$\bullet~$MEMO & 7.4  & 8.6  & 8.9  & 19.8  & 13.2  & 20.8  & 27.5  & 25.6  & 28.6  & 32.3  & 60.8  & 11.0  & 23.8  & 33.2  & 37.7  & 24.0 \\ %
        ~~$\bullet~$DDA & 32.2  & 33.1  & 32.0  & 14.6  & 16.4  & 16.6  & 24.4  & 20.0  & 25.5  & 17.2  & 52.2  & 3.2  & 35.7  & 41.8  & 45.4  & 27.2  \\

        ~~$\bullet~$Tent & 1.2  & 1.4  & 1.4  & 1.0  & 0.9  & 1.2  & 2.6  & 1.7  & 1.8  & 3.6  & 5.0  & 0.5  & 2.6  & 3.2  & 3.1  & 2.1  \\ 
        ~~$\bullet~$EATA & 0.3  & 0.3  & 0.3  & 0.2  & 0.2  & 0.5  & 0.9  & 0.8  & 0.9  & 1.8  & 3.5  & 0.2  & 0.8  & 1.2  & 0.9  & 0.9  \\ 

\cmidrule{1-17}

        ResNet50 (GN)   & 17.9  & 19.9  & 17.9  & 19.7 & 11.3  & 21.3  & 24.9  & 40.4  & 47.4  & 33.6  & 69.2  & 36.3  & 18.7  & 28.4  & 52.2  & 30.6  \\
~~$\bullet~$MEMO & 18.4  & 20.6  & 18.4  & 17.1  & 12.7  & 21.8  & 26.9  & 40.7  & 46.9  & 34.8  & 69.6  & 36.4  & 19.2  & 32.2  & 53.4  & 31.3  \\ 
~~$\bullet~$DDA    & 42.5 & 43.4 & 42.3 & 16.5 & 19.4 & 21.9 & 26.1 & 35.8 & 40.2 & 13.7 & 61.3 & 25.2 & 37.3 & 46.9 & 54.3 & 35.1 \\

~~$\bullet~$Tent       &   2.6  & 3.3  & 2.7  & 13.9  & 7.9  & 19.5  & 17.0  & 16.5  & 21.9  & 1.8  & 70.5  & 42.2  & 6.6  & 49.4  & 53.7  & 22.0  \\ 
~~$\bullet~$EATA       &  27.0  & 28.3  & 28.1  & 14.9  & 17.1  & 24.4  & 25.3  & 32.2  & 32.0  & 39.8  & 66.7  & 33.6  & 24.5  & 41.9  & 38.4  & 31.6  \\ 
~~$\bullet~$ROID & 32.3 & 35.0 & 32.5 & 15.9 & 19.8 & 27.3 & 30.9 & 48.0 & 44.6 & 52.8 & 72.2 & 46.0 & 34.2 & 46.9 & 54.6 & 39.5 \\
~~$\bullet~$DeYO & 42.5 & 44.9 & 43.8 & 22.2 & 16.3 & 41.0 & 13.2 & 52.2 & 51.5 & 39.7 & 73.4 & 52.6 & 46.9 & 59.3 & 59.3 & 43.9 \\
~~$\bullet~$ReCAP & 42.0 & 44.1 & 42.7 & 19.8 & 24.3 & 39.7 & 40.2 & 46.0 & 52.2 & 57.3 & 73.1 & 52.4 & 33.7 & 59.4 & 59.5 & 45.8 \\
\rowcolor{pink!15}~~$\bullet~$SAR (ours)  & 33.1$_{\pm1.0}$ & 36.5$_{\pm0.4}$ & 35.5$_{\pm1.1}$ & 19.2$_{\pm0.4}$ & 19.5$_{\pm1.2}$ & 33.3$_{\pm0.5}$ & 27.7$_{\pm4.0}$ & 23.9$_{\pm5.1}$ & 45.3$_{\pm0.4}$ & 50.1$_{\pm1.0}$ & 71.9$_{\pm0.1}$ & 46.7$_{\pm0.2}$ & 7.1$_{\pm1.8}$ & 52.1$_{\pm0.5}$ & 56.3$_{\pm0.1}$ & 37.2$_{\pm0.6}$ \\

\rowcolor{pink!30}~~$\bullet~$\mysarE (ours) & \textbf{43.0$_{\pm 0.3}$} & \textbf{45.8$_{\pm 0.5}$} & \textbf{43.9$_{\pm 0.5}$} & \textbf{35.5$_{\pm 0.6}$} & \textbf{36.5$_{\pm 0.5}$} & \textbf{44.5$_{\pm 0.2}$} & \textbf{48.7$_{\pm 0.2}$} & \textbf{56.1$_{\pm 0.3}$} & \textbf{55.0$_{\pm 0.2}$} & \textbf{61.0$_{\pm 0.3}$} & \textbf{73.4$_{\pm 0.1}$} & \textbf{55.3$_{\pm 0.1}$} & \textbf{54.0$_{\pm 0.2}$} & \textbf{61.9$_{\pm 0.1}$} & \textbf{61.0$_{\pm 0.2}$} & \textbf{51.7$_{\pm 0.1}$} \\
        
        \cmidrule{1-17}
        
        VitBase (LN)   & 9.4  & 6.7  & 8.3  & 29.1  & 23.4  & 34.0  & 27.0  & 15.8  & 26.3  & 47.4  & 54.7  & 43.9  & 30.5  & 44.5  & 47.6  & 29.9  \\ 
~~$\bullet~$MEMO & 21.6  & 17.4  & 20.6  & 37.1  & 29.6  & 40.6  & 34.4  & 25.0  & 34.8  & 55.2  & 65.0  & 54.9  & 37.4  & 55.5  & 57.7  & 39.1  \\  
~~$\bullet~$DDA    & 41.3 & 41.3 & 40.6 & 24.6 & 27.4 & 30.7 & 26.9 & 18.2 & 27.7 & 34.8 & 50.0 & 32.3 & 42.2 & 52.5 & 52.7 & 36.2 \\

~~$\bullet~$Tent       & 32.7  & 1.4  & 34.6  & 54.4  & 52.3  & 58.2  & 52.2  & 7.7  & 12.0  & 69.3  & 76.1  & 66.1  & 56.7  & 69.4  & 66.4  & 47.3  \\ 
~~$\bullet~$EATA       & 35.9  & 34.6  & 36.7  & 45.3  & 47.2  & 49.3  & 47.7  & 56.5  & 55.4  & 62.2  & 72.2  & 21.7  & 56.2  & 64.7  & 63.7  & 49.9  \\
~~$\bullet~$ROID & 44.9 & 44.7 & 45.3 & 48.6 & 48.9 & 54.1 & 50.5 & 59.9 & 57.4 & 67.0 & 75.2 & 32.5 & 60.3 & 67.5 & 64.6 & 54.8 \\
~~$\bullet~$DeYO & 53.5 & 36.0 & 54.6 & 57.6 & 58.7 & 63.7 & 46.2 & 67.6 & 66.0 & 73.2 & 77.9 & 66.7 & 69.0 & 73.5 & 70.3 & 62.3 \\
~~$\bullet~$ReCAP & 53.1 & 38.5 & 49.6 & 57.3 & 59.0 & \textbf{63.8} & 60.7 & 67.8 & \textbf{66.3} & 72.9 & 77.7 & 66.8 & 68.2 & 73.0 & 70.0 & 63.0 \\
\rowcolor{pink!15}~~$\bullet~$SAR (ours) & 46.5$_{\pm3.0}$ & 43.1$_{\pm7.4}$ & 48.9$_{\pm0.4}$ & 55.3$_{\pm0.1}$ & 54.3$_{\pm0.2}$ & 58.9$_{\pm0.1}$ & 54.8$_{\pm0.2}$ & 53.6$_{\pm7.1}$ & 46.2$_{\pm3.5}$ & 69.7$_{\pm0.3}$ & 76.2$_{\pm0.1}$ & 66.2$_{\pm0.3}$ & 60.9$_{\pm0.3}$ & 69.6$_{\pm0.1}$ & 66.6$_{\pm0.1}$ & 58.0$_{\pm0.5}$ \\
\rowcolor{pink!30}~~$\bullet~$\mysarE (ours) & \textbf{53.7$_{\pm 0.1}$} & \textbf{54.5$_{\pm 0.3}$} & \textbf{55.0$_{\pm 0.3}$} & \textbf{57.8$_{\pm 0.2}$} & \textbf{59.0$_{\pm 0.1}$} & 63.6$_{\pm 0.2}$ & \textbf{61.7$_{\pm 0.1}$} & \textbf{67.8$_{\pm 0.2}$} & 66.0$_{\pm 0.1}$ & \textbf{73.3$_{\pm 0.1}$} & \textbf{77.8$_{\pm 0.1}$} & \textbf{67.4$_{\pm 0.1}$} & \textbf{68.7$_{\pm 0.2}$} & \textbf{73.4$_{\pm 0.2}$} & \textbf{70.1$_{\pm 0.1}$} & \textbf{64.6$_{\pm 0.1}$} \\
	\end{tabular}
	}
	 \end{threeparttable}
	 \end{center}
\end{table*}

\section{Comparison with State-of-the-arts}\label{sec:main_exp}

\textbf{Dataset and Methods.} We conduct experiments based on ImageNet-C~\cite{hendrycks2019benchmarking}, a large-scale and widely used benchmark for out-of-distribution generalization. It contains 15 types of 4 main categories (noise, blur, weather, digital) corrupted images and each type has 5 severity levels. We compare with the following state-of-the-art methods. DDA~\cite{gao2022back} performs input adaptation at test time via a diffusion model. Tent~\cite{wang2021tent} minimizes prediction entropy for fully test-time adaptation. MEMO~\cite{zhang2021memo} minimizes marginal entropy over different augmented copies \wrt a given test sample. 
EATA~\cite{niu2022EATA}, ROID~\cite{kundu2020universal}, DeYO~\cite{lee2024deyo}, and ReCAP~\cite{hu2025beyond} are entropy-based methods with active sample selection for online fully TTA.

\noindent\textbf{Models and Implementation Details.} We conduct experiments on ResNet50-BN/GN and VitBase-LN that are obtained from \texttt{torchvision} or \texttt{timm}~\cite{rw2019timm}. 
\rpami{For both \mysar and \mysarE, we use SGD as the update rule, with a momentum of 0.9 and batch size of 64 (except for the experiments of batch size=1). Learning rates are 0.00025/0.001 in \mysar, and 0.001/0.005 in \mysarE for ResNet/Vit models, respectively. }
The threshold $E_0$ in Eqn.~(\ref{eq:reliable_entropy}) is set to 0.4$\times\ln 1000$ per EATA~\cite{niu2022EATA}. $\rho$ in Eqn.~(\ref{eq:sa_entropy}) is set by the default value 0.05 in \cite{foret2020sharpness}. \rpami{The coefficient $\alpha$ and $\beta$ in Eqn.~(\ref{eq:sar_extend_optimization}) are set to $10^3/D$ and 50, respectively, where $D$ is the feature dimension.} For trainable parameters of our methods during TTA, following Tent \cite{wang2021tent}, we \textbf{adapt the affine parameters} of group/layer normalization layers in models.  $\zeta$ in \mysarE is set to 100. More details and hyperparameters of compared methods are put in Appendix~\ref{sec:more_exp_details}.

\subsection{Robustness to Corruption under Wild Test Settings}
\textbf{Results under Online Imbalanced Label Distribution Shifts.} As illustrated in Section~\ref{sec:label_shift_effects}, as the imbalance ratio $q_{max}/q_{min}$ increases, TTA degrades more and more severe. Here, we make comparisons under the most difficult case: $q_{max}/q_{min}\small{=}\infty$, where test samples come in class order. We evaluate all methods under different corruptions via the same sample sequence for fairness.

From Table \ref{tab:imagenet-c-label-shift-infity}, \rpami{our \mysar consistently outperforms Tent over the 15 corruption types on ResNet50-GN and VitBase-LN, suggesting its effectiveness}. It is worth noting that Tent works well for many corruption types on VitBase-LN (\eg, \textit{defocus} and \textit{motion blur}) and ResNet50-GN (\eg, \textit{pixel}), while consistently failing on ResNet50-BN. This further verifies our observations in Section~\ref{sec:label_shift_effects} that entropy minimization on LN/GN models has the potential to perform well under online imbalanced label distribution shifts. Meanwhile, Tent also suffers from many failure cases, \eg, VitBase-LN on \textit{shot noise} and \textit{snow}. For these cases, our \methodname works well. \rpami{Moreover, our \mysarE further substantially improves the robustness by incorporating the test-time feature regularization scheme, \eg, increasing the average accuracy from 37.2\% (\mysar) to 51.7\%, with a +7.8\% gain compared to DeYO on ResNet50-GN.
Notably, \mysarE remains effective even in scenarios where all existing methods perform poorly (\eg, +16.3\% on \textit{defocus} with ResNet50-GN),
suggesting the distinct strength of our feature regularization design.}

\begin{table}[t]
\caption{Characteristics of state-of-the-art methods. We evaluate the efficiency of different methods with ResNet50 (GN) on ImageNet-C (Gaussian noise, severity level 5), which consists of 50,000 images. The real run time is tested via a single V100 GPU. 
DDA \cite{gao2022back} pre-trains an additional diffusion model and then performs input adaptation/diffusion at test time. 
SAR$^{2\dagger}$ is an efficient implementation variant that also selects test samples for feature regularizers, but it incurs a slight accuracy drop compared with \mysar (0.4\% in average accuracy on ImageNet-C under mixed shifts).
}
\label{tab:methods_summary_supp}
\newcommand{\tabincell}[2]{\begin{tabular}{@{}#1@{}}#2\end{tabular}}
\begin{center}
\begin{threeparttable}
\Huge
    \resizebox{1.0\linewidth}{!}{
 	\begin{tabular}{l|cc|cc}
 	 Method & \#Forward & \#Backward & Other computation & GPU time (s)\\
 	\midrule
        MEMO \cite{zhang2021memo}  & 50,000$\times$65 & 50,000$\times$64 & AugMix~\cite{hendrycks2020augmix} & 55,980  \\ %
        DDA \cite{gao2022back} & 50,000$\times$2 & 0 & 50,000 diffusion &  146,220  \\ %
        TTT \cite{sun2020test} & 50,000$\times$21 & 50,000$\times$20 & rotation aug. & 3,600  \\ %
        Tent \cite{wang2021tent} & 50,000 & 50,000 & n/a & 110  \\
        EATA \cite{niu2022EATA} & 50,000 & 26,196 & regularizer & 114  \\ %
    \midrule
        \methodname (ours) & 50,000 + 12,710 & 12,710$\times$2 & n/a & 115   \\
        \mysarE (ours) & 50,000$\times$2 & 50,000$\times$2 & feature bank & 226 \\
        SAR$^{2\dagger}$ (ours) & 50,000 + 23,198 & 23,198$\times$2 & feature bank & 163\\
	\end{tabular}
	}
\end{threeparttable}
\end{center}
\end{table}

\begin{table*}[t]
    \caption{Comparisons with state-of-the-art methods on ImageNet-C (severity level 5) with \textbf{\textsc{Batch Size=1}} regarding \textbf{Accuracy (\%)}.
    ``BN"/``GN"/``LN" is short for Batch/Group/Layer normalization. The \textbf{bold} number indicates the best result.
    }
    \vspace{-0.15in}
    \label{tab:imagenet-c-bs1}
\newcommand{\tabincell}[2]{\begin{tabular}{@{}#1@{}}#2\end{tabular}}
 \begin{center}
 \begin{threeparttable}
 \LARGE
    \resizebox{1.0\linewidth}{!}{
 	\begin{tabular}{l|ccc|cccc|cccc|cccc|>{\columncolor{blue!8}}c}
 	\multicolumn{1}{c}{} & \multicolumn{3}{c}{Noise} & \multicolumn{4}{c}{Blur} & \multicolumn{4}{c}{Weather} & \multicolumn{4}{c}{Digital}  \\
 	 Model+Method & Gauss. & Shot & Impul. & Defoc. & Glass & Motion & Zoom & Snow & Frost & Fog & Brit. & Contr. & Elastic & Pixel & JPEG & Avg.  \\
 	\midrule
        ResNet50 (BN) & 2.2  & 2.9  & 1.9  & 17.9  & 9.8  & 14.8  & 22.5  & 16.9  & 23.3  & 24.4  & 58.9  & 5.4  & 17.0  & 20.6  & 31.6  & 18.0  \\ 
        ~~$\bullet~$MEMO & 7.5  & 8.7  & 8.9  & 19.7  & 13.0  & 20.8  & 27.6  & 25.4  & 28.7  & 32.2  & 60.9  & 11.0  & 23.8  & 32.9  & 37.5  & 23.9  \\ %
        ~~$\bullet~$DDA & 32.1  & 32.8  & 31.8  & 14.7  & 16.6  & 16.6 & 24.2  & 20.0  & 25.4  & 17.2  & 52.1  & 3.2 & 35.7  & 41.5  & 45.3  & 27.3  \\

        ~~$\bullet~$Tent & 0.1  & 0.1  & 0.1  & 0.1  & 0.1  & 0.1  & 0.2  & 0.2  & 0.2  & 0.2  & 0.2  & 0.1  & 0.1  & 0.2  & 0.1  & 0.1  \\ 
        ~~$\bullet~$EATA & 0.1  & 0.1  & 0.1  & 0.1  & 0.1  & 0.1  & 0.2  & 0.2  & 0.2  & 0.1  & 0.2  & 0.1  & 0.1  & 0.2  & 0.1  & 0.1  \\

\cmidrule{1-17}

        ResNet50 (GN)  & 18.0  & 19.8  & 17.9  & 19.8  & 11.4  & 21.4  & 24.9  & 40.4  & 47.3  & 33.6  & 69.3  & 36.3  & 18.6  & 28.4  & 52.3  & 30.6 \\
~~$\bullet~$MEMO & 18.5  & 20.5  & 18.4  & 17.1  & 12.6  & 21.8  & 26.9  & 40.4  & 47.0  & 34.4  & 69.5  & 36.5  & 19.2  & 32.1  & 53.3  & 31.2  \\ 
~~$\bullet~$DDA       & 42.4 & 43.3 & 42.3 & 16.6 & 19.6 & 21.9 & 26.0 & 35.7 & 40.1 & 13.7 & 61.2 & 25.2 & 37.5 & 46.6 & 54.1 & 35.1 \\

~~$\bullet~$Tent       & 2.5  & 2.9  & 2.5  & 13.5  & 3.6  & 18.6  & 17.6  & 15.3  & 23.0  & 1.4  & 70.4  & 42.2  & 6.2  & 49.2  & 53.8  & 21.5  \\
~~$\bullet~$EATA       & 24.8  & 28.3  & 25.7  & 18.1  & 17.3  & 28.5  & 29.3  & 44.5  & 44.3  & 41.6& 70.9  & 44.6  & 27.0  & 46.8  & 55.7  & 36.5  \\
~~$\bullet~$DeYO & 41.8 & 44.7 & 43.0 & 22.5 & 24.7 & 41.8 & 24.4 & 54.5 & 52.2 & 20.7 & 73.5 & 53.5 & 48.5 & 60.2 & 59.8 & 44.4 \\
~~$\bullet~$ReCAP & 42.5 & 44.4 & 42.9 & 19.4 & 25.0 & 42.2 & 44.0 & 49.7 & 52.4 & 57.5 & 72.9 & 53.6 & 29.5 & 60.4 & 60.0 & 46.4 \\
\rowcolor{pink!15}~~$\bullet~$SAR (ours)  & 23.4$_{\pm0.3}$ & 26.6$_{\pm0.4}$ & 23.9$_{\pm0.0}$ & 18.4$_{\pm0.1}$ & 15.4$_{\pm0.3}$ & 28.6$_{\pm0.3}$ & 30.4$_{\pm0.2}$ & 44.9$_{\pm0.3}$ & 44.7$_{\pm0.2}$ & 25.7$_{\pm0.6}$ & 72.3$_{\pm0.2}$ & 44.5$_{\pm0.1}$ & 14.8$_{\pm2.7}$ & 47.0$_{\pm0.1}$ & 56.1$_{\pm0.0}$ & 34.5$_{\pm0.2}$ \\
\rowcolor{pink!30}~~$\bullet~$\mysarE (ours) & \textbf{46.2$_{\pm 0.2}$} & \textbf{48.3$_{\pm 0.1}$} & \textbf{46.7$_{\pm 0.1}$} & \textbf{37.4$_{\pm 0.1}$} & \textbf{38.6$_{\pm 0.3}$} & \textbf{45.9$_{\pm 0.3}$} & \textbf{50.9$_{\pm 0.1}$} & \textbf{58.9$_{\pm 0.1}$} & \textbf{56.6$_{\pm 0.1}$} & \textbf{63.4$_{\pm 0.2}$} & \textbf{73.9$_{\pm 0.0}$} & \textbf{57.2$_{\pm 0.1}$} & \textbf{57.5$_{\pm 0.1}$} & \textbf{63.7$_{\pm 0.1}$} & \textbf{62.4$_{\pm 0.1}$} & \textbf{53.8$_{\pm 0.02}$} \\
        
        \cmidrule{1-17}
        VitBase (LN)   & 9.5  & 6.7  & 8.2  & 29.0  & 23.4  & 33.9  & 27.1  & 15.9  & 26.5  & 47.2  & 54.7  & 44.1  & 30.5  & 44.5  & 47.8  & 29.9 \\
~~$\bullet~$MEMO & 21.6  & 17.3  & 20.6  & 37.1  & 29.6  & 40.4  & 34.4  & 24.9  & 34.7  & 55.1  & 64.8  & 54.9  & 37.4  & 55.4  & 57.6  & 39.1  \\
~~$\bullet~$DDA       & 41.3 & 41.1 & 40.7 & 24.4 & 27.2 & 30.6 & 26.9 & 18.3 & 27.5 & 34.6 & 50.1 & 32.4 & 42.3 & 52.2 & 52.6 & 36.1 \\

~~$\bullet~$Tent       & 42.2  & 1.0  & 43.3 & 52.4  & 48.2  & 55.5  & 50.5  & 16.5  & 16.9  & 66.4  & 74.9  & 64.7  & 51.6  & 67.0  & 64.3  & 47.7  \\
~~$\bullet~$EATA       & 29.7  & 25.1  & 34.6  & 44.7  & 39.2  & 48.3  & 42.4  & 37.5  & 45.9  & 60.0  & 65.9  & 61.2  & 46.4  & 58.2  & 59.6  & 46.6  \\
~~$\bullet~$DeYO & 54.0 & 52.1 & 55.1 & 58.8 & 59.5 & 64.2 & 53.5 & 68.2 & 66.4 & 73.7 & 78.3 & 68.2 & 68.9 & 73.8 & 70.8 & 64.4 \\
~~$\bullet~$ReCAP & 53.5 & 56.7 & 56.9 & 59.2 & 60.5 & 65.3 & 64.0 & 69.6 & 67.2 & 74.1 & 78.4 & 64.6 & 70.2 & 74.4 & 71.5 & 65.7 \\
\rowcolor{pink!15}~~$\bullet~$SAR (ours)  & 40.8$_{\pm0.4}$ & 36.4$_{\pm0.7}$ & 41.5$_{\pm0.3}$ & 53.7$_{\pm0.2}$ & 50.7$_{\pm0.1}$ & 57.5$_{\pm0.1}$ & 52.8$_{\pm0.3}$ & 59.1$_{\pm0.4}$ & 50.7$_{\pm0.6}$ & 68.1$_{\pm1.4}$ & 74.6$_{\pm0.7}$ & 65.7$_{\pm0.0}$ & 57.9$_{\pm0.1}$ & 68.9$_{\pm0.1}$ & 65.9$_{\pm0.0}$ & 56.3$_{\pm0.1}$ \\

\rowcolor{pink!30}~~$\bullet~$\mysarE (ours) & \textbf{57.2$_{\pm 0.0}$} & \textbf{58.0$_{\pm 0.2}$} & \textbf{58.1$_{\pm 0.2}$} & \textbf{61.0$_{\pm 0.1}$} & \textbf{62.0$_{\pm 0.1}$} & \textbf{66.6$_{\pm 0.1}$} & \textbf{65.8$_{\pm 0.1}$} & \textbf{70.4$_{\pm 0.1}$} & \textbf{68.8$_{\pm 0.1}$} & \textbf{75.2$_{\pm 0.1}$} & \textbf{79.1$_{\pm 0.1}$} & \textbf{69.5$_{\pm 0.1}$} & \textbf{71.4$_{\pm 0.2}$} & \textbf{75.2$_{\pm 0.1}$} & \textbf{72.0$_{\pm 0.1}$} & \textbf{67.4$_{\pm 0.04}$} \\
	\end{tabular}
	}
	 \end{threeparttable}
	 \end{center}
\end{table*}

\begin{table}[t]
    \caption{Comparisons with state-of-the-arts  on ImageNet-C under \textbf{\textsc{mixture of 15 corruption types}} regarding \textbf{Accuracy (\%)}.
    }
    \setlength{\tabcolsep}{5pt}
    \vspace{-0.15in}
    \label{tab:imagenet-c-mix-only}
\newcommand{\tabincell}[2]{\begin{tabular}{@{}#1@{}}#2\end{tabular}}
 \begin{center}
 \begin{threeparttable}
    \resizebox{1\linewidth}{!}{
 	\begin{tabular}{l|cc|cc|>{\columncolor{blue!8}}c}
     	\multicolumn{1}{c}{} & \multicolumn{2}{c}{ResNet50 (GN)} & \multicolumn{2}{c}{VitBase (LN)} & \multicolumn{1}{c}{}  \\
 	 Method  & Level 5 & Level 3 & Level 5 & Level 3 & Avg.\\
        \midrule
        NoAdapt & 30.6 & 54.0 & 29.9 & 53.8 & 42.1 \\
        MEMO~\cite{zhang2021memo} & 31.2 & 54.5 & 39.1 & 62.1 & 46.7 \\
        DDA~\cite{gao2022back} & 35.1 & 52.3 & 36.1 & 53.2 & 44.2 \\
        Tent~\cite{wang2021tent} & 13.4 & 33.1 & 16.5 & 70.2 & 33.3 \\
        EATA~\cite{niu2022EATA} & 38.1 & 56.1 & 55.7 & 69.6 & 54.9 \\
        ROID~\cite{marsden2024universal} & 36.9 & 56.9 & 52.9 & 68.2 & 53.7 \\
        DeYO~\cite{lee2024deyo} & 38.6 & 59.2 & 59.4 & 72.1 &  57.3 \\
        ReCAP~\cite{hu2025beyond} & 41.5 & 59.8 & 59.4 & 72.1 & 58.2\\ 
        \rowcolor{pink!15} \mysar (ours) & 38.3$_{\pm0.1}$ & 57.4$_{\pm0.1}$ & 57.1$_{\pm0.1}$   & 70.7$_{\pm0.01}$ & 55.9$_{\pm0.01}$ \\
        \rowcolor{pink!30} \mysarE (ours) & \textbf{43.3$_{\pm0.1}$} & \textbf{59.8$_{\pm0.01}$} & \textbf{61.2$_{\pm0.03}$}   & \textbf{72.9$_{\pm0.01}$} & \textbf{59.3$_{\pm0.01}$} \\

	\end{tabular}
	}
	 \end{threeparttable}
	 \end{center}
\end{table}

\begin{table*}[t]
    \caption{Comparisons with state-of-the-art methods on ImageNet-C (severity level 5) under \textbf{\textsc{continuous adaptation}} with online imbalanced label shifts (imbalance ratio = $\infty$)  regarding \textbf{Accuracy (\%)}.
    ``GN"/``LN" is short for Group/Layer normalization. The \textbf{bold} number indicates the best result.
    }
    \vspace{-0.15in}
    \label{tab:continuous_label_shifts}
\newcommand{\tabincell}[2]{\begin{tabular}{@{}#1@{}}#2\end{tabular}}
 \begin{center}
 \begin{threeparttable}
 \LARGE
    \resizebox{1.0\linewidth}{!}{
 	\begin{tabular}{l|ccc|cccc|cccc|cccc|>{\columncolor{blue!8}}c}
 	\multicolumn{1}{c}{} & \multicolumn{3}{c}{Noise} & \multicolumn{4}{c}{Blur} & \multicolumn{4}{c}{Weather} & \multicolumn{4}{c}{Digital}  \\
 	 Model+Method & Gauss. & Shot & Impul. & Defoc. & Glass & Motion & Zoom & Snow & Frost & Fog & Brit. & Contr. & Elastic & Pixel & JPEG & Avg.  \\
    \cmidrule{1-17}
 	
        ResNet50 (GN)   & 17.9  & 19.9  & 17.9  & 19.7 & 11.3  & 21.3  & 24.9  & 40.4  & 47.4  & 33.6  & 69.2  & 36.3  & 18.7  & 28.4  & 52.2  & 30.6  \\
~~$\bullet~$MEMO & 18.4  & 20.6  & 18.4  & 17.1  & 12.7  & 21.8  & 26.9  & 40.7  & 46.9  & 34.8  & 69.6  & 36.4  & 19.2  & 32.2  & 53.4  & 31.3  \\ 
~~$\bullet~$DDA    & 42.5 & 43.4 & 42.3 & 16.5 & 19.4 & 21.9 & 26.1 & 35.8 & 40.2 & 13.7 & 61.3 & 25.2 & 37.3 & 46.9 & 54.3 & 35.1 \\

~~$\bullet~$Tent       & 2.6 & 0.2 & 0.1 & 1.4 & 0.1 & 0.1 & 0.2 & 0.3 & 0.1 & 0.2 & 0.3 & 0.1 & 0.1 & 0.1 & 0.3 & 0.4  \\ 
~~$\bullet~$EATA       & 25.8 & 23.7 & 20.2 & 14.4 & 13.7 & 18.4 & 20.2 & 20.8 & 23.2 & 32.0 & 41.4 & 20.5 & 22.4 & 23.1 & 26.5 & 23.1  \\ 
~~$\bullet~$ROID & 32.6 & 35.2 & 33.1 & 15.7 & 19.3 & 27.1 & 31.3 & 48.0 & 45.1 & 52.4 & \textbf{72.0} & 45.9 & 35.1 & 46.7 & 54.2 & 39.6 \\
~~$\bullet~$DeYO        & 41.9 & 47.8 & 46.1 & 7.4 & 0.4 & 0.2 & 0.1 & 0.1 & 0.1 & 0.1 & 0.1 & 0.1 & 0.1 & 0.1 & 0.1 & 9.7 \\
~~$\bullet~$ReCAP & 42.2 & 47.8 & 46.6 & 5.8 & 0.2 & 0.1 & 0.1 & 0.2 & 0.2 & 0.1 & 4.3 & \textbf{53.6} & 15.4 & 44.8 & 58.3 & 21.3 \\
\rowcolor{pink!15}~~$\bullet~$SAR (ours) & 33.7$_{\pm1.0}$ & 46.1$_{\pm0.2}$ & 46.6$_{\pm0.1}$ & 10.9$_{\pm1.5}$ & 11.9$_{\pm3.4}$ & 14.1$_{\pm7.6}$ & 3.0$_{\pm2.1}$ & 25.6$_{\pm7.1}$ & 17.8$_{\pm8.9}$ & 38.7$_{\pm19.2}$ & 70.6$_{\pm1.3}$ & 48.6$_{\pm0.5}$ & 5.7$_{\pm0.9}$ & 0.9$_{\pm0.2}$ & 36.7$_{\pm3.4}$ & 27.4$_{\pm1.1}$ \\

\rowcolor{pink!30}~~$\bullet~$\mysarE (ours) & \textbf{43.2$_{\pm0.5}$} & \textbf{48.8$_{\pm0.2}$} & \textbf{47.8$_{\pm0.2}$} & \textbf{29.8$_{\pm0.4}$} & \textbf{35.2$_{\pm0.2}$} & \textbf{41.1$_{\pm0.4}$} & \textbf{46.2$_{\pm0.4}$} & \textbf{50.4$_{\pm0.3}$} & \textbf{53.1$_{\pm0.2}$} & \textbf{58.6$_{\pm0.2}$} & 69.1$_{\pm0.1}$ & 50.9$_{\pm0.2}$ & \textbf{48.8$_{\pm0.4}$} & \textbf{57.3$_{\pm0.1}$} & \textbf{59.1$_{\pm0.1}$} & \textbf{49.3$_{\pm0.1}$} \\
        
        \cmidrule{1-17}
        
        VitBase (LN)   & 9.4  & 6.7  & 8.3  & 29.1  & 23.4  & 34.0  & 27.0  & 15.8  & 26.3  & 47.4  & 54.7  & 43.9  & 30.5  & 44.5  & 47.6  & 29.9  \\ 
~~$\bullet~$MEMO & 21.6  & 17.4  & 20.6  & 37.1  & 29.6  & 40.6  & 34.4  & 25.0  & 34.8  & 55.2  & 65.0  & 54.9  & 37.4  & 55.5  & 57.7  & 39.1  \\  
~~$\bullet~$DDA    & 41.3 & 41.3 & 40.6 & 24.6 & 27.4 & 30.7 & 26.9 & 18.2 & 27.7 & 34.8 & 50.0 & 32.3 & 42.2 & 52.5 & 52.7 & 36.2 \\

~~$\bullet~$Tent       & 32.5 & 0.3 & 0.1 & 0.4 & 0.1 & 0.1 & 0.1 & 0.1 & 0.1 & 0.1 & 0.2 & 0.1 & 0.1 & 0.1 & 0.1 & 2.3 \\ 
~~$\bullet~$EATA      & 36.1 & 27.7 & 22.4 & 14.7 & 20.0 & 13.4 & 5.7 & 8.2 & 11.4 & 6.6 & 24.2 & 0.3 & 10.2 & 15.6 & 19.2 & 15.7 \\
~~$\bullet~$ROID & 45.1 & 45.3 & 45.4 & 48.0 & 50.4 & 54.8 & 51.3 & 60.1 & 57.5 & 67.1 & 75.1 & 60.0 & 59.4 & 67.7 & 64.7 & 56.8 \\
~~$\bullet~$DeYO        & 52.9 & 56.9 & 57.1 & 35.5 & 37.6 & 40.0 & 18.0 & 39.1 & 39.7 & 45.1 & 50.6 & 39.3 & 42.7 & 47.1 & 45.6  & 43.2  \\
~~$\bullet~$ReCAP  & 53.3 & 56.7 & 57.2 & 53.8 & 57.1 & 60.9 & 39.1 & 61.1 & 62.8 & 69.3 & 76.3 & 62.3 & \textbf{67.1} & 71.4 & 68.7 & 61.1 \\
\rowcolor{pink!15}~~$\bullet~$SAR (ours) & 47.2$_{\pm1.9}$ & 54.0$_{\pm0.2}$ & 56.2$_{\pm0.2}$ & 53.0$_{\pm0.3}$ & 55.8$_{\pm0.2}$ & 59.0$_{\pm0.2}$ & 56.2$_{\pm0.6}$ & 61.1$_{\pm0.2}$ & 63.1$_{\pm0.2}$ & 68.8$_{\pm0.2}$ & 76.4$_{\pm0.1}$ & 63.7$_{\pm0.1}$ & 60.8$_{\pm1.9}$ & 69.7$_{\pm0.3}$ & 67.8$_{\pm0.1}$ & 60.9$_{\pm0.2}$ \\
\rowcolor{pink!30}~~$\bullet~$\mysarE (ours) & \textbf{53.7$_{\pm0.1}$} & \textbf{57.6$_{\pm0.3}$} & \textbf{58.2$_{\pm0.1}$} & \textbf{54.6$_{\pm0.7}$} & \textbf{58.4$_{\pm0.2}$} & \textbf{61.4$_{\pm0.2}$} & \textbf{60.7$_{\pm0.4}$} & \textbf{64.0$_{\pm0.2}$} & \textbf{64.3$_{\pm0.2}$} & \textbf{70.4$_{\pm0.1}$} & \textbf{76.7$_{\pm0.1}$} & \textbf{63.6$_{\pm0.2}$} & 66.6$_{\pm0.4}$ & \textbf{71.8$_{\pm0.3}$} & \textbf{69.1$_{\pm0.1}$} & \textbf{63.4$_{\pm0.1}$} \\
	\end{tabular}
	}
	 \end{threeparttable}
	 \end{center}
\end{table*}

\noindent\textbf{Results under Batch Size = 1.} 
From Table \ref{tab:imagenet-c-bs1}, \rpami{our \mysar mitigates the collapse issue in Tent and achieves robust performance gain in many cases, suggesting its effectiveness. It is worth noting that MEMO/DDA does not suffer from the instability issue and achieves the same results across different wild testing scenarios, by resetting model parameters when adapting to each sample. However, they require significantly more computational overhead (see Table~\ref{tab:methods_summary_supp}) and yield only a limited performance gain as they cannot leverage knowledge from previously seen samples. In contrast, our \mysar enables robust and efficient online adaptation through reliable and sharpness-aware updates. 
Furthermore, our \mysarE unleashes the effectiveness and stability of TTA with the test-time feature regularization design,
\eg, increasing the average accuracy from 56.3\% (\mysar) to 67.4\% and surpassing ReCAP by 1.7\% on Vitbase-LN, consistently outperforming all baselines. Notably, \mysarE also dramatically reduces performance variance (\ie, by up to \textit{10}-fold on ResNet50-GN compared to \mysar) and remains effective even in scenarios where all existing methods underperform.
These results highlight the crucial role of test-time feature regularization in online TTA, enabling \mysarE to achieve both substantially higher accuracy and enhanced stability across diverse wild test settings.
}

\noindent\textbf{Results under Mixed Distribution Shifts.} 
We evaluate different methods on the mixture of 15 corruption types (total of 15$\times$50,000 images) at different severity levels (5\&3). From Table \ref{tab:imagenet-c-mix-only}, our \mysar \rpami{~achieves robust performance across models and severity levels}, suggesting its effectiveness. Tent fails (occurs collapse) on ResNet50-GN levels 5\&3 and VitBase-LN level 5 and achieves inferior accuracy than the no-adapt model, showing the instability of long-range online entropy minimization. Compared with Tent, although MEMO and DDA achieve better results, they rely on much more computation (inefficient at test time) as in Table \ref{tab:methods_summary_supp}, 
\rpami{and DDA also need to train an additional diffusion model. 
DeYO and ReCAP mitigate the collapse issue in Tent by selectively adapting to individual samples. However, they use solely output signals for sample selection, which remains insufficient and achieves a performance plateau on VitBase-LN (\eg, 59.3\%-59.4\% at level 5). In \mysarE, by integrating test-time feature regularization, we break through the DeYO/ReCAP performance ceiling, improving over \mysar by 5.0\% on ResNet50-GN and 4.1\% on VitBase-LN at level 5, consistently outperforming all baselines. These results suggest that regularizing the model representation at test time provides distinct robustness and effectiveness for TTA.
}

\rpami{
\noindent\textbf{Results under Continuous Adaptation with Online Imbalanced Label Shifts.}
We further evaluate the long-term stability of TTA methods under a more challenging yet practical test scenario in the wild, where the model is continuously adapted across domains under online imbalanced label shifts, without parameter resets. 
From Table~\ref{tab:continuous_label_shifts}, our \mysar improves upon Tent by a large margin, suggesting our effectiveness, but it still suffers from instability during prolonged TTA.
Prior methods such as DeYO and ReCAP also exhibit severe degradation over time, \eg, collapsing with accuracy close to zero on \textit{blurs} and \textit{weathers} with ResNet50-GN. 
This indicates that while selecting pseudo labels for entropy minimization helps stabilize early TTA, model biases such as feature inequity may still accumulate and amplify during the long-term TTA process, hindering TTA stability. In contrast, by explicitly regulating feature redundancy and inequity, our \mysarE prevents collapse and remains highly effective throughout continuous TTA, \eg, increasing the average accuracy from 27.4\% (\mysar) to 49.3\% with up to \textit{10}-fold reduction in performance variance, markedly outperforming all baselines. 
These further underscore the importance of our feature regularization design in enabling robust TTA across both short-term and long-term adaptation in the wild.
}

\subsection{Ablate Experiments}\label{sec:ablate_results_main}

\begin{table}[t]
  \centering
  \setlength{\tabcolsep}{4pt}
  \caption{Effects of components in \mysar. We report \textbf{
  Accuracy (\%)} on ImageNet-C (level 5) with Vitbase-LN in the wild. ``reliable (RE)" and ``sharpness (SA)" denote Eqns.~(\ref{eq:reliable_entropy} \& \ref{eq:sa_entropy}), “recover (RC)” denotes model recovery scheme.
  }
  \label{tab:ablation_sar}
\vspace{-0.15in}
\begin{center}
 \begin{threeparttable}
    \resizebox{1\linewidth}{!}{
        \begin{tabular}{l|ccc|ccc|>{\columncolor{blue!8}}c}
          Exp. & ~RE & SA & RC~ & ~Label Shifts & Mix Shifts & $BS = 1$~ & ~Avg.~ \\
        \midrule
        Tent & ~ & ~ &  & 47.3 & 16.5 & 47.7 & 37.2 \\
        1 & \checkmark & ~ &  & 53.1 & 55.2 & 50.5 & 52.9 \\
        2 & \checkmark & \checkmark & & 54.5 & 57.1 & 55.7 & 55.8 \\
        \rowcolor{pink!30} 3 (\mysar)~ & \checkmark & \checkmark & \checkmark & \textbf{58.0} & \textbf{57.1} & \textbf{56.3} & \textbf{57.1} \\
        \end{tabular}
        }
         \end{threeparttable}
         \end{center}
\end{table}

\begin{table}[t]
  \centering
  \caption{Effects of components in \mysarE. We report the \textbf{Accuracy (\%)} on ImageNet-C (level 5) with VitBase-LN under wild test scenarios. Feature redundancy $R(\cdot)$ and inequity $I(\cdot)$ are calculated based on feature bank.}
  \label{tab:ablation_fr&fi}
\vspace{-0.15in}
\begin{center}
 \begin{threeparttable}
    \resizebox{1\linewidth}{!}{
        \begin{tabular}{l|cc|ccc|>{\columncolor{blue!8}}c}
          Exp. & $R(\cdot)$ & $I(\cdot)$ & Label Shifts & Mix Shifts & $BS = 1$ & Avg. \\
        \midrule     
        SAR & ~ & ~ & 58.0 & 57.1 & 56.3 & 57.1 \\
        1 & \checkmark & ~ & 63.8 & 60.3 & 64.2 & 62.8 \\
        2 & ~ & \checkmark & 63.3 & 58.9 & 61.1 & 61.1 \\
        \rowcolor{pink!30}3 (SAR$^2)$ & \checkmark & \checkmark & \textbf{64.6} & \textbf{61.2} & \textbf{67.4} & \textbf{64.4} \\
        \end{tabular}
        }
         \end{threeparttable}
         \end{center}
\end{table}

\rpami{
\noindent\textbf{Effects of Components in \methodname.}
From Table~\ref{tab:ablation_sar}, compared with plain entropy minimization, the reliable entropy in Eqn. (\ref{eq:reliable_entropy}) clearly improves the TTA performance by removing partial noisy updates, \eg, $37.2\%\rightarrow 52.9\%$ w.r.t. the average accuracy over wild test settings. Meanwhile, sharpness-aware (sa) minimization in Eqn.~(\ref{eq:sa_entropy}) promotes a flat entropy minima in the parameter space, which further enhances the robustness against large and noisy updates, \eg, yielding a $+5.2\%$ improvement under TTA with a single sample. 
With both reliable entropy and sharpness optimization, our \mysar performs stably except for very few cases, \eg, on \textit{snow} under label shifts (\textit{see} Table~\ref{tab:ablation_ls_infty} in Appendix for more details). For this case, our model recovery scheme takes effect, \ie, $55.8\%\rightarrow 57.1\%$ regarding the average accuracy under wild test scenarios.
}

\rpami{\noindent\textbf{Effects of Components in \mysarE.}
As discussed in Section~\ref{sec:overall_sar_method}, inequity $I(\cdot)$ operates at the centroid level to prevent the representation bias towards a specific class. In contrast, redundancy $R(\cdot)$ is centroid-invariant, measuring inter-dimension feature correlations independent of the centroid. They thus focus complementarily on the centroid and centroid-invariant perspectives to maintain both balanced/unbiased centroids and decorrelated features during TTA. As shown in Table~\ref{tab:ablation_fr&fi}, each regularizer independently improves the efficacy of TTA in the wild, \eg, 57.1\% (\mysar) \textit{vs.} 62.8\% (\mysar + redundancy) \textit{vs.} 61.1\% (\mysar + inequity) regarding average accuracy over wild scenarios. By combining both regularizers, the efficacy and stability of TTA are further improved by a large margin, \eg, achieving an accuracy of 67.4\% under TTA with a single sample. 
These results highlight both the effectiveness and the complementarity of our regularizer designs.
}

\begin{figure}[t]
\vspace{-0.02in}
\centering
\includegraphics[width=1.0\linewidth]{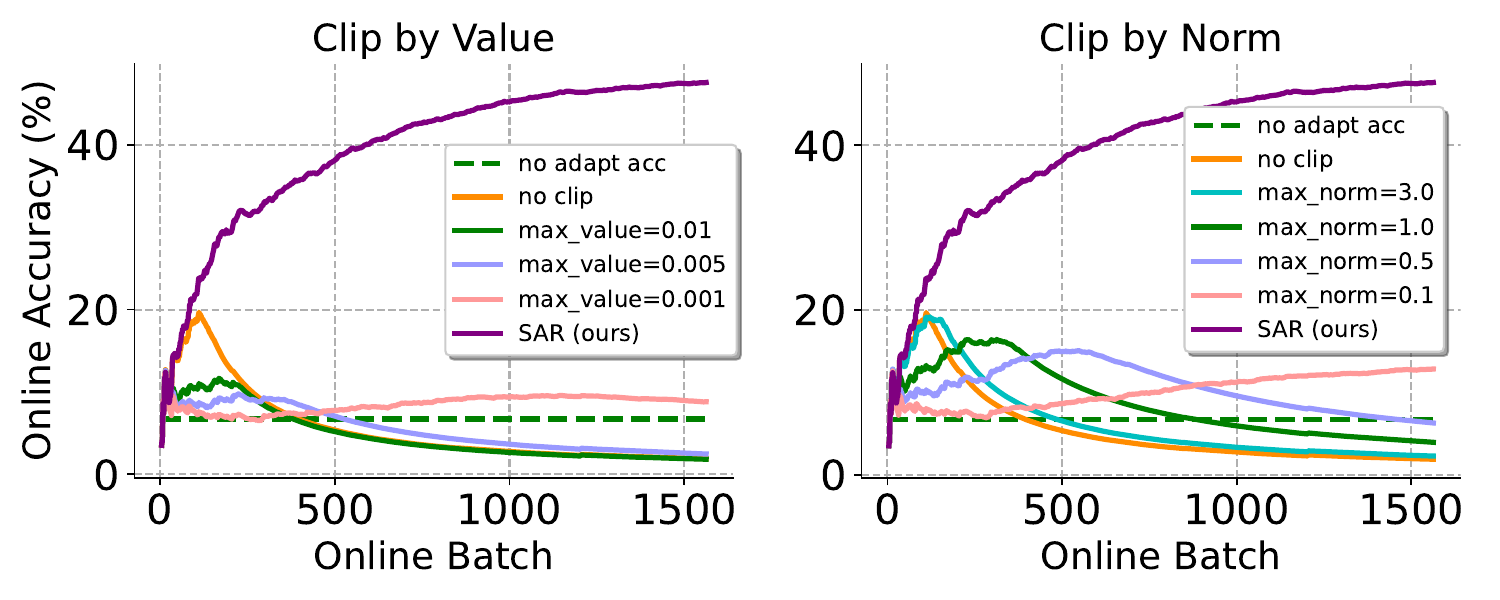}
\vspace{-0.3in}
\caption{Comparison with gradient clipping. Results on VitBase-LN, ImageNet-C, shot noise, severity level 5, online imbalanced (ratio = $\infty$) label shift. \textit{Accuracy is calculated over all previous test samples.}}
\label{fig:ablation_grad_clip}
\vspace{-0.1in}
\end{figure}

\noindent\textbf{Comparison with Gradient Clipping.}\label{sec:compare_grad_clip}
As mentioned in Section \ref{sec:sar_method}, gradient clipping is a straightforward solution to alleviate model collapse. Here, we compare our \methodname with two variants of gradient clip, \ie, by value and by norm. From Figure~\ref{fig:ablation_grad_clip}, for both two variants, it is hard to set a proper threshold $\delta$ for clipping, since the gradients for different models and test data have different scales and thus the $\delta$ selection would be sensitive. We carefully select $\delta$ on a specific test set (\textit{shot noise} level 5). Then, we select a very small $\delta$ to make gradient clip work, \ie, clip by value 0.001 and by norm 0.1. Nonetheless, the performance gain over ``no adapt" is  very marginal, since the small $\delta$ would limit the learning ability of the model and in this case the clipped gradients may point in a very different direction from the true gradients.
However, a large $\delta$ fails to stabilize the adaptation process and the accuracy will degrade after the model collapses (\eg, clip by value 0.005 and by norm 1.0). In contrast, \methodname does not need to tune such a parameter and achieves significant improvements than gradient clipping.

\begin{figure*}[ht!]
    \setlength{\subfigcapskip}{-0.05in}     
    \centering
    \subfigure[VitBase-LN, NoAdapt]{\label{fig:consistency_entropy_comparison}\includegraphics[width=44mm]{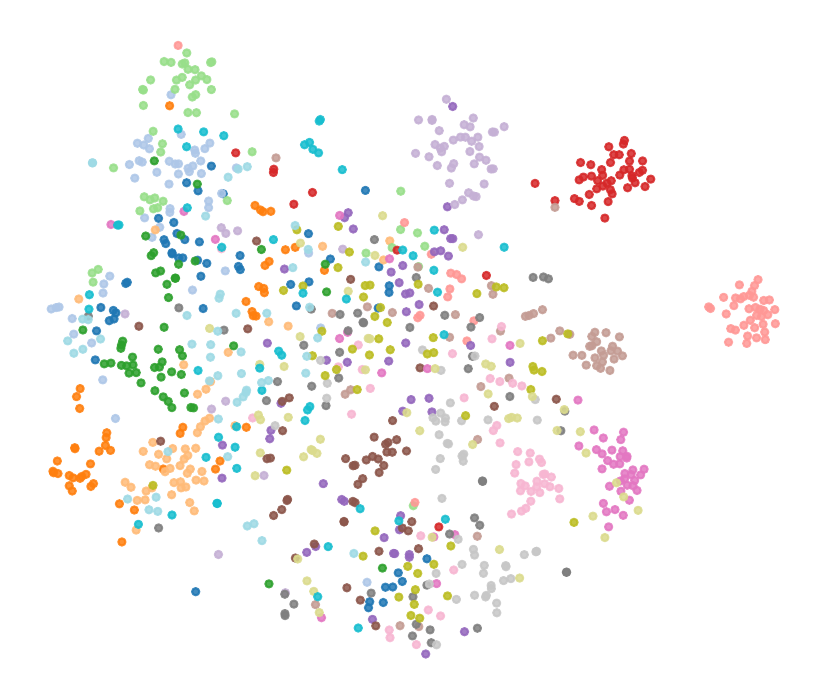}}
    \hfill
    \subfigure[Tent~\cite{wang2021tent}]{\label{fig:consistent_heatmap}\includegraphics[width=44mm]{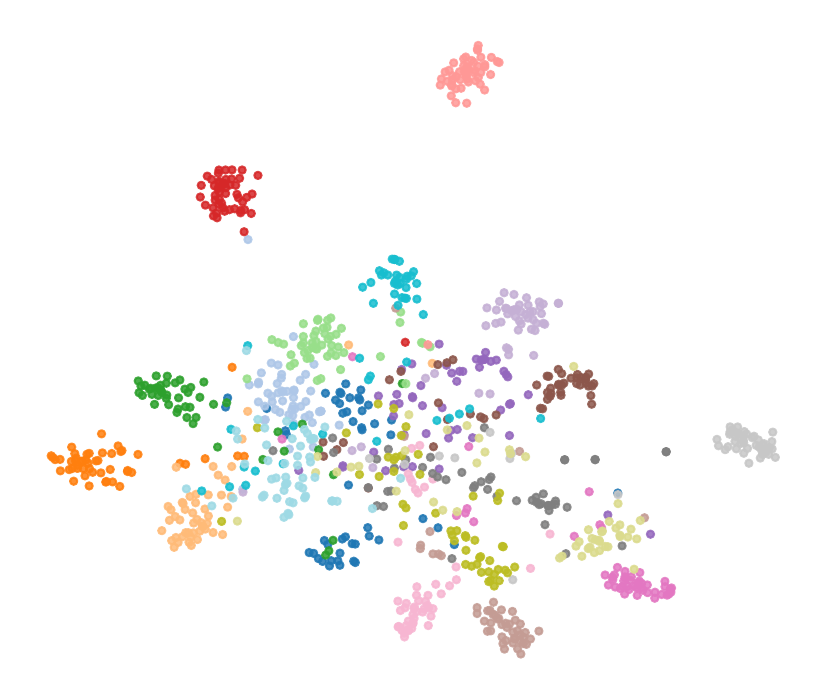}}
    \hfill
    \subfigure[\mysar (ours)]{\label{fig:sar_tsne}\includegraphics[width=44mm]{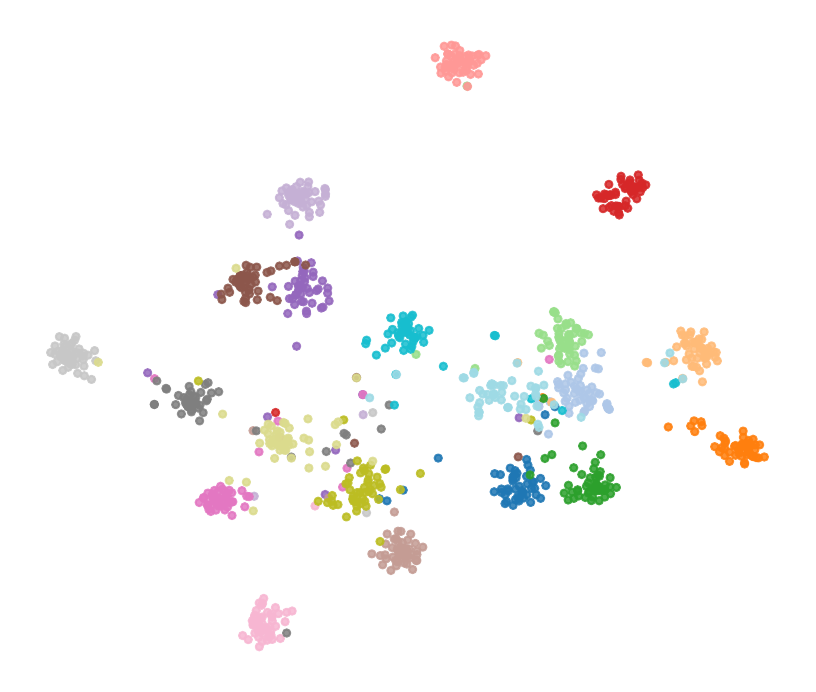}}
    \hfill
    \subfigure[\mysarE (ours)]{\label{fig:sar2_tsne}\includegraphics[width=44mm]{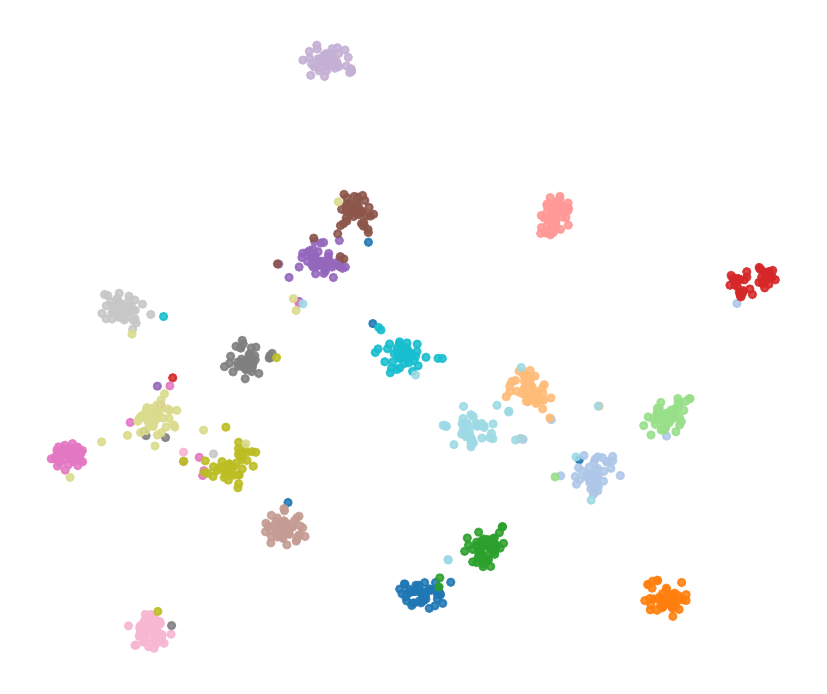}}
    \vspace{-0.02in}
    \caption{t-SNE visualization of the feature space after adapting VitBase-LN on ImageNet-C (Snow, level 5) under online imbalanced label shifts. 
    }
    \label{fig:tsne-visualization}
\end{figure*}

\begin{table}[t]
    \caption{
    Effects of batch-imbalance penalties. Results are reported on ImageNet -C (level 5) under a mild scenario w.r.t. \textbf{Accuracy (\%, $\uparrow$)} and \textbf{ECE (\%, $\downarrow$)}.
    }
    \vspace{-0.15in}
    \label{tab:inequity_information_maximization}
\newcommand{\tabincell}[2]{\begin{tabular}{@{}#1@{}}#2\end{tabular}}
 \begin{center}
 \begin{threeparttable}
    \resizebox{1\linewidth}{!}{
    \begin{tabular}{l|cc|cc|>{\columncolor{blue!8}}c>{\columncolor{blue!8}}c}
     	\multicolumn{1}{c}{} & \multicolumn{2}{c}{ResNet50 (GN)} & \multicolumn{2}{c}{VitBase (LN)} & \multicolumn{2}{>{\columncolor{blue!8}}c}{Avg.}  \\
     	 Method  & Acc. & ECE &   Acc. & ECE & Acc. & ECE\\
        \midrule
        NoAdapt & 30.6 & 8.4 & 29.9 & 7.0 & 30.3 & 7.7 \\
        Tent~\cite{wang2021tent} & 25.1 & 48.7 & 47.9 & 22.3 & 36.5 & 35.5 \\
        \ding{59} InfoMax~\cite{liang2020we} & 38.2 & 20.2 & 53.8 & 8.2 & 46.0 & 14.2 \\
        \rowcolor{pink!30} \ding{59} Our $I(\cdot)$ (Eqn.~\ref{eq:initial_inequity}) & 48.9 & 7.5 & 59.3 & 4.9 & \textbf{54.1} & \textbf{6.2} \\
    
	\end{tabular}
	}
	 \end{threeparttable}
	 \end{center}

\end{table}

\rpami{
\noindent\textbf{Advantages of Inequity over Information Maximization~\cite{liang2020we}.}
We further compare our inequity regularizer $I(\cdot)$ with the diversity term in information maximization (IM) loss~\cite{hu2017learning}, which both penalize biased representation/prediction towards a specific class, while our $I(\cdot)$ operates at the representation's centroid level and IM loss operates at the output level. From Table~\ref{tab:inequity_information_maximization}, although both inequity $I(\cdot)$ and IM loss alleviate the collapse issue in Tent, our inequity consistently outperforms IM loss regarding both accuracy and calibration across architectures, \eg, increasing the accuracy from 46.0\% to 54.1\% and reducing ECE from 14.2\% to 6.2\%. 
This is because IM loss averages the softmax output of batch samples for entropy estimation, while $I(\cdot)$ first computes the centroid of batch features before calculating its prediction entropy. Under a limited batch size $B$, IM loss constrains the per-class total probability to $B/C$ to achieve a uniform batch-average prediction, implying an implicit confidence flattening effect to TTA.
In contrast, our inequity loss $I(\cdot)$ only ensures the representation's centroid is debiased, while each sample may retain a confident prediction. Detailed results across 15 domains (Table~\ref{tab:detailed_inequity_infomax} in Appendix) further highlight that IM loss can decrease Tent’s accuracy on domains where Tent itself does not collapse, whereas our inequity $I(\cdot)$ consistently improves Tent's accuracy across all domains, suggesting our effectiveness for online TTA.
}

\begin{table}[t]
    \caption{Impacts of initial representation health on TTA effectiveness. Results are reported on ImageNet-C (Gaussian, level 5) under label shifts regarding \textbf{Accuracy (\%)}. We perform $K$ steps of feature regularization (Eqns.~\ref{eq:initial_redundancy}-\ref{eq:initial_inequity}) on a random mini-batch to boost initial representation health before TTA. 
    }
    \vspace{-0.15in}
    \label{tab:health_impact}
\newcommand{\tabincell}[2]{\begin{tabular}{@{}#1@{}}#2\end{tabular}}
 \begin{center}
 \begin{threeparttable}
    \resizebox{1\linewidth}{!}{
    \begin{tabular}{l|ccc|>{\columncolor{blue!8}}c}
 	 Model  & Tent & SAR & DeYO  & Avg.\\
        \midrule
        ResNet50 (GN) & 2.6 & 33.7 & 41.1 & 25.8 \\
        \ding{59} feature regularizer ($K=2$) & 5.1 & 35.0 & 42.2 & 27.4$_{(~+1.6)}$ \\
        \ding{59} feature regularizer ($K=4$) & 16.3 & 36.9 & 42.8 & 32.0$_{(~+6.2)}$ \\
        \rowcolor{pink!30} \ding{59} feature regularizer ($K=6$) & 33.5 & 38.4 & 43.1 & \textbf{38.4$_{(+12.6)}$} \\
	\end{tabular}
	}
	 \end{threeparttable}
	 \end{center}
\end{table}

\begin{figure}[ht!]
\centering
\vspace{-0.02in}
\includegraphics[width=1.0\linewidth]{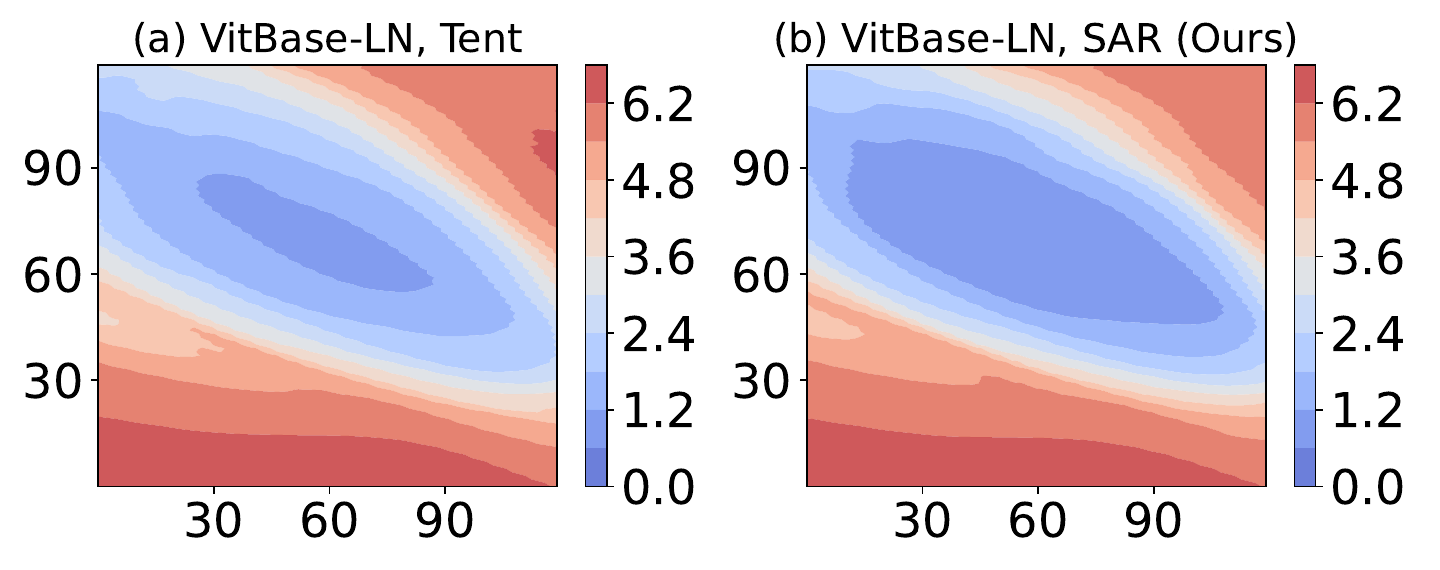}
\vspace{-0.3in}
\caption{
\rpami{Visualization of entropy-loss surface for VitBase-LN adapted via Tent and our \mysar on ImageNet-C, Gaussian noise, severity level 5.}
}
\label{fig:loss_landscape}
\vspace{-0.05in}
\end{figure}

\rpami{\noindent\textbf{Impacts of Initial Representation Health on TTA.} As shown in Figure~\ref{fig:feature_motivation}, a more severe distribution shift induces larger redundancy and inequity in model representation before TTA takes place. To further examine how this initial representation health impacts TTA efficacy, 
we conduct analysis in Table~\ref{tab:health_impact} by applying feature regularization (Eqns.~\ref{eq:initial_redundancy}-\ref{eq:initial_inequity}) to $K$ random mini-batches before we start TTA.
From the results, online TTA significantly benefits from better initial representation, \eg, 25.8\%$\rightarrow$38.4\% with feature regularizer ($K=6$) \wrt average accuracy. Notably, poor initial health, \eg, high inequity, can mislead online adaptation and worsen during TTA (Figure~\ref{fig:feature_motivation}), causing methods like Tent to collapse on the original ResNet50-GN. In contrast, by simply enhancing initial representation, Tent avoids collapse and achieves an accuracy of 33.5\%. These highlight the crucial role of well-conditioned initial representations for stable TTA, and also the potential of our test-time feature regularization as a plug-in to existing methods.
}

\rpami{
\noindent\textbf{Separability in Feature Space.} 
We compare the separability in feature space based on t-SNE, where each point is colored by its ground-truth label, and 20 classes are selected for visualization.
Here, Tent uses a smaller learning rate to mitigate model collapse.
From Figure~\ref{fig:tsne-visualization}, we have the following observations: 1)~Without adaptation, features are scattered and poorly separated; 2)~Tent slightly improves separability but also produces a large and overly dense cluster. This coincides with the increase in feature inequity (Figure~\ref{fig:feature_motivation}), where representations from multiple classes gradually collapse toward a dominant one, leading to higher representation similarity; 
3)~\mysar further improves separability compared to Tent, but several clusters remain highly similar;
4)~\mysarE incorporates the redundancy and inequity regularizers, which explicitly regulate representation overlap across classes and balance the spread of features, resulting in well-separated, distributed, and more compact clusters, thereby providing a more stable feature space for TTA.

}

\noindent\textbf{Sharpness of Loss Surface.} 
We visualize the loss surface by adding perturbations to model weights, as done in \cite{li2018visualizing}. We plot Figure \ref{fig:loss_landscape} via the model weights obtained after the adaptation on the whole test set. By comparing Figures \ref{fig:loss_landscape} (a) and (b), the area (the deepest blue) within the lowest loss contour line of our \methodname is larger than Tent, suggesting that our solution achieves flatter entropy minima and thus is more robust to noisy/large gradients.

\section{Conclusions}
In this paper, we seek to stabilize online test-time adaptation (TTA) under wild test settings, \ie, mix shifts, small batch, and imbalanced label shifts. To this end, we first analyze and conduct extensive empirical studies to verify why wild TTA fails. Then, we point out that batch norm acts as a crucial obstacle to stable TTA. Meanwhile, though batch-agnostic norms (\ie, group and layer norm) perform more stably under wild settings, they still suffer from many failure cases. We then investigate these failures by examining model gradients and feature behaviors before and after collapse during TTA. Our findings indicate that collapse occurrence is consistently characterized by a sharp gradient explosion followed by severe degradation, accompanied by substantial deterioration of feature representations. To address these, we first propose a sharpness-aware and reliable entropy minimization method (\methodname) by suppressing the effect of certain noisy test samples with large gradients. Furthermore, we introduce two feature regularizers to boost \mysar, supported by a prototype feature bank, to facilitate regularization in online wild test settings, namely \mysarE. Specifically, the redundancy regularizer reduces inter-dimensional correlations, while the inequity regularizer mitigates feature bias toward specific classes, thereby preventing representation deterioration. Extensive experimental results demonstrate the stability and efficiency of our \methodname and \mysarE under wild test settings.

\bibliographystyle{IEEEtran}
{
\bibliography{main}
}

\newpage

\appendices
\onecolumn

\begin{LARGE}
~~~\vspace{1pt}
\begin{center}
    \bf Supplementary Materials for ``\mytitle''
\end{center}
\end{LARGE}
\vspace{2pt}

\rpami{In the supplementary, we provide more implementation details, more experimental results, more ablative results, and additional discussions of our \mysar and \mysarE.
We organize our supplementary as follows.
}

\begin{itemize}[leftmargin=*]
    \item In Section~\ref{sec:more_implementation_details}, we provide more implementation details of our proposed \mysar and \mysarE, and the compared baselines.
    \item In Section~\ref{sec:more_experimental_results}, we supplement more experimental results to further demonstrate our superiority for TTA under wild test settings.
    \item In Section~\ref{sec:more_ablation}, we include extra ablation studies on component efficacy and demonstrate our robustness against hyperparameter choices.
    \item In Section~\ref{sec:more_discussions}, we provide extended observations and discussions to further validate our design choices in \mysar and \mysarE. 
\end{itemize}


\section{More Details of \mysar and \mysarE}\label{sec:more_implementation_details}

\subsection{More Details on Datasets}\label{sec:more_datasets}

\begin{figure*}[h]
\renewcommand\thefigure{A}
\centering\includegraphics[width=0.6\linewidth]{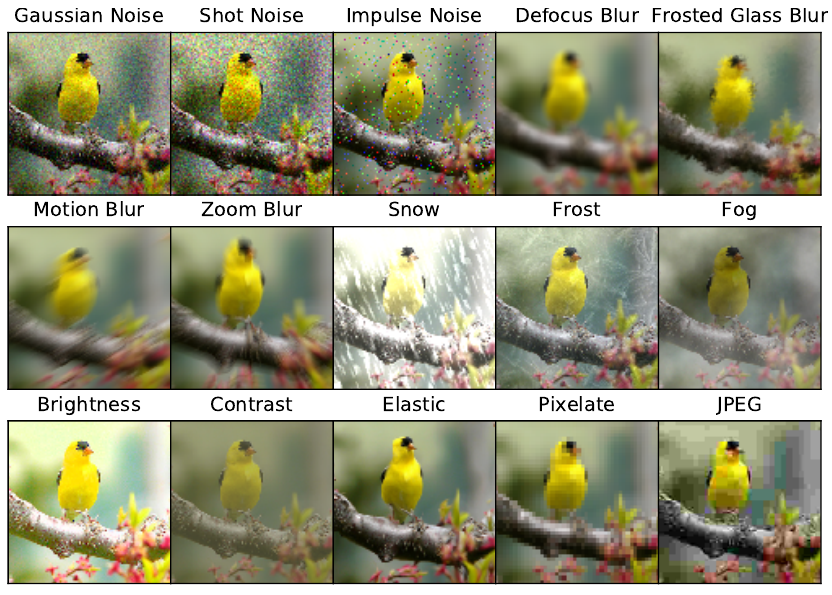}\caption{Visualizations of different corruption types in ImageNet corruption benchmark, which are taken from the original paper of ImageNet-C~\cite{hendrycks2019benchmarking}.}
\label{fig:corruption_types}
\end{figure*}

In this paper, we mainly evaluate the out-of-distribution generalization ability of all methods on a large-scale and widely used benchmark, namely \textbf{ImageNet-C}\footnote{\url{https://zenodo.org/record/2235448\#.YzQpq-xBxcA}}~\cite{hendrycks2019benchmarking}. ImageNet-C is constructed by corrupting the original ImageNet~\cite{deng2009imagenet} test set. The corruption (as shown in Figure~\ref{fig:corruption_types}) consists of 15 different types, \ie, Gaussian noise, shot noise, impulse noise, defocus blur, glass blur, motion blur, zoom blur, snow, frost, fog, brightness, contrast, elastic transformation, pixelation, and JPEG compression, where each corruption type has 5 severity levels and the larger severity level means more severe distribution shift. 
Then, we further conduct experiments on \textbf{ImageNet-R}~\cite{hendrycks2021many} to verify the effectiveness of our method. ImageNet-R contains 30,000 images with various artistic renditions of 200 ImageNet classes, primarily collected from Flickr and filtered by Amazon MTurk annotators.

\subsection{More Experimental Protocols on Evaluation}\label{sec:more_exp_details}

All pre-trained models involved in our paper for test-time adaptation are publicly available,
including ResNet50-BN\footnote{\url{https://download.pytorch.org/models/resnet50-19c8e357.pth}} obtained from \texttt{torchvision} library, ResNet-50-GN\footnote{\url{https://github.com/rwightman/pytorch-image-models/releases/download/v0.1-rsb-weights/resnet50\_gn\_a1h2-8fe6c4d0.pth}} and VitBase-LN\footnote{\url{https://storage.googleapis.com/vit\_models/augreg/B\_16-i21k-300ep-lr\_0.001-aug\_medium1-wd\_0.1-do\_0.0-sd\_0.0--imagenet2012-steps\_20k-lr\_0.01-res\_224.npz}} and ConvNeXt-LN\footnote{\url{https://dl.fbaipublicfiles.com/convnext/convnext_base_1k_224_ema.pth}} obtained from \texttt{timm} repository~\cite{rw2019timm}. We summarize the detailed characteristics of all involved methods in Table~\ref{tab:methods_summary_supp} and introduce their implementation details in the following. 

\noindent\textbf{Ours \mysar and \mysarE.} We use SGD as the update rule, with a momentum of 0.9, batch size of 64 (except for the experiments of batch size = 1), with learning rates of 0.00025/0.001 in \mysar, and 0.001/0.005 in \mysarE for ResNet/Vit models, respectively. The learning rate for batch size = 1 is scaled down to (0.00025/16) for ResNet and (0.001/32) for Vit in \mysar, and (0.001/16) for ResNet and (0.005/16) for Vit in \mysarE. The threshold $E_0$ in Eqn.~(\ref{eq:reliable_entropy}) is set to 0.4$\times\ln 1000$ per EATA~\cite{niu2022EATA}. $\rho$ in Eqn.~(\ref{eq:sa_entropy}) is set by the default value 0.05 in \cite{foret2020sharpness}. In Eqn.~(\ref{eq:sar_extend_optimization}), $\alpha$ and $\beta$ are set to $10^3$ and 50, respectively, and moving average factor $\lambda$ in Eqn.~(\ref{eq:bank_update}) is set to 0.9, $\zeta$ is set to 100. \rpami{For ResNet50-GN, since it tends to produce constant zero-valued dimensions (ResNet50-GN applies both group normalization and ReLU function on feature maps, see Section~\ref{sec:resnet-gn-sparsity}), we instead apply mean-centering on the feature dimension for Eqn.~(\ref{eq:initial_redundancy}), while keeping Eqn.~(\ref{eq:initial_redundancy}) unchanged for other models.} For model recovery, we record the entropy loss values with a moving average factor of 0.9 for $e_m$, and the reset threshold $e_0$ is set to 0.2. For \textbf{learnable parameters}, we only update affine parameters in normalization layers by following Tent~\cite{wang2021tent}. However, since the top/deep layers are more sensitive and more important to the original model than shallow layers as mentioned in~\cite{choi2022improving}, we freeze the top layers and update the affine parameters of layer or group normalization in the remaining shallow layers. Specifically, for ResNet50-GN that has 4 layer groups (layer1, 2, 3, 4), we freeze the layer4. For ViTBase-LN that has 11 blocks groups (blocks1-11), we freeze blocks9, blocks10, blocks11. \rpami{DeYO and ReCAP also adopt the same parameters for updating at test time.}
\vspace{0.05in}

\noindent\textbf{TTT\footnote{\url{https://github.com/yueatsprograms/ttt_imagenet_release}}~\cite{sun2020test}.} 
For fair comparisons, we seek to compare all methods based on the same model weights. However, TTT alters the model training process and requires the model contains a self-supervised rotation prediction branch for test-time training. Therefore, we modify TTT so that it can be applied to any pre-trained model. Specifically, given a pre-trained model, we add a new branch (random initialized) from the end of a middle layer (2nd layer group of ResNet-50-GN and 6th blocks group of VitBase-LN) for the rotation prediction task. We first freeze all original parameters of the pre-trained model and train the newly added branch for 10 epochs on the original ImageNet training set. Here, we apply an SGD optimizer, with a momentum of 0.9, an initial learning rate of 0.1/0.005 for ResNet50-GN/VitBase-LN, and
decrease it at epochs 4 and 7 by decreasing factor 0.1.  Then, we take the newly obtained model (with two branches) as the base model to perform test-time training. During test time, we use SGD as the update rule with a learning rate of 0.001 for ResNet0-GN (following TTT) and 0.0001 for VitBase-LN, and the data augmentation size is set to 20 (following~\cite{niu2022EATA}).
\vspace{0.05in}

\noindent\textbf{Tent\footnote{\url{https://github.com/DequanWang/tent}}~\cite{wang2021tent}.} We follow all hyper-parameters that are set in Tent unless it does not provide. Specifically, we use SGD as the update rule, with a momentum of 0.9, batch size of 64 (except for the
experiments of batch size = 1 and effects of small test batch sizes (in Section~\ref{sec:bs_effects})), and learning rate of 0.00025/0.001 for ResNet/Vit models. The learning rate for batch size = 1 is set to (0.00025/32) for ResNet models and (0.001/64) for Vit models. The trainable parameters are all affine parameters in normalization layers.
\vspace{0.05in}

\noindent\textbf{EATA\footnote{\url{https://github.com/mr-eggplant/EATA}}~\cite{niu2022EATA}.} We follow all hyper-parameters that are set in EATA unless it does not provide. Specifically, the entropy constant $E_0$ (for reliable sample identification) is set to $0.4\times\ln 1000$. The $\epsilon$ for redundant sample identification is set to 0.05. The trade-off parameter $\beta$ for entropy loss and regularization loss is set to 2,000. The number of pre-collected in-distribution test samples for Fisher importance calculation is 2,000. The update rule is SGD, with a momentum of 0.9, batch size of 64 (except for the
experiments of batch size = 1), and learning rate of 0.00025/0.001 for ResNet/Vit models. The learning rate for batch size = 1 is set to (0.00025/32) for ResNet models and (0.001/64) for Vit models. The trainable parameters are all affine parameters in normalization layers. 
\vspace{0.05in}

\noindent\textbf{DeYO\footnote{\url{https://github.com/Jhyun17/DeYO}}~\cite{lee2024deyo}.} We follow all hyper-parameters that are set in DeYO unless it does not provide. Specifically, the entropy constant $E_0$ (for reliable sample identification) is set to $0.4\times\ln 1000$, and the factor $\tau_{\text{Ent}}$ is set to 0.5$\times\ln{1000}$. The Pseudo-Label Probability Difference (PLPD) threshold $\tau_{\text{PLPD}}$ is set to 0.2. The update rule is SGD, with a momentum of 0.9, batch size of 64 (except for the
experiments of batch size = 1), and learning rate of 0.00025/0.001 for ResNet/Vit models. The learning rate for batch size = 1 is set to (0.00025/16) for ResNet models and (0.001/32) for Vit models. The trainable parameters are the affine parameters of normalization layers from layer 1 to layer 3 in ResNet models, and the affine parameters of the layer normalization layers from blocks 1 to blocks 8 in Vit models. 
\vspace{0.05in}

\noindent\textbf{ReCAP\footnote{\url{https://github.com/hzcar/ReCAP}}~\cite{hu2025beyond}.} We follow all hyper-parameters that are set in ReCAP unless it does not provide. Specifically, the reset threshold $e_0$ is set to 0.2. The balancing coefficient $\lambda$ is set to 0.5. The range $\tau$ of the local scope is set to 1.2, and the Pseudo-Label Probability Difference (PLPD) threshold $\tau_{\text{PLPD}}$ is set to 0.2. For ResNet50-GN, the sample selection threshold $L_0$ and the update scaling threshold $\tau_{RE}$ are set to $0.8\times\ln 1000$, and the maximum scaling of each update is clipped by 3.0. For Vitbase-LN, the sample selection threshold $L_0$ and the update scaling threshold $\tau_{RE}$ are set to $1.0\times\ln 1000$, and the maximum scaling of each update is clipped by 1.5. The update rule is SGD, with a momentum of 0.9, batch size of 64 (except for the
experiments of batch size = 1), and learning rate of 0.00025/0.001 for ResNet/Vit models. When batch size = 1, the learning rate for set to (0.00025/16) for ResNet and (0.001/32) for Vit, while the maximum scaling value of update is clipped by 5.0. The trainable parameters are the affine parameters of normalization layers from layer 1 to layer 3 in ResNet models, and the affine parameters of the layer normalization layers from blocks 1 to blocks 8 in Vit models. 
\vspace{0.05in}

\noindent\textbf{ROID\footnote{\url{https://github.com/mariodoebler/test-time-adaptation}}~\cite{marsden2024universal}.} We follow all hyper-parameters that are set in ROID unless it does not provide. Specifically, the weight ensembling $\alpha$ is set to 0.99. The temperature $\tau$ is set to $\frac{1}{3}$. We do not use the prior correction strategy, as we empirically find that it would significantly harm TTA performance under wild test settings. The update rule is SGD, with a momentum of 0.9, batch size of 64, and learning rate of 0.00025/0.001 for ResNet/Vit models. Trainable parameters are the affine parameters of all normalization layers.
\vspace{0.05in}

\noindent\textbf{MEMO\footnote{\url{https://github.com/zhangmarvin/memo}}~\cite{zhang2021memo}.} We follow all hyper-parameters that are set in MEMO. Specifically, we use the AugMix~\cite{hendrycks2020augmix} as a set of data augmentations and the augmentation size is set to 64. For Vit models, the optimizer is AdamW~\cite{loshchilov2019decoupled}, with learning rate 0.00001 and weight decay 0.01. For ResNet models, the optimizer is SGD, with learning rate 0.00025 and no weight decay. The trainable parameters are the entire model.

\noindent\textbf{DDA\footnote{\url{https://github.com/shiyegao/DDA}}~\cite{gao2022back}.} We reproduce DDA according to its official GitHub repository and use the default hyper-parameters.
\vspace{0.05in}

\noindent\textbf{More Details on Experiments in Section~\ref{sec:empirical_norm_effects}: Normalization Layer Effects in TTA.} In Section~\ref{sec:empirical_norm_effects}, we investigate the effects of TTT and Tent with models that have different norm layers under \{small test batch sizes, mixed distribution shifts, online imbalanced label distribution shifts\}. For each experiment, we only consider one of the three above test settings. To be specific, for experiments regarding batch size effects (Section~\ref{sec:bs_effects} (1)), we only tune the batch size and the test set does not contain multiple types of distribution shifts and its label distribution is always uniform. For experiments of mixed domain shifts (Section~\ref{sec:mix_domain_effects} (2)), the test samples come from the mixture of 15 corruption types, while the batch size is 64 for Tent and 1 for TTT, and the label distribution of test data is always uniform. For experiments of online label shifts (Section~\ref{sec:label_shift_effects} (3)), the label distribution of test data is online shifted and imbalanced, while the BS is 64 for Tent and 1 for TTT, and test data only consist of one corruption type. 
Moreover, it is worth noting that we re-scale the learning rate for entropy minimization (Tent) according to the batch size, since entropy minimization is sensitive to the learning rate and a fixed learning rate often fails to work well. Specifically, the learning rate is re-scaled as $(0.00025/32)\times \text{BS} \textsc{~~if~~} \text{BS} < 32 \textsc{~~else~~} 0.00025$ for ResNet models and $(0.001/64)\times \text{BS}$ for Vit models. Compared with Tent, the single sample adaptation method TTT is not very sensitive to the learning rate, and thus we set the same learning rate for various batch sizes. We also provide the results of TTT under different batch sizes with dynamic re-scaled learning rates in Table \ref{tab:bs_effects_ttt_rescaled_lrs}. 

\begin{table*}[h]
    \renewcommand\thetable{A}
    \caption{Batch size (BS) effects in TTT \cite{sun2020test} with different models (different norm layers). The learning rate is dynamically re-scaled by $0.001\times$\textsc{BS}. We report the accuracy (\%) on ImageNet-C, Gaussian noise, with severity level 5.
    }
    \label{tab:bs_effects_ttt_rescaled_lrs}
\newcommand{\tabincell}[2]{\begin{tabular}{@{}#1@{}}#2\end{tabular}}
 \begin{center}
 \begin{threeparttable}
    \resizebox{0.55\linewidth}{!}{
 	\begin{tabular}{lccccc}
 	 Model & $BS=1$ & $BS=2$ & $BS=4$ & $BS=8$ & $BS=16$   \\
 	\midrule
    ResNet50-BN   & 21.2 & 23.4 & 23.4 & 24.7 &	24.6   \\
    ResNet50-GN  & 40.9 & 40.5 & 40.8 &	41.1 & 40.7  \\
	\end{tabular}
	}
	 \end{threeparttable}
	 \end{center}
\end{table*}

\section{More Experimental Results}\label{sec:more_experimental_results}

\subsection{Comparisons with State-of-the-arts under Online Imbalanced Label Shift on Severity Level 3}

We provide more results regarding online imbalanced label distribution shift (imbalance ratio $=\infty$) of all compared methods in Table~\ref{tab:imagenet-c-label-shift-infity-level3}. The results are consistent with that of the main paper (severity level 5), where our \mysar achieves robust performance gain in the wild, and our \mysarE performs best in the average of 15 different corruption types.
It is worth noting that DDA achieves competitive results under \textit{noise} corruptions while performing worse for other corruption types. The reason is that the diffusion model used in DDA for input adaptation is trained via noise diffusion, and thus its generalization ability to diffuse other corruptions is still limited.

\begin{table*}[h]
    \renewcommand\thetable{B}

    \caption{Comparisons with state-of-the-art methods on ImageNet-C of severity level 3 under \textbf{\textsc{online imbalanced label distribution shifts}} (imbalance ratio $q_{max}/q_{min}=\infty$)  regarding \textbf{Accuracy (\%)}. ``BN"/``GN"/``LN" is short for Batch/Group/Layer normalization.
    }
    \label{tab:imagenet-c-label-shift-infity-level3}
\newcommand{\tabincell}[2]{\begin{tabular}{@{}#1@{}}#2\end{tabular}}
 \begin{center}
 \begin{threeparttable}
 \Huge
    \resizebox{1.0\linewidth}{!}{
 	\begin{tabular}{l|ccc|cccc|cccc|cccc|>{\columncolor{blue!8}}c}
 	\multicolumn{1}{c}{} & \multicolumn{3}{c}{Noise} & \multicolumn{4}{c}{Blur} & \multicolumn{4}{c}{Weather} & \multicolumn{4}{c}{Digital}  \\
 	 Model+Method & Gauss. & Shot & Impul. & Defoc. & Glass & Motion & Zoom & Snow & Frost & Fog & Brit. & Contr. & Elastic & Pixel & JPEG & Avg.\\
 	\midrule

        ResNet50 (BN) & 27.7  & 25.2  & 25.1  & 37.8  & 16.7  & 37.8  & 35.3  & 35.2  & 32.1  & 46.7  & 69.5  & 46.2  & 55.4  & 46.2  & 59.4  & 39.8  \\ 
        ~~$\bullet~$MEMO & 37.6  & 34.5  & 36.7  & 41.4  & 23.4  & 44.4  & 40.9  & 44.6  & 37.3  & 52.4  & 70.5  & 56.3  & 58.7  & 55.2  & 60.9  & 46.3 \\ %
        ~~$\bullet~$DDA & 49.9  & 50.0  & 49.2  & 33.2  & 31.9  & 38.0  & 36.7  & 35.1  & 34.1  & 35.01  & 64.9  & 33.7  & 59.3 & 53.9  & 59.0  & 44.3  \\

        ~~$\bullet~$Tent & 3.4  & 3.2  & 3.2  & 2.3  & 2.0  & 2.4  & 3.4  & 2.4  & 2.4  & 4.6  & 5.4  & 3.0  & 4.8  & 4.6  & 4.5  & 3.4  \\ 
        ~~$\bullet~$EATA & 1.3  & 0.9  & 1.1  & 0.6  & 0.6  & 1.2  & 1.4  & 1.3  & 1.3  & 1.9  & 4.1  & 1.6  & 2.7  & 2.4  & 3.1  & 1.7  \\

\cmidrule{1-17}

        ResNet50 (GN)    & 54.5  & 52.9  & 53.1  & 44.4  & 21.2  & 49.8  & 39.3  & 54.9  & 54.1  & 55.8  & 75.3  & 69.7  & 59.6  & 59.7  & 66.4  & 54.1  \\
~~$\bullet~$MEMO & 55.9  & 54.3  & 54.1  & 40.1  & 23.1  & 49.5  & 41.4  & 54.8  & 54.1  & 57.6  & 75.7  & 70.2  & 60.2  & 61.5  & 66.7  & 54.6  \\
~~$\bullet~$DDA       & 61.0  & 61.0  & 60.5  & 39.3  & 37.3  & 46.4  & 39.7  & 47.7  & 48.1  & 29.9  & 69.9  & 57.8  & 62.8  & 60.1  & 63.8  & 52.4 \\

~~$\bullet~$Tent       &   59.1  & 58.6  & 58.3  & 39.0  & 27.9  & 54.7  & 41.1  & 51.3  & 41.4  & 62.0  & 75.2  & 70.1  & 62.3  & 63.7  & 66.4  & 55.4  \\ 
~~$\bullet~$EATA       & 52.3  & 52.9  & 51.7  & 35.7  & 30.1  & 46.4  & 39.6  & 43.8  & 39.8  & 55.7  & 72.4  & 66.6  & 54.7  & 56.0  & 56.2  & 50.3  \\ 
\rowcolor{pink!15}~~$\bullet~$SAR (ours)  & 60.8$_{\pm0.1}$ & 60.5$_{\pm0.3}$ & 60.2$_{\pm0.2}$ & 47.9$_{\pm0.5}$ & 36.7$_{\pm0.7}$ & 58.2$_{\pm0.2}$ & 49.7$_{\pm0.5}$ & 57.9$_{\pm0.3}$ & 53.6$_{\pm0.0}$ & 65.0$_{\pm0.1}$ & 76.4$_{\pm0.2}$ & 71.0$_{\pm0.0}$ & 67.0$_{\pm0.2}$ & 65.8$_{\pm0.1}$ & 67.6$_{\pm0.0}$ & 59.9$_{\pm0.1}$ \\
\rowcolor{pink!30}~~$\bullet~$\mysarE (ours) & 65.3$_{\pm0.0}$ & 65.3$_{\pm0.1}$ & 64.2$_{\pm0.1}$ & 56.0$_{\pm0.1}$ & 51.5$_{\pm0.1}$ & 63.4$_{\pm0.2}$ & 58.5$_{\pm0.2}$ & 62.7$_{\pm0.1}$ & 60.3$_{\pm0.2}$ & 69.9$_{\pm0.2}$ & 77.2$_{\pm0.1}$ & 73.6$_{\pm0.1}$ & 71.8$_{\pm0.2}$ & 70.5$_{\pm0.2}$ & 70.6$_{\pm0.2}$ & \textbf{65.4$_{\pm0.1}$} \\

 	\cmidrule{1-17}
        VitBase (LN)   & 51.5  & 46.8  & 50.4  & 48.7  & 37.1  & 54.7  & 41.6  & 35.1  & 33.3  & 68.0  & 69.3  & 74.9  & 65.9  & 66.0  & 63.6  & 53.8 \\ 
~~$\bullet~$MEMO & 62.1  & 57.9  & 61.5  & 57.2  & 45.6  & 62.0  & 49.9  & 46.5  & 43.1  & 74.1  & 75.8  & 79.7  & 72.6  & 72.3  & 70.6  & 62.1 \\ 
~~$\bullet~$DDA       & 59.7  & 58.2  & 59.4  & 43.5  & 43.3  & 50.5  & 41.0  & 34.3  & 34.4  & 55.4  & 65.0  & 64.2  & 64.1  & 63.8  & 62.9  & 53.3 \\

~~$\bullet~$Tent       &  68.7  & 68.0  & 68.1  & 68.2  & 63.8  & 70.9  & 63.8  & 67.6  & 41.9  & 76.3  & 78.8  & 79.5  & 75.9  & 76.7  & 73.7  & 69.5  \\ 
~~$\bullet~$EATA       & 65.3  & 62.6  & 63.6  & 63.0  & 57.1  & 66.3  & 59.3  & 64.5  & 61.0  & 73.3  & 76.9  & 75.9  & 74.2  & 74.8  & 73.1  & 67.4  \\ 
\rowcolor{pink!15}~~$\bullet~$SAR (ours)  & 68.8$_{\pm0.1}$ & 68.2$_{\pm0.1}$ & 68.4$_{\pm0.2}$ & 68.3$_{\pm0.2}$ & 64.7$_{\pm0.0}$ & 71.0$_{\pm0.2}$ & 64.2$_{\pm0.3}$ & 68.1$_{\pm0.1}$ & 66.0$_{\pm0.1}$ & 76.4$_{\pm0.1}$ & 79.0$_{\pm0.1}$ & 79.6$_{\pm0.1}$ & 76.2$_{\pm0.3}$ & 77.1$_{\pm0.1}$ & 74.1$_{\pm0.2}$ & 71.3$_{\pm0.1}$ \\
\rowcolor{pink!30}~~$\bullet~$\mysarE (ours)  & 71.6$_{\pm0.1}$ & 71.5$_{\pm0.0}$ & 71.2$_{\pm0.0}$ & 70.5$_{\pm0.2}$ & 68.7$_{\pm0.1}$ & 73.7$_{\pm0.1}$ & 69.0$_{\pm0.2}$ & 72.2$_{\pm0.1}$ & 69.8$_{\pm0.0}$ & 78.2$_{\pm0.0}$ & 80.3$_{\pm0.0}$ & 80.1$_{\pm0.1}$ & 78.1$_{\pm0.2}$ & 78.6$_{\pm0.1}$ & 76.7$_{\pm0.1}$ & \textbf{74.0$_{\pm0.04}$} \\
        
	\end{tabular}
	}
	 \end{threeparttable}
	 \end{center}
\end{table*}

\subsection{Comparisons with State-of-the-arts under Batch Size of 1 on Severity Level 3}

We provide more results regarding batch size = 1 of all compared methods in Table~\ref{tab:imagenet-c-bs1-level3}. The results are consistent with that of the main paper (severity level 5), and our \mysarE performs best in the average of 15 different corruption types.

\begin{table*}[h]
    \renewcommand\thetable{C}

    \caption{Comparisons with state-of-the-art methods on ImageNet-C of severity level 3 under \textbf{\textsc{Batch Size=1}} regarding \textbf{Accuracy (\%)}. ``BN"/``GN"/``LN" is short for Batch/Group/Layer normalization.
    }
    \label{tab:imagenet-c-bs1-level3}
\newcommand{\tabincell}[2]{\begin{tabular}{@{}#1@{}}#2\end{tabular}}
 \begin{center}
 \begin{threeparttable}
 \Huge
    \resizebox{1.0\linewidth}{!}{
 	\begin{tabular}{l|ccc|cccc|cccc|cccc|>{\columncolor{blue!8}}c}
 	\multicolumn{1}{c}{} & \multicolumn{3}{c}{Noise} & \multicolumn{4}{c}{Blur} & \multicolumn{4}{c}{Weather} & \multicolumn{4}{c}{Digital}  \\

 	 Model+Method & Gauss. & Shot & Impul. & Defoc. & Glass & Motion & Zoom & Snow & Frost & Fog & Brit. & Contr. & Elastic & Pixel & JPEG & Avg.\\
 	\midrule
        ResNet50 (BN) & 27.6  & 25.0  & 25.1  & 38.0  & 16.9  & 37.7  & 35.2  & 35.2  & 32.1  & 46.6  & 69.6  & 46.0  & 55.6  & 46.2  & 59.3  & 39.7  \\ 
        ~~$\bullet~$MEMO & 37.5  & 34.3  & 36.6  & 41.2  & 23.3  & 44.2  & 41.0  & 44.5  & 37.4  & 52.3  & 70.5  & 56.0  & 58.7  & 55.0  & 60.8  & 46.2 \\ %
        ~~$\bullet~$DDA & 49.8  & 49.9  & 49.2  & 33.2  & 32.0  & 37.9  & 36.6  & 35.2  & 34.2  & 34.9  & 64.9  & 33.5  & 59.3  & 53.9  & 59.0  & 44.2  \\

        ~~$\bullet~$Tent & 0.1  & 0.2  & 0.2  & 0.2  & 0.1  & 0.1  & 0.2  & 0.2  & 0.2  & 0.2  & 0.2  & 0.2  & 0.2  & 0.2  & 0.2  & 0.2  \\ 
        ~~$\bullet~$EATA & 0.2  & 0.2  & 0.1  & 0.2  & 0.1  & 0.2  & 0.2  & 0.1  & 0.2  & 0.2  & 0.2  & 0.2  & 0.2  & 0.2  & 0.2  & 0.2  \\

\cmidrule{1-17}

        ResNet50 (GN)    &  54.5  & 52.8  & 53.1  & 44.3  & 21.2  & 49.7  & 39.2  & 54.8  & 54.0  & 55.8  & 75.4  & 69.8  & 59.6  & 59.7  & 66.3  & 54.0  \\
~~$\bullet~$MEMO & 55.7  & 54.2  & 53.9  & 40.0  & 22.8  & 49.2  & 41.2  & 54.8  & 54.1  & 57.6  & 75.5  & 69.9  & 60.0  & 61.3  & 66.6  & 54.5  \\ 
~~$\bullet~$DDA       & 61.0  & 60.9  & 60.4  & 39.2  & 37.2  & 46.4  & 39.7  & 47.7  & 48.0  & 29.8  & 70.0  & 58.0  & 62.7  & 60.2  & 63.8  & 52.3 \\

~~$\bullet~$Tent       & 58.8  & 58.5  & 58.7  & 38.2  & 26.8  & 54.9  & 42.6  & 51.6  & 38.8  & 61.9  & 75.3  & 70.0  & 62.3  & 63.6  & 66.3  & 55.2  \\
~~$\bullet~$EATA       & 59.2  & 58.7  & 58.8  & 45.7  & 32.6  & 55.5  & 45.9  & 56.4  & 52.7  & 63.6  & 75.9  & 71.1  & 64.7  & 64.5  & 67.8  & 58.2  \\
\rowcolor{pink!15}~~$\bullet~$SAR (ours)  & 60.3$_{\pm0.1}$ & 59.6$_{\pm0.1}$ & 59.5$_{\pm0.1}$ & 46.6$_{\pm1.1}$ & 33.0$_{\pm0.5}$ & 57.5$_{\pm0.1}$ & 47.8$_{\pm0.1}$ & 57.8$_{\pm0.2}$ & 52.8$_{\pm0.1}$ & 65.1$_{\pm0.1}$ & 76.7$_{\pm0.1}$ & 71.4$_{\pm0.1}$ & 67.3$_{\pm0.2}$ & 66.0$_{\pm0.1}$ & 67.8$_{\pm0.0}$ & 59.3$_{\pm0.1}$ \\

\rowcolor{pink!30}~~$\bullet~$\mysarE (ours) & 65.9$_{\pm0.0}$ & 65.7$_{\pm0.1}$ & 65.0$_{\pm0.0}$ & 57.3$_{\pm0.1}$ & 53.5$_{\pm0.1}$ & 64.8$_{\pm0.0}$ & 60.8$_{\pm0.1}$ & 64.8$_{\pm0.1}$ & 61.8$_{\pm0.1}$ & 71.4$_{\pm0.1}$ & 77.3$_{\pm0.0}$ & 74.1$_{\pm0.0}$ & 72.9$_{\pm0.0}$ & 71.7$_{\pm0.0}$ & 71.5$_{\pm0.1}$ & \textbf{66.6$_{\pm0.02}$} \\

        \cmidrule{1-17}

        VitBase (LN)   &  51.6  & 46.9  & 50.5  & 48.7  & 37.2  & 54.7  & 41.6  & 35.1  & 33.5  & 67.8  & 69.3  & 74.8  & 65.8  & 66.0  & 63.7  & 53.8  \\ 
~~$\bullet~$MEMO & 61.9  & 57.7  & 61.4  & 57.0  & 45.4  & 61.8  & 49.8  & 46.6  & 43.1  & 73.9  & 75.7  & 79.6  & 72.6  & 72.1  & 70.5  & 61.9  \\ 
~~$\bullet~$DDA       & 59.8  & 58.2  & 59.5  & 43.4  & 43.2  & 50.4  & 40.9  & 34.2  & 34.3  & 55.2  & 64.9  & 64.0  & 64.2  & 63.7  & 62.8  & 53.2 \\

~~$\bullet~$Tent       & 67.1  & 66.2  & 66.3  & 66.3  & 60.9  & 69.1  & 61.4  & 65.2  & 60.4  & 75.2  & 78.1  & 78.8  & 74.9  & 75.8  & 72.4  & 69.2  \\ 
~~$\bullet~$EATA       & 60.7  & 58.5  & 61.6  & 60.1  & 51.8  & 64.2  & 54.8  & 53.3  & 52.6  & 72.5  & 73.6  & 77.9  & 71.3  & 71.3  & 69.7  & 63.6  \\
\rowcolor{pink!15}~~$\bullet~$SAR (ours)  & 68.5$_{\pm0.1}$ & 67.8$_{\pm0.1}$ & 68.0$_{\pm0.1}$ & 67.8$_{\pm0.2}$ & 63.1$_{\pm0.0}$ & 70.7$_{\pm0.1}$ & 63.5$_{\pm0.1}$ & 66.9$_{\pm0.2}$ & 62.8$_{\pm2.1}$ & 75.8$_{\pm0.5}$ & 77.7$_{\pm0.8}$ & 78.4$_{\pm0.4}$ & 74.7$_{\pm1.5}$ & 75.7$_{\pm0.5}$ & 72.7$_{\pm1.1}$ & 70.3$_{\pm0.3}$ \\
\rowcolor{pink!30}~~$\bullet~$\mysarE (ours) & 73.5$_{\pm0.1}$ & 73.5$_{\pm0.0}$ & 73.3$_{\pm0.1}$ & 72.2$_{\pm0.1}$ & 71.0$_{\pm0.1}$ & 75.3$_{\pm0.1}$ & 71.7$_{\pm0.0}$ & 74.5$_{\pm0.0}$ & 72.0$_{\pm0.1}$ & 79.5$_{\pm0.0}$ & 81.3$_{\pm0.0}$ & 80.9$_{\pm0.0}$ & 79.5$_{\pm0.0}$ & 80.0$_{\pm0.1}$ & 78.1$_{\pm0.1}$ & \textbf{75.8$_{\pm0.01}$} \\

	\end{tabular}
	}
	 \end{threeparttable}
	 \end{center}
\end{table*}

\subsection{Additional Results on ImageNet-R}
We further conduct experiments on ImageNet-R under two wild test settings: online imbalanced label distribution shifts (in Table~\ref{tab:imagenet_r_label_shift}) and batch size = 1 (in Table~\ref{tab:imagenet_r_bs1}). The overall results are consistent with that on ImageNet-C: 1) ResNet50-GN and VitBase-LN perform more stable than ResNet50-BN; 2) Compared with Tent and EATA, \mysarE achieves the best performance on ResNet50-GN and VitBase-LN.

\begin{table}[!ht]
\centering
\begin{minipage}[t]{0.49\linewidth}
    \renewcommand\thetable{D}
\centering
    \caption{Comparison in terms of accuracy (\%) under the wild setting \textbf{online imbalanced label distribution shifts} on ImageNet-R.
    }
    \label{tab:imagenet_r_label_shift}
     \begin{threeparttable}
        \resizebox{1.0\linewidth}{!}{
     	\begin{tabular}{l|cccc}
     	 Method & ResNet50-BN & ResNet50-GN & VitBase-LN  \\
     	\midrule
        No Adapt.  & 36.2 & 40.8 &43.1 \\
        Tent & 6.6 & 41.7 & 45.2 \\
        EATA & 5.8 & 40.9 & 47.5 \\
        \rowcolor{pink!30}SAR (ours) & - & 42.9 & 52.0 \\
        \rowcolor{pink!30}\mysarE (ours) & - & \textbf{48.4} & \textbf{56.2} \\
    	\end{tabular}
    	}
    	\end{threeparttable}
\end{minipage}\hfill%
\begin{minipage}[t]{0.49\linewidth}
\renewcommand\thetable{E}
\centering
    \caption{Comparison in terms of accuracy (\%) under the wild setting \textbf{single sample adaptation (batch size = 1)} on ImageNet-R.
    }
    \label{tab:imagenet_r_bs1}
    \begin{threeparttable}
    \resizebox{1.0\linewidth}{!}{
 	\begin{tabular}{l|cccc}
 	 Method & ResNet50-BN & ResNet50-GN & VitBase-LN  \\
 	\midrule
    No Adapt.  & 36.2 & 40.8 & 43.1 \\
    Tent & 0.6 & 42.2 & 40.5 \\
    EATA & 0.6 & 42.3 & 52.5 \\
    \rowcolor{pink!30}SAR (ours) & - & 43.9 & 53.1 \\
    \rowcolor{pink!30}\mysarE (ours) & - & \textbf{49.9} & \textbf{62.8} \\
	\end{tabular}
	}
	 \end{threeparttable}
\end{minipage}
\end{table}

\clearpage
\section{Additional Ablate Results}\label{sec:more_ablation}

\subsection{Effects of Components in \methodname}\label{sec:more_ablation_for_sar}
\begin{figure*}[h]
\renewcommand\thefigure{B}
\centering
\includegraphics[width=0.4\linewidth]{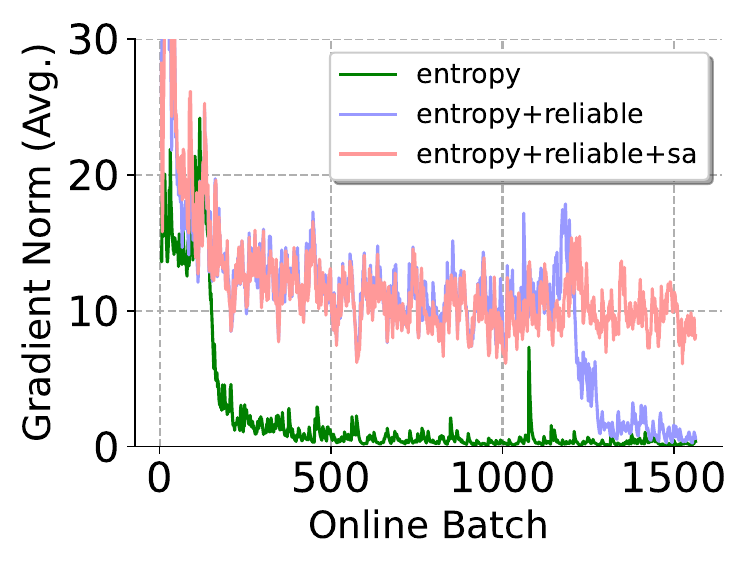}
\vspace{-0.2in}
\caption{The evolution of gradients norm during online test-time adaptation. Results on VitBase, ImageNet-C, shot noise, severity level 5, online imbalance label shift (imbalance ratio = $\infty$). ``reliable" and ``sharpness-aware (sa)" are short for Eqn.~(\ref{eq:reliable_entropy}) and Eqn.~(\ref{eq:sa_entropy}), respectively.}
\label{fig:ablation_grad_norm_changes}
\end{figure*}

We further ablate the effects of components in our \methodname in Figure \ref{fig:ablation_grad_norm_changes} by plotting the changes of gradients norms of our methods with different components. From the results, both the reliable (Eqn.~\ref{eq:reliable_entropy}) and sharpness-aware (sa) (Eqn.~\ref{eq:sa_entropy}) modules together ensure the model's gradients keep in a normal range during the whole online adaptation process, which is consistent with our previous results in Section~\ref{sec:ablate_results_main}.

We also present more detailed ablation experiments under the wild test settings to demonstrate the effectiveness of reliable entropy, sharpness-aware minimization, and the recovery scheme in \mysar. From Tables~\ref{tab:ablation_ls_infty}-\ref{tab:ablatin_mix_shifts}, both the reliable entropy and sharpness-aware optimization work together to enable stable online TTA in the wild except for very few cases, \ie, VitBase-LN on \textit{snow} and \textit{frost} under imbalanced label shifts, and on \textit{frost} under BS=1. For these cases, our model recovery scheme takes effect, \ie, $54.5\%\rightarrow 58.2\%$ and $55.7\%\rightarrow 56.4\%$ on VitBase-LN regarding average accuracy under imbalanced label shifts and BS=1, respectively.

\begin{table*}[h]
    \renewcommand\thetable{F}
    \caption{Effects of components in \methodname. We report the \textbf{Accuracy (\%)} on ImageNet-C (level 5) under \textbf{\textsc{online imbalanced label shifts}} (imbalance ratio $q_{max}/q_{min}$ = $\infty$). ``reliable" and ``sharpness-aware (sa)" denote Eqn.~(\ref{eq:reliable_entropy}) and Eqn.~(\ref{eq:sa_entropy}), ``recover" denotes the model recovery scheme.
    }
    \vspace{-0.05in}
    \label{tab:ablation_ls_infty}
\newcommand{\tabincell}[2]{\begin{tabular}{@{}#1@{}}#2\end{tabular}}
 \begin{center}
 \begin{threeparttable}
 \large
    \resizebox{1.0\linewidth}{!}{
 	\begin{tabular}{l|ccc|cccc|cccc|cccc|>{\columncolor{blue!8}}c}
 	\multicolumn{1}{c}{} & \multicolumn{3}{c}{Noise} & \multicolumn{4}{c}{Blur} & \multicolumn{4}{c}{Weather} & \multicolumn{4}{c}{Digital}  \\
 	 Model+Method & Gauss. & Shot & Impul. & Defoc. & Glass & Motion & Zoom & Snow & Frost & Fog & Brit. & Contr. & Elastic & Pixel & JPEG & Avg.  \\
 	\midrule

        ResNet50 (GN)+Entropy  & 2.6  & 3.3  & 2.7  & 13.9  & 7.9  & 19.5  & 17.0  & 16.5  & 21.9  & 1.8  & 70.5  & 42.2  & 6.6  & 49.4  & 53.7  & 22.0  \\ 
~~\ding{59}~reliable      & 34.5  & 36.8  & 36.2  & 19.5  & 3.1  & 33.6  & 14.5  & 20.5  & 38.3  & 2.4  & 71.9  & 47.0  & 8.3  & 52.1  & 56.4  & 31.7  \\ 
~~\ding{59}~reliable+sa      & 33.8  & 35.9  & 36.4  & 19.2  & 18.7  & 33.6  & 24.5  & 23.5  & 45.2  & 49.3  & 71.9  & 46.6  & 9.2  & 51.6  & 56.4  & \textbf{37.0}  \\ 
\rowcolor{pink!30}~~\ding{59}~reliable+sa+recover      & 33.6  & 36.1  & 36.2  & 19.1  & 18.6  & 33.9  & 24.7  & 22.5  & 45.7  & 49.0  & 71.9  & 46.6  & 9.2  & 51.5  & 56.3  & \textbf{37.0}  \\ %

\cmidrule{1-17}
        VitBase (LN)+Entropy   & 32.7  & 1.4  & 34.6  & 54.4  & 52.3  & 58.2  & 52.2  & 7.7  & 12.0  & 69.3  & 76.1  & 66.1  & 56.7  & 69.4  & 66.4  & 47.3  \\
~~\ding{59}~reliable       &  47.8  & 35.7  & 48.4  & 55.2  & 54.1  & 58.6  & 54.4  & 13.3  & 21.4  & 69.5  & 76.2  & 66.1  & 60.2  & 69.3  & 66.7  & 53.1  \\ 
~~\ding{59}~reliable+sa       & 47.9  & 47.6  & 48.5  & 55.4  & 54.2 & 58.8  & 54.6  & 19.7  & 22.1  & 69.4  & 76.3  & 66.2  & 60.9  & 69.4  & 66.6  & 54.5  \\ 
\rowcolor{pink!30}~~\ding{59}~reliable+sa+recover      & 47.9  & 47.6  & 48.5  & 55.4  & 54.2  & 58.8  & 54.6  & 49.1  & 48.3 & 69.4  & 76.3  & 66.2  & 60.9  & 69.4  & 66.6  & \textbf{58.2} \\

	\end{tabular}
	}
	 \end{threeparttable}
	 \end{center}
	 \vspace{-0.15in}
\end{table*}

\begin{table*}[h]
    \renewcommand\thetable{G}
    \caption{Effects of components in \methodname. We report the \textbf{Accuracy (\%)} on ImageNet-C (level 5) under \textbf{\textsc{Batch Size=1}}. ``reliable" and ``sharpness-aware (sa)" denote Eqn.~(\ref{eq:reliable_entropy}) and Eqn.~(\ref{eq:sa_entropy}), ``recover" denotes the model recovery scheme.}
    \label{tab:ablation_bs1}
\newcommand{\tabincell}[2]{\begin{tabular}{@{}#1@{}}#2\end{tabular}}
 \begin{center}
 \begin{threeparttable}
 \large
    \resizebox{1.0\linewidth}{!}{
 	\begin{tabular}{l|ccc|cccc|cccc|cccc|>{\columncolor{blue!8}}c}
 	\multicolumn{1}{c}{} & \multicolumn{3}{c}{Noise} & \multicolumn{4}{c}{Blur} & \multicolumn{4}{c}{Weather} & \multicolumn{4}{c}{Digital}  \\
 	 Model+Method & Gauss. & Shot & Impul. & Defoc. & Glass & Motion & Zoom & Snow & Frost & Fog & Brit. & Contr. & Elastic & Pixel & JPEG & Avg.  \\
 	\midrule

        ResNet50 (GN)+Entropy   &  2.5  & 2.9  & 2.5  & 13.5  & 3.6  & 18.6  & 17.6  & 15.3  & 23.0  & 1.4  & 70.4  & 42.2  & 6.2  & 49.2  & 53.8  & 21.5\\ 
        ~~\ding{59}~reliable       &  22.8  & 25.5 &  23.0  &  18.4  & 14.8  &  27.0  & 28.6 & 40.0 & 43.3  & 18.4 & 71.5  &  43.1  &  15.2   & 45.6  & 55.4  & 32.8 \\ 
        ~~\ding{59}~reliable+sa       &  23.8  & 26.4 &  24.0  &  18.6  & 15.4  &  28.3  & 30.5 & 44.8 & 44.8  & 26.7 & 72.4  &  44.5  &  12.2   & 46.9  & 65.1 & \textbf{34.4} \\ 
        \rowcolor{pink!30}~~\ding{59}~reliable+sa+recover      &  23.8  & 26.4 &  23.9  &  18.5  & 15.4  &  28.3  & 30.6 & 44.6 & 44.9  & 24.7 & 72.4  &  44.4  &  12.3   & 47.0  & 65.1 & 34.2 \\
\cmidrule{1-17}
        VitBase (LN)+Entropy    &  42.2  & 1.0  & 43.3  & 52.4  & 48.2  & 55.5  & 50.5  & 16.5  & 16.9  & 66.4  & 74.9  & 64.7  & 51.6  & 67.0  & 64.3  & 47.7 \\
        ~~\ding{59}~reliable       &  34.8  & 4.2  &  35.5  &  50.5  & 45.9  &  54.0  & 48.6 & 52.5 & 47.8  & 65.5 & 74.5  &  63.4  &  51.4   & 65.6  & 63.0  & 50.5  \\ 
        ~~\ding{59}~reliable+sa        &  40.4  & 37.3 &  41.2  &  53.6  & 50.6  &  57.3  & 53.0 & 58.7 & 40.9  & 68.8 & 75.9  &  65.7  &  57.9   & 69.0  & 66.0 & 55.7\\ 
        \rowcolor{pink!30}~~\ding{59}~reliable+sa+recover       &  40.4  & 37.3 &  41.2  &  53.6  & 50.6  &  57.3  & 53.0 & 58.7 & 50.7  & 68.8 & 75.4  &  65.7  &  57.9   & 69.0  & 66.0 & \textbf{56.4} \\
	\end{tabular}
	}
	 \end{threeparttable}
	 \end{center}
\end{table*}

\begin{table}[h]
    \renewcommand\thetable{H}

    \centering
    \caption{Effects of components in SAR. We report the \textbf{Accuracy (\%)} under \textbf{\textsc{Mixed Domain Shifts}}, \ie, mixture of 15 corruption types of ImageNet-C with severity level 5.
    ``reliable" and ``sharpness-aware (sa)" denote Eqn.~(\ref{eq:reliable_entropy}) and Eqn.~(\ref{eq:sa_entropy}), ``recover" denotes the model recovery scheme.}
    \label{tab:ablatin_mix_shifts}
    \begin{minipage}{0.35\textwidth}
            \centering
            \newcommand{\tabincell}[2]{\begin{tabular}{@{}#1@{}}#2\end{tabular}}
 \begin{center}
 \begin{threeparttable}
    \resizebox{1.0\linewidth}{!}{
        \begin{tabular}{lc}
 	 Model + Method  & Accuracy (\%)  \\
 	\midrule
 
        ResNet50 (GN)+Entropy  & 13.4  \\ 
        ~~\ding{59}~reliable      & 38.1  \\ 
        ~~\ding{59}~reliable+sa      & 38.2 \\ 
        \rowcolor{pink!30}~~\ding{59}~reliable+sa+recover      & 38.2  \\
	\end{tabular}
        }
         \end{threeparttable}
         \end{center}
    \end{minipage}
    ~~~
    \begin{minipage}{0.35\textwidth}
            \centering
            \newcommand{\tabincell}[2]{\begin{tabular}{@{}#1@{}}#2\end{tabular}}
 \begin{center}
 \begin{threeparttable}
    \resizebox{1.0\linewidth}{!}{
        \begin{tabular}{lc}
 	 Model + Method  & Accuracy (\%)  \\
 	\midrule
        ViTBase (LN)+Entropy  & 16.5  \\ 
        ~~\ding{59}~reliable      & 55.2  \\ 
        ~~\ding{59}~reliable+sa      & 57.1 \\ 
        \rowcolor{pink!30}~~\ding{59}~reliable+sa+recover      & 57.1  \\
	\end{tabular}
        }
         \end{threeparttable}
         \end{center}
    \end{minipage}
\end{table}

\subsection{Visualization of Loss Surface Learned by \methodname}

\begin{figure}[h]
\renewcommand\thefigure{C}
\centering
\includegraphics[width=0.5\linewidth]{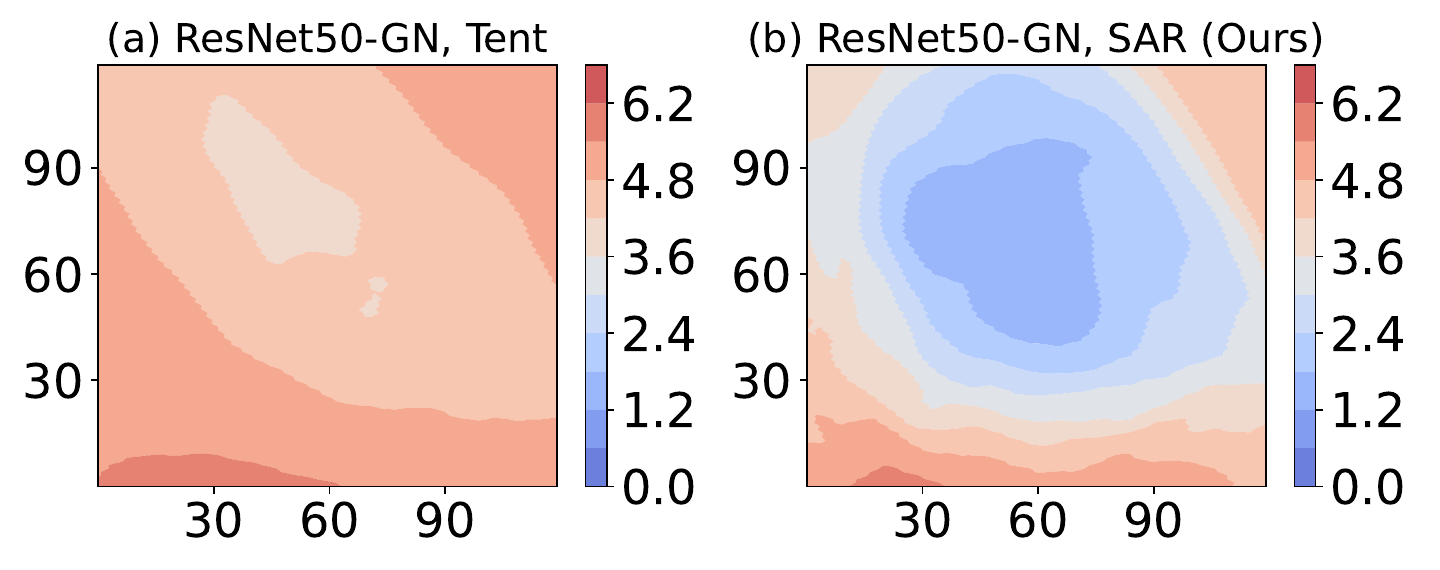}
\vspace{-0.1in}
\caption{Visualization of loss (entropy) surface. Models are learned on ImageNet-C of Gaussian noise with severity level 5.}
\label{fig:loss_landscape_r50_gn}
\end{figure}

In Figure \ref{fig:loss_landscape} in the main paper, we have visualized the loss surface of Tent and our \methodname on VitBase-LN. In this section, we further provide visualizations of ResNet50-GN. We select a checkpoint at batch 120 to plot the loss surface for Tent, since after batch 120 this model starts to collapse. In this case, the loss (entropy) is hard to degrade and cannot find a proper minimum. For our \methodname, the model weights for plotting are obtained after the adaptation on the whole test set. By comparing Figures \ref{fig:loss_landscape_r50_gn} (a)-(b), \methodname helps to stabilize the entropy minimization process and find proper minima. 

\subsection{Sensitivity of $\rho$ in SAR}
The hyper-parameter $\rho$ (in Eqn.~(\ref{eq:sa_entropy})) for sharpness-aware optimization is not hard to tune in our TTA. Following~\cite{foret2020sharpness}, we set $\rho=0.05$ in all experiments, and it works well with different architectures (ResNet50-BN, ResNet50-GN, VitBase-LN) on different datasets (ImageNet-C, ImageNet-R). In Table~\ref{tab:sensitivity_rho}, we also conduct a sensitivity analysis of $\rho$, where SAR works well under the range [0.03, 0.1].

\begin{table}[h]
    \renewcommand\thetable{I}
    \setlength{\tabcolsep}{8pt}
    \caption{Sensitivity of $\rho$ in SAR. We report \textbf{Accuracy (\%)} on ImageNet-C (shot noise, severity level 5) under \textbf{\textsc{online imbalanced label shifts}}, where the imbalance ratio is $\infty$.}
     \vspace{-0.05in}
    \label{tab:sensitivity_rho}
\newcommand{\tabincell}[2]{\begin{tabular}{@{}#1@{}}#2\end{tabular}}
 \begin{center}
 \begin{threeparttable}
    \resizebox{1.0\linewidth}{!}{
 	\begin{tabular}{lccccccc}
 	 Method & No adapt & Tent & SAR ($\rho$=0.01) & SAR ($\rho$=0.03) & SAR ($\rho$=0.05) & SAR ($\rho$=0.07) & SAR ($\rho$=0.1) \\
 	\midrule
    Accuracy (\%) & 6.7    & 1.4  &       40.3        &       47.7        &       47.6        &       47.2        &       47.2       \\
	\end{tabular}
	}
	 \end{threeparttable}
	 \end{center}
\end{table}

\rpami{
\subsection{Sensitivity of $\alpha$ and $\beta$ in \mysarE}
As shown in Tables~\ref{tab:sensitivity_alpha} \& \ref{tab:sensitivity_beta}, our redundancy and inequity regularizer consistently improves upon \mysar across a wide range of values, \ie, even using a very large coefficient such as $\alpha=1e4$ and $\beta=200$, confirming that the performance gains are not dependent 
on precise tuning. For all models across testing scenarios, we fix $\alpha=1e3$ and $\beta=50$ to balance the regularization strength for online TTA.  
}
\begin{table}[h]
    \renewcommand\thetable{J}
    \setlength{\tabcolsep}{2.5pt}
    \caption{
    Sensitivity of $\alpha$ in \mysarE. We report \textbf{Accuracy (\%)} on ImageNet-C (Gauss noise, severity level 5) under \textbf{\textsc{online imbalanced label shifts}}, where the imbalance ratio is $\infty$. \mysarE-R denotes SAR with redundancy regularizer $R(\cdot)$.
    }
     \vspace{-0.05in}
    \label{tab:sensitivity_alpha}
\newcommand{\tabincell}[2]{\begin{tabular}{@{}#1@{}}#2\end{tabular}}
 \begin{center}
 \begin{threeparttable}
    \resizebox{1.0\linewidth}{!}{
 	\begin{tabular}{lcccccccc}
 	 Method & No adapt & SAR & \mysarE-R ($\alpha$=1e1) & \mysarE-R ($\alpha$=1e2) & \mysarE-R ($\alpha$=5e2) & \mysarE-R ($\alpha$=1e3) & \mysarE-R ($\alpha$=5e3) & \mysarE-R ($\alpha$=1e4) \\
 	\midrule
    Accuracy (\%) & 17.9 & 33.1 & 34.2 & 38.4 & 40.8 & 41.7& 42.0 & 41.6 \\
	\end{tabular}
	}
	 \end{threeparttable}
	 \end{center}
     \vspace{-0.2in}
\end{table}
\begin{table}[h!]
    \renewcommand\thetable{K}
    \setlength{\tabcolsep}{3pt}
    \caption{
    Sensitivity of $\beta$ in \mysarE. We report \textbf{Accuracy (\%)} on ImageNet-C (zoom blur, severity level 5) under \textbf{\textsc{online imbalanced label shifts}}, where the imbalance ratio is $\infty$. \mysarE-I denotes SAR with inequity regularizer $I(\cdot)$.
    }
     \vspace{-0.05in}
    \label{tab:sensitivity_beta}
\newcommand{\tabincell}[2]{\begin{tabular}{@{}#1@{}}#2\end{tabular}}
 \begin{center}
 \begin{threeparttable}
    \resizebox{1.0\linewidth}{!}{
 	\begin{tabular}{lcccccccc}
 	 Method & No adapt & SAR & \mysarE-I ($\beta$=5) & \mysarE-I ($\beta$=10) & \mysarE-I ($\beta$=20) & \mysarE-I ($\beta$=50) & \mysarE-I ($\beta$=100) & \mysarE-I ($\beta$=200) \\
 	\midrule
    Accuracy (\%) & 27.0 & 58.0 & 58.6 & 58.8 & 59.0 & 59.3 & 59.4 & 59.5 \\
	\end{tabular}
	}
	 \end{threeparttable}
	 \end{center}
\end{table}

\clearpage

\section{Additional Discussions}\label{sec:more_discussions}

\begin{figure*}[h]
\renewcommand\thefigure{D}
\centering
\includegraphics[width=1\linewidth]{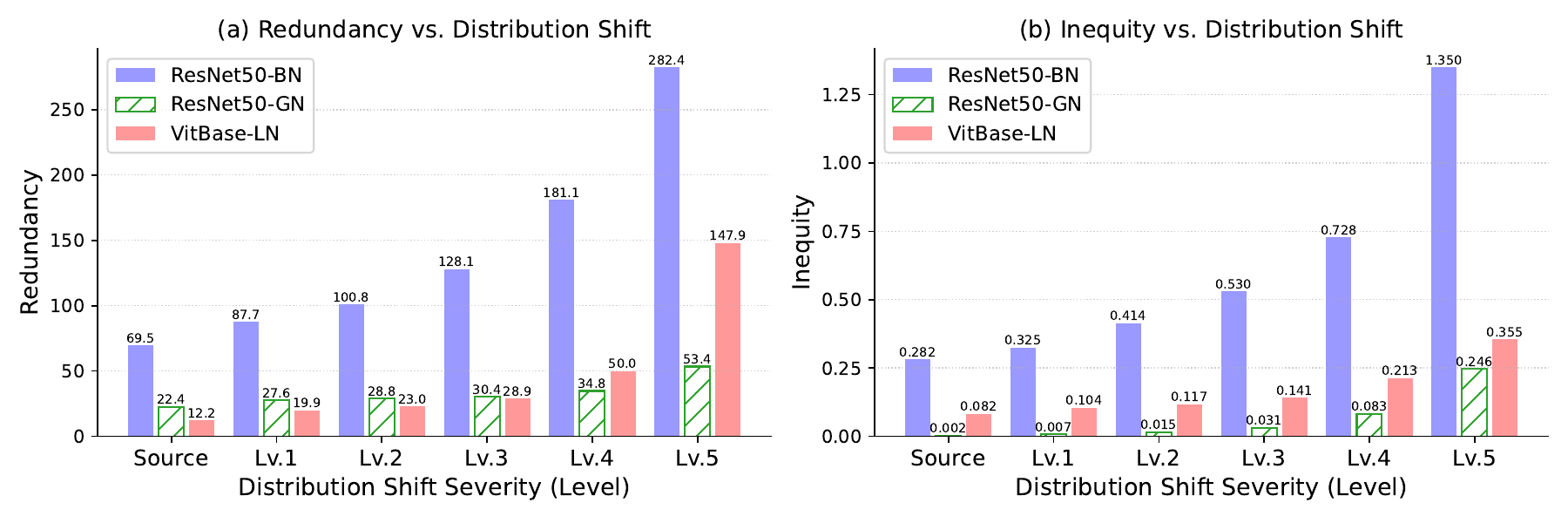}
\vspace{-0.3in}
\caption{\rpami{Impacts of distribution shifts severity on representation health. Feature redundancy and inequity are estimated over 50,000 images from different severity levels of ImageNet-C (Gaussian noise) or the original ImageNet-1k validation set (source).}}
\label{fig:redundancy_inequity_severity}
\end{figure*}

\rpami{
\subsection{Impacts of Distribution Shifts Severity on Models' Representation Health}\label{sec:shift_on_health}
In this section, we provide a more detailed analysis of how distribution shift severity influences models’ representation health, measured by feature redundancy and inequity. Our results in Figure~\ref{fig:redundancy_inequity_severity} reveal a monotonic increase in both redundancy and inequity as shift severity grows, confirming the trend observed in Figure~\ref{fig:feature_motivation}. Notably, the growth in feature redundancy and inequity accelerates at higher severities, \eg, feature redundancy increases by 3.1 (level 1{$\rightarrow$}level 2) \vs~ 97.9 (level 4{$\rightarrow$}level 5) on Vitbase-LN, suggesting that severe shifts rapidly undermine the model’s ability to maintain diverse and discriminative representations. As highlighted in Table~\ref{tab:inequity_collapse_correlation}, degraded initial health of representations exhibits a high correlation with eventual TTA collapse. These results collectively explain the increasing instability of TTA under severe shifts and motivate our feature regularization design at test time to stabilize adaptation.
}

\begin{table*}[h]
    \renewcommand\thetable{L}
    \caption{
    Correlations between inequity and TTA collapse. We report results on ImageNet-C (severity level 5) under a mild scenario. Initial Inequity $I(\cdot)$ is computed using the non-adapted model over 50,000 images from the target domain. c$_i$ denotes the \textit{i}-th class.
    }
    \label{tab:inequity_collapse_correlation}
\newcommand{\tabincell}[2]{\begin{tabular}{@{}#1@{}}#2\end{tabular}}
 \begin{center}
 \begin{threeparttable}
    \resizebox{0.8\linewidth}{!}{
 	\begin{tabular}{l|ccccc|ccc}
 	\multicolumn{1}{c}{} & \multicolumn{5}{c}{ResNet50-GN} & \multicolumn{3}{c}{Vitbase-LN} \\
 	 Method & Gauss. & Shot & Impul. & Fog & Elastic & Shot & Snow & Frost \\
 	\midrule
        \textbf{Initial} inequity $I(\cdot)$ towards & c$_{107}$ & c$_{107}$ & c$_{607}$ & c$_{974}$ & c$_{996}$ & c$_{971}$ & c$_{917}$ & c$_{728}$ \\
        Tent collapses into predicting & c$_{107}$ & c$_{107}$ & c$_{107}$ & c$_{974}$ & c$_{996}$ & c$_{971}$ & c$_{562}$ & c$_{728}$ \\

	\end{tabular}
	}
	 \end{threeparttable}
	 \end{center}
\end{table*}

\begin{figure}[h!]
\renewcommand\thefigure{E}
\centering
\includegraphics[width=0.8\linewidth]{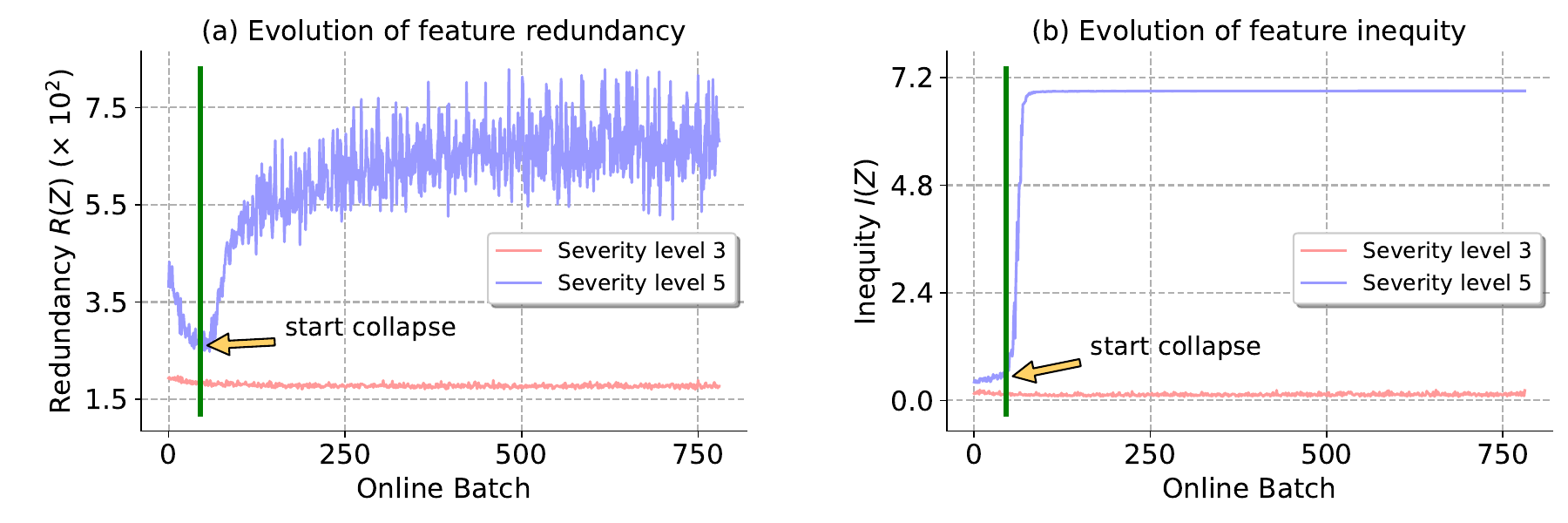}
\vspace{-0.15in}
\caption{\rpami{
Failure case analyses of online Tent~\cite{wang2021tent}. 
We illustrate the evolution of feature redundancy and inequity during collapse and non-collapse TTA. Experiments are conducted on ImageNet-C (shot noise) with VitBase-LN under a mild setting, \ie, data comes with shuffled labels.
}}
\vspace{0.1in}
\label{fig:vit_feature_evolution}
\end{figure}

\rpami{
\subsection{Evolutions of Feature Redundancy and Inequity on VitBase-LN}\label{sec:vit_feature_evolution}
As shown in Figure~\ref{fig:vit_feature_evolution} (a), under a severe distribution shift (severity level 5), feature redundancy first decreases when TTA still improves accuracy, suggesting an enhancement in model discrimination against shifts, but explodes immediately when collapse occurs. 
In contrast, feature redundancy stably decreases during a non-collapse TTA process at severity level 3.
Meanwhile, feature inequity exhibits an accelerated growth after TTA begins under severity level 5, which then triggers catastrophic model collapse. Conversely, under severity level 3, inequity remains low and stable throughout TTA, consistent with the trend observed in Figure~\ref{fig:feature_motivation}.
Together, these observations demonstrate how redundancy and inequity capture the dynamics that distinguish between successful and collapsing TTA, which motivates our design of feature redundancy and inequity regularizer on test data to stabilize test-time entropy minimization in the wild.
}

\rpami{
\subsection{Normalization Strategy for Feature Redundancy Estimation on ResNet50-GN}\label{sec:resnet-gn-sparsity}
ResNet50-GN applies group normalization across channels within each sample, followed by a ReLU function before producing the final feature maps. However, when several channels consistently have large positive values, GN forces other channels to become negative, which are then zeroed out by ReLU. As a result, many feature dimensions in ResNet50-GN can become inactive, \ie, nearly zero across all samples. In contrast, ResNet50-BN applies batch normalization (BN) across samples per dimension, ensuring that each dimension retains both positive and negative values across the batch to prevent the inactivity issue. As verified in Table~\ref{tab:gn_feature_sparsity}, 75.6\%/0.9\% of dimensions in ResNet50-GN already have a mean absolute value below 0.1/0.01 before adaptation, and the ratio increases to 87.0\%/32.7\% after Tent adaptation, whereas ResNet50-BN exhibits no sparsity across thresholds. 
When batch-wise normalization (zero-mean, unit-variance across the batch dimension) is applied to GN features, such near-zero dimensions can cause division by small variances, leading to extreme outliers and instability in redundancy estimation (Eqn.~\ref{eq:initial_redundancy}). Thus, on ResNet50-GN, we instead perform normalization along the feature dimension for Eqn.~(\ref{eq:initial_redundancy}), while keeping Eqn.~(\ref{eq:initial_redundancy}) unchanged for other models to calculate the defined feature redundancy.
}

\begin{table*}[h]
    \renewcommand\thetable{M}

    \caption{
    Prevalence of near-zero dimensions in ResNet50-GN.
    We report the \textbf{Percentage (\%)} of dimensions with mean absolute values below threshold $m$, before and after Tent on ImageNet-C (Gaussian, level 5) under a mild scenario, \ie, data comes in domain orders with shuffled label distribution.
    }
    \label{tab:gn_feature_sparsity}
\newcommand{\tabincell}[2]{\begin{tabular}{@{}#1@{}}#2\end{tabular}}
 \begin{center}
 \begin{threeparttable}
    \resizebox{0.7\linewidth}{!}{
 	\begin{tabular}{l|ccc|ccc}
 	\multicolumn{1}{c}{} & \multicolumn{3}{c}{ResNet50-GN} & \multicolumn{3}{c}{ResNet50-BN} \\
 	 Method & $m=0.1$ & $m=0.01$ & $m=0.001$ & $m=0.1$ & $m=0.01$ & $m=0.001$ \\
 	\midrule
        NoAdapt & 75.6 & 0.9 & 0.1 & 0.0 & 0.0 & 0.0 \\
        Tent Adapted & 87.0 & 32.7 & 1.4 & 0.0 & 0.0 & 0.0 \\

	\end{tabular}
	}
	 \end{threeparttable}
	 \end{center}
\end{table*}

\rpami{
\subsection{Effects of Redundancy Minimization as a Standalone Method}\label{sec:standalone_redundancy}
As shown in Figure~\ref{fig:tsne-visualization}, reducing feature redundancy penalizes representation overlap on test data, which can lead to more well-separated representations across classes.
This delivers a similar effect to Tent~\cite{wang2021tent}, which enforces entropy minimization per sample, whereas our $R(\cdot)$ operates at the batch level.
In this section, we further investigate the efficacy of redundancy minimization as a standalone method.
From Table~\ref{tab:standalone_redundancy}, minimizing redundancy consistently improves OOD accuracy across different architectures. Remarkably, redundancy minimization not only enhances performance on top of Tent, \ie, Tent+Redundancy $R(\cdot)$, but in some cases even surpasses Tent itself. For instance, +22.6\% \vs~+18.0\% (Tent)  on VitBase-LN and +18.5\% \vs~-5.5\% (Tent) on ResNet50-GN, regarding the accuracy improvement over the non-adapted baseline. These results highlight that: (1) redundancy minimization alone can serve as a strong and efficient adaptation strategy, particularly when conventional entropy minimization collapses; (2) combining $R(\cdot)$ with Tent yields the largest gains, indicating their complementary nature. Overall, this demonstrates the potential of redundancy minimization on the feature level as both a standalone and synergistic regularizer to enhance the effectiveness and stability of TTA.
}

\begin{table*}[h]
    \renewcommand\thetable{N}
    \caption{\rpami{Effects of our redundancy regularizer $R(\cdot)$. We report \textbf{Accuracy (\%)} on ImageNet-C (severity level 5) under a mild scenario, \ie, data comes in domain orders with shuffled label distribution. Redundancy $R(\cdot)$ is directly calculated over the current test batch without using the feature bank.}
    }
    \label{tab:standalone_redundancy}
\newcommand{\tabincell}[2]{\begin{tabular}{@{}#1@{}}#2\end{tabular}}
 \begin{center}
 \begin{threeparttable}
    \resizebox{0.65\linewidth}{!}{
 	\begin{tabular}{l|cccc}
 	 Model & NoAdapt & Tent & Redundancy $R(\cdot)$ & Tent+Redundancy $R(\cdot)$ \\
 	\midrule
        ResNet50 (BN) & 31.5 & 42.6$_{(+11.1)}$ & 32.9$_{(+1.4)}$ & 43.8$_{(+12.3)}$ \\
        ResNet50 (GN) & 30.6 & 25.1$_{(-5.5)}$ & 49.1$_{(+18.5)}$ & 50.7$_{(+20.1)}$ \\
        VitBase (LN) & 29.9 & 47.9$_{(+18.0)}$ & 52.5$_{(+22.6)}$ & 60.9$_{(+31.0)}$\\

	\end{tabular}
	}
	 \end{threeparttable}
	 \end{center}
\end{table*}

\rpami{
\subsection{Batch-imbalance Penalizing Strategies: Inequity (Ours) \vs~ Information Maximization}
In this section, we present the detailed results of Table~\ref{tab:inequity_information_maximization} in the main paper to further demonstrate the advantage of our inequity regularization. From Table~\ref{tab:detailed_inequity_infomax}, we highlight the following observations: 1) on domains where Tent alone fails, both our inequity regularizer and the diversity term in infomax~\cite{liang2020we} help mitigate collapse, while our inequity consistently achieves much higher OOD accuracy, \eg, 6.1\% (Tent) \textit{vs.} 31.1\% (Tent+InfoMax) \textit{vs.} 43.0\% (Tent+Inequity) on \textit{shot noise} with ResNet50-GN, demonstrating the superiority of our inequity regularizer to stabilize adaptation. 2) on domains where Tent alone works well, InfoMax often deteriorates the effectiveness of Tent due to its implicit confidence flattening effect, as discussed in Section~\ref{sec:ablate_results_main}. In contrast, our inequity regularization ensures a debiased representation centroid while each sample can retain a confident prediction, consistently improving upon Tent by a large margin, \eg, 66.1\% (Tent) \textit{vs.} 64.1\% (Tent+InfoMax) \textit{vs.} 70.0\% (Tent+Inequity) on \textit{fog} with VitBase-LN. 
Overall, these results show that our inequity regularizer not only stabilizes adaptation when Tent collapses, but also strengthens adaptation when Tent already works well, \eg, +11.4\% on VitBase-LN \wrt the average accuracy, suggesting the superiority of our inequity $I(\cdot)$ for online TTA.
}

\begin{table*}[ht]
    \renewcommand\thetable{O}
    \caption{Effects of different batch-imbalance penalties. Results are reported on ImageNet-C (severity level 5) \wrt \textbf{Accuracy (\%)} under a mild scenario.
    }
    \label{tab:detailed_inequity_infomax}
\newcommand{\tabincell}[2]{\begin{tabular}{@{}#1@{}}#2\end{tabular}}
 \begin{center}
 \begin{threeparttable}
 \large
    \resizebox{1.0\linewidth}{!}{
 	\begin{tabular}{l|ccc|cccc|cccc|cccc|>{\columncolor{blue!8}}c}
 	\multicolumn{1}{c}{} & \multicolumn{3}{c}{Noise} & \multicolumn{4}{c}{Blur} & \multicolumn{4}{c}{Weather} & \multicolumn{4}{c}{Digital}  \\
 	 Model+Method & Gauss. & Shot & Impul. & Defoc. & Glass & Motion & Zoom & Snow & Frost & Fog & Brit. & Contr. & Elastic & Pixel & JPEG & Avg.  \\
 	\midrule
ResNet50 (GN)+Tent & 5.1 & 6.1 & 5.4 & 15.0 & 10.8 & 22.1 & 22.9 & 25.8 & 33.2 & 3.1 & 70.4 & 42.7 & 11.0 & 48.1 & 54.2 & 25.1  \\
~~\ding{59}~InfoMax~\cite{liang2020we} & 28.8 & 31.1 & 29.9 & 21.4 & 20.2 & 29.4 & 30.9 & 44.3 & 44.0 & 47.5 & 70.1 & 43.4 & 30.2 & 47.6 & 54.5 & 38.2 \\
\rowcolor{pink!30}~~\ding{59}~Ours $I(\cdot)$ (Eqn.~\ref{eq:initial_inequity}) & 41.0 & 43.0 & 41.8 & 32.7 & 31.8 & 40.4 & 44.3 & 54.5 & 54.2 & 58.8 & 72.9 & 52.9 & 47.0 & 58.3 & 59.5 & \textbf{48.9} \\
\cmidrule{1-17}
VitBase (LN)+Tent         &  42.4 & 1.4 & 43.3 & 52.2 & 47.6 & 55.4 & 49.9 & 19.7 & 18.1 & 66.1 & 74.9 & 64.7 & 52.5 & 66.8 & 64.1 & 47.9   \\
~~\ding{59}~InfoMax~\cite{liang2020we} & 40.3 & 38.5 & 41.3 & 50.5 & 46.4 & 52.8 & 48.8 & 55.5 & 56.0 & 64.1 & 72.7 & 62.4 & 53.3 & 63.2 & 61.2 & 53.8 \\
\rowcolor{pink!30}~~\ding{59}~Ours $I(\cdot)$ (Eqn.~\ref{eq:initial_inequity}) & 46.1 & 44.3 & 47.1 & 52.7 & 52.1 & 58.6 & 55.5 & 61.8 & 62.1 & 70.0 & 75.9 & 65.9 & 61.7 & 68.9 & 66.3 & \textbf{59.3}
	\end{tabular}
	}
	 \end{threeparttable}
	 \end{center}
\end{table*}

\subsection{Effects of Large Batch Sizes in BN models under Mixed Domain Shifts} 
In Section~\ref{sec:norm_effects}, we mentioned that a standard batch size (\eg, 64 on ImageNet) works well when there is only one type of distribution shift. However, when test data contains multiple shifts, this batch size fails to estimate an accurate mean and variance in batch normalization layers. Here, we investigate the effects of super large batch sizes in this setting. From Table \ref{tab:superbs_effects_bn_mixed}, the adapted performance increases as the batch size increases, indicating that a larger batch size helps to estimate statistics more accurately. It is worth noting that the performance on severity level 3 degrades when BS is larger than 1024. This is because we fix the learning rate for various batch sizes and in this sense, BS=1024 may lead to insufficient model updates. Moreover, although enlarging batch sizes is able to boost the performance, the adapt performance is still inferior to the average accuracy of adapting on each corruption type separately (\ie, average adapt). This further emphasizes the necessity of exploiting models with group or layer norm layers to perform test-time entropy minimization.

\begin{table*}[h]
    \renewcommand\thetable{P}
    \caption{Effects of large batch sizes (BS) in Tent \cite{wang2021tent} with ResNet-50-BN under \textsc{\textbf{mixture of 15 corruption types}} on ImageNet-C regarding \textbf{Accuracy (\%)}.
    }
    \label{tab:superbs_effects_bn_mixed}
\newcommand{\tabincell}[2]{\begin{tabular}{@{}#1@{}}#2\end{tabular}}
 \begin{center}
 \begin{threeparttable}
    \resizebox{0.8\linewidth}{!}{
 	\begin{tabular}{l|c|c|cccccc}
 	\multicolumn{2}{c}{} & \multicolumn{1}{c}{avg. adapt } & \multicolumn{6}{c}{mix adapt} \\
 	 Severity & Base & BS$=$64 & BS=32 & BS$=$64 & BS$=$128 & BS$=$256 & BS$=$512 & BS$=$1,024  \\
 	\midrule
    Level = 5   & 18.0  & 42.6  & 0.9 & 2.3  & 4.0 & 8.3& 12.4 & 16.4  \\
    Level = 3  & 39.8  & 59.0  & 20.1 & 40.7  & 46.7 & 49.1 & 49.0 & 47.8  \\

	\end{tabular}
	}
	 \end{threeparttable}
	 \end{center}
\end{table*}

\subsection{Performance of Tent with ConvNeXt-LN}

For mainstream neural network models, the normalization layers are often coupled with network architecture. Specifically, group norm (GN) and batch norm (BN) are often combined with conventional networks, while layer norm (LN) is more suitable for transformer networks. Therefore, we investigate the layer normalization effects in TTA in Section~\ref{sec:empirical_norm_effects} through VitBase-LN.
Here, we conduct more experiments to compare the performance of online entropy minimization~\cite{wang2021tent} on ResNet50-BN and ConvNeXt-LN~\cite{liu2022convnet}. ConvNeXt is a convolutional network equipped with LN. The authors conduct significant modifications over ResNet to make this LN-based convolutional network work well, such as modifying the architecture (ResNet block to ConvNeXt block, activation functions, etc.), various training strategies (stochastic depth, random erasing, EMA, etc.). From Table~\ref{tab:convnext}, Tent+ConvNeXt-LN performs more stable than Tent+ResNet50-BN, but still suffers several failure cases. These results are consistent with that of ResNet50-BN \vs~VitBase-LN.

\begin{table*}[ht]
    \renewcommand\thetable{Q}
    \caption{Results of Tent on ResNet50-BN and ConvNeXt-LN. We report \textbf{Accuracy (\%)} on ImageNet-C under \textbf{online imbalanced label distribution shifts} and the imbalance ratio is $\infty$.
    }
    \label{tab:convnext}
\newcommand{\tabincell}[2]{\begin{tabular}{@{}#1@{}}#2\end{tabular}}
 \begin{center}
 \begin{threeparttable}
 \large
    \resizebox{1.0\linewidth}{!}{
 	\begin{tabular}{l|ccc|cccc|cccc|cccc|>{\columncolor{blue!8}}c}
 	\multicolumn{1}{c}{} & \multicolumn{3}{c}{Noise} & \multicolumn{4}{c}{Blur} & \multicolumn{4}{c}{Weather} & \multicolumn{4}{c}{Digital}  \\
 	 Severity level 5 & Gauss. & Shot & Impul. & Defoc. & Glass & Motion & Zoom & Snow & Frost & Fog & Brit. & Contr. & Elastic & Pixel & JPEG & Avg.  \\
 	\midrule

        ResNet50-BN  & 2.2    & 2.9  & 1.8    & 17.8   & 9.8   & 14.5   & 22.5 & 16.8 & 23.4  & 24.6 & 59.0  & 5.5    & 17.1    & 20.7  & 31.6 & 18.0  \\ 
        ~~$\bullet~$Tent~\cite{wang2021tent}        & 1.2    & 1.4  & 1.4    & 1.0    & 0.9   & 1.2    & 2.6  & 1.7  & 1.8   & 3.6  & 5.0   & 0.5    & 2.6     & 3.2   & 3.1  & 2.1  \\ %
        ConvNeXt-LN  & 52.3   & 52.7 & 52.3   & 31.7   & 18.7  & 42.5   & 38.1 & 54.2 & 58.3  & 50.6 & 75.6  & 56.8   & 32.2    & 39.2  & 60.4 & 47.7  \\ 
        ~~$\bullet~$Tent~\cite{wang2021tent}        & 26.3   & 11.9 & 36.9   & 31.1   & 12.7  & 14.6   & 5.1  & 7.8  & 5.3   & 6.6  & 79.0  & 67.6   & 1.5     & 68.4  & 65.8 & 29.4 \\
\cmidrule{1-17}
 	 Severity level 3 & Gauss. & Shot & Impul. & Defoc. & Glass & Motion & Zoom & Snow & Frost & Fog & Brit. & Contr. & Elastic & Pixel & JPEG & Avg.  \\
 	\cmidrule{1-17}
        ResNet50-BN  & 27.7   & 25.2 & 25.1   & 37.8   & 16.7  & 37.8   & 35.3 & 35.2 & 32.1  & 46.7 & 69.5  & 46.2   & 55.4    & 46.2  & 59.4 & 39.8 \\ 
        ~~$\bullet~$Tent~\cite{wang2021tent}        & 3.4    & 3.2  & 3.2    & 2.3    & 2.0   & 2.4    & 3.4  & 2.4  & 2.4   & 4.6  & 5.4   & 3.0    & 4.8     & 4.6   & 4.5  & 3.4  \\ %
        ConvNeXt-LN  & 71.1   & 69.8 & 72.4   & 55.2   & 36.6  & 64.3   & 52.8 & 63.2 & 64.0  & 66.0 & 79.4  & 75.8   & 69.1    & 68.5  & 71.9 & 65.4  \\ 
        ~~$\bullet~$Tent ~\cite{wang2021tent}       & 73.4   & 73.5 & 74.4   & 66.8   & 61.6  & 71.4   & 64.2 & 22.6 & 39.5  & 76.8 & 81.2  & 78.9   & 76.8    & 75.6  & 75.8 & 67.5 \\
	\end{tabular}
	}
	 \end{threeparttable}
	 \end{center}
\end{table*}

\subsection{Effectiveness of Model Recovery Scheme with Tent and EATA}

In this subsection, we apply our Model Recovery scheme to Tent~\cite{wang2021tent} and EATA~\cite{niu2022EATA}. From Table~\ref{tab:tent_eata_recover}, the model recovery indeed helps Tent a lot (\eg, the average accuracy from 22.0\% to 26.1\% on ResNet50-GN) while its performance gain on EATA is a bit marginal. Compared with Tent+recovery and EATA+recovery, our \mysar / \mysarE greatly boosts the adaptation performance, \eg, the average accuracy 26.1\% (Tent+recovery) \vs~37.2\% (SAR) on ResNet50-GN, suggesting the effectiveness of our proposed methods. \rpami{It is worth noting that our model recovery scheme has also been adopted by the recent state-of-the-art method ReCAP~\cite{hu2025beyond}, where it is triggered to restore model performance during continuous TTA under online imbalanced label shifts, as shown in Table~\ref{tab:continuous_label_shifts} (\textit{contrast} on ResNet50-GN).}

\begin{table*}[ht]
    \renewcommand\thetable{R}
    \caption{Results of combining model recovery scheme with Tent and EATA. We report the \textbf{Accuracy (\%)} on ImageNet-C severity level 5 under \textbf{\textsc{online imbalanced label distribution shifts}} and the imbalance ratio is $\infty$.
    }
    \label{tab:tent_eata_recover}
\newcommand{\tabincell}[2]{\begin{tabular}{@{}#1@{}}#2\end{tabular}}
 \begin{center}
 \begin{threeparttable}
 \large
    \resizebox{1.0\linewidth}{!}{
 	\begin{tabular}{l|ccc|cccc|cccc|cccc|>{\columncolor{blue!8}}c}
 	\multicolumn{1}{c}{} & \multicolumn{3}{c}{Noise} & \multicolumn{4}{c}{Blur} & \multicolumn{4}{c}{Weather} & \multicolumn{4}{c}{Digital}  \\
 	 Model+Method & Gauss. & Shot & Impul. & Defoc. & Glass & Motion & Zoom & Snow & Frost & Fog & Brit. & Contr. & Elastic & Pixel & JPEG & Avg.  \\
 	\midrule
ResNet50 (GN) &   18.0    & 19.8 &  17.9  &  19.8  & 11.4  &  21.4  & 24.9 & 40.4 & 47.3  & 33.6 & 69.3  &  36.3  &  18.6   & 28.4  & 52.3 &  30.6  \\
~~$\bullet~$Tent            &  2.6   & 3.3  &  2.7   &  13.9  &  7.9  &  19.5  & 17.0 & 16.5 & 21.9  & 1.8  & 70.5  &  42.2  &   6.6   & 49.4  & 53.7 &   22.0 \\
~~$\bullet~$Tent+recover &  10.1  & 12.2 &  10.6  &  13.9  &  8.5  &  19.5  & 20.6 & 24.3 & 33.5  & 8.9  & 70.5  &  42.2  &  13.5   & 49.4  & 53.7 &   26.1    \\
~~$\bullet~$EATA         &  27.0  & 28.3 &  28.1  &  14.9  & 17.1  &  24.4  & 25.3 & 32.2 & 32.0  & 39.8 & 66.7  &  33.6  &  24.5   & 41.9  & 38.4 &   31.6   \\
~~$\bullet~$EATA+recover  &  26.1  & 31.0 &  27.2  &  19.9  & 18.5  &  25.7  & 25.7 & 35.9 & 28.6  & 40.4 & 68.2  &  35.3  &  27.6   & 42.9  & 40.9 &   32.9   \\
\rowcolor{pink!15}~~$\bullet~$SAR (ours)  & 33.1$_{\pm1.0}$ & 36.5$_{\pm0.4}$ & 35.5$_{\pm1.1}$ & 19.2$_{\pm0.4}$ & 19.5$_{\pm1.2}$ & 33.3$_{\pm0.5}$ & 27.7$_{\pm4.0}$ & 23.9$_{\pm5.1}$ & 45.3$_{\pm0.4}$ & 50.1$_{\pm1.0}$ & 71.9$_{\pm0.1}$ & 46.7$_{\pm0.2}$ & 7.1$_{\pm1.8}$ & 52.1$_{\pm0.5}$ & 56.3$_{\pm0.1}$ & 37.2$_{\pm0.6}$ \\

\rowcolor{pink!30}~~$\bullet~$\mysarE (ours) & \textbf{43.0$_{\pm 0.3}$} & \textbf{45.8$_{\pm 0.5}$} & \textbf{43.9$_{\pm 0.5}$} & \textbf{35.5$_{\pm 0.6}$} & \textbf{36.5$_{\pm 0.5}$} & \textbf{44.5$_{\pm 0.2}$} & \textbf{48.7$_{\pm 0.2}$} & \textbf{56.1$_{\pm 0.3}$} & \textbf{55.0$_{\pm 0.2}$} & \textbf{61.0$_{\pm 0.3}$} & \textbf{73.4$_{\pm 0.1}$} & \textbf{55.3$_{\pm 0.1}$} & \textbf{54.0$_{\pm 0.2}$} & \textbf{61.9$_{\pm 0.1}$} & \textbf{61.0$_{\pm 0.2}$} & \textbf{51.7$_{\pm 0.1}$} \\
\cmidrule{1-17}
VitBase (LN)  & 9.4 & 6.7  &  8.3   &  29.1  & 23.4  &  34.0  & 27.0 & 15.8 & 26.3  & 47.4 & 54.7  &  43.9  &  30.5   & 44.5  & 47.6 &  29.9   \\
~~$\bullet~$Tent         &  32.7  & 1.4  &  34.6  &  54.4  & 52.3  &  58.2  & 52.2 & 7.7  & 12.0  & 69.3 & 76.1  &  66.1  &  56.7   & 69.4  & 66.4 &   47.3   \\
~~$\bullet~$Tent+recover &  40.3  & 10.1 &  42.4  &  54.4  & 52.3  &  58.1  & 52.2 & 31.6 & 39.2  & 69.3 & 76.1  &  66.1  &  56.7   & 69.4  & 66.4 &   52.3   \\
~~$\bullet~$EATA          &  35.9  & 34.6 &  36.7  &  45.3  & 47.2  &  49.3  & 47.7 & 56.5 & 55.4  & 62.2 & 72.2  &  21.7  &  56.2   & 64.7  & 63.7 &   50.0  \\
~~$\bullet~$EATA+recover &  35.9  & 34.6 &  36.7  &  45.3  & 47.2  &  49.3  & 47.7 & 56.5 & 55.4  & 62.2 & 72.2  &  21.7  &  56.2   & 64.7  & 63.7 &   49.9   \\
\rowcolor{pink!15}~~$\bullet~$SAR (ours) & 46.5$_{\pm3.0}$ & 43.1$_{\pm7.4}$ & 48.9$_{\pm0.4}$ & 55.3$_{\pm0.1}$ & 54.3$_{\pm0.2}$ & 58.9$_{\pm0.1}$ & 54.8$_{\pm0.2}$ & 53.6$_{\pm7.1}$ & 46.2$_{\pm3.5}$ & 69.7$_{\pm0.3}$ & 76.2$_{\pm0.1}$ & 66.2$_{\pm0.3}$ & 60.9$_{\pm0.3}$ & 69.6$_{\pm0.1}$ & 66.6$_{\pm0.1}$ & 58.0$_{\pm0.5}$ \\
\rowcolor{pink!30}~~$\bullet~$\mysarE (ours) & \textbf{53.7$_{\pm 0.1}$} & \textbf{54.5$_{\pm 0.3}$} & \textbf{55.0$_{\pm 0.3}$} & \textbf{57.8$_{\pm 0.2}$} & \textbf{59.0$_{\pm 0.1}$} & 63.6$_{\pm 0.2}$ & \textbf{61.7$_{\pm 0.1}$} & \textbf{67.8$_{\pm 0.2}$} & 66.0$_{\pm 0.1}$ & \textbf{73.3$_{\pm 0.1}$} & \textbf{77.8$_{\pm 0.1}$} & \textbf{67.4$_{\pm 0.1}$} & \textbf{68.7$_{\pm 0.2}$} & \textbf{73.4$_{\pm 0.2}$} & \textbf{70.1$_{\pm 0.1}$} & \textbf{64.6$_{\pm 0.1}$} \\
	\end{tabular}
	}
	 \end{threeparttable}
	 \end{center}
\end{table*}

\end{document}